\RequirePackage{fix-cm}
\documentclass[twocolumn]{svjour3}          %
\smartqed  %
\usepackage[dvipsnames]{xcolor}
\usepackage{graphicx}
\usepackage{epsfig}
\usepackage{multirow}
\usepackage{hyperref}
\usepackage{amsmath}
\usepackage{amssymb}
\usepackage[caption=false]{subfig}
\graphicspath{{imgs/}{imgs/intro/}{imgs/visual_mapper/}{imgs/eval/visual_compare/}{imgs/eval/on_color_norm/}{imgs/eval/dense/}{imgs/eval/on_weak_area/}{imgs/eval/on_clusters/}}

\renewcommand{\d}[1]{\mbox{\boldmath$#1$}}

\newcommand{\s}[1]{{\un{#1}}}

\newcommand{\q}[1]{{{\bf #1}}}

\newcommand{\mq}[1]{{\q #1}}

\newcommand{\mC}{\mathchoice {\setbox0=\hbox{$\displaystyle\rm
C$}\hbox{\hbox to0pt{\kern0.4\wd0\vrule height0.9\ht0\hss}\box0}}
{\setbox0=\hbox{$\textstyle\rm C$}\hbox{\hbox
to0pt{\kern0.4\wd0\vrule height0.9\ht0\hss}\box0}}
{\setbox0=\hbox{$\scriptstyle\rm C$}\hbox{\hbox
to0pt{\kern0.4\wd0\vrule height0.9\ht0\hss}\box0}}
{\setbox0=\hbox{$\scriptscriptstyle\rm C$}\hbox{\hbox
to0pt{\kern0.4\wd0\vrule height0.9\ht0\hss}\box0}}}

\newcommand{\mG}{\mathchoice {\setbox0=\hbox{$\displaystyle\rm
G$}\hbox{\hbox to0pt{\kern0.4\wd0\vrule height0.9\ht0\hss}\box0}}
{\setbox0=\hbox{$\textstyle\rm G$}\hbox{\hbox
to0pt{\kern0.4\wd0\vrule height0.9\ht0\hss}\box0}}
{\setbox0=\hbox{$\scriptstyle\rm G$}\hbox{\hbox
to0pt{\kern0.4\wd0\vrule height0.9\ht0\hss}\box0}}
{\setbox0=\hbox{$\scriptscriptstyle\rm G$}\hbox{\hbox
to0pt{\kern0.4\wd0\vrule height0.9\ht0\hss}\box0}}}

\newcommand{\mJ}{\mathchoice {\setbox0=\hbox{$\displaystyle\rm
J$}\hbox{\hbox to0pt{\kern0.4\wd0\vrule height0.9\ht0\hss}\box0}}
{\setbox0=\hbox{$\textstyle\rm J$}\hbox{\hbox
to0pt{\kern0.4\wd0\vrule height0.9\ht0\hss}\box0}}
{\setbox0=\hbox{$\scriptstyle\rm J$}\hbox{\hbox
to0pt{\kern0.4\wd0\vrule height0.9\ht0\hss}\box0}}
{\setbox0=\hbox{$\scriptscriptstyle\rm J$}\hbox{\hbox
to0pt{\kern0.4\wd0\vrule height0.9\ht0\hss}\box0}}}

\newcommand{\mO}{\mathchoice {\setbox0=\hbox{$\displaystyle\rm
O$}\hbox{\hbox to0pt{\kern0.4\wd0\vrule height0.9\ht0\hss}\box0}}
{\setbox0=\hbox{$\textstyle\rm O$}\hbox{\hbox
to0pt{\kern0.4\wd0\vrule height0.9\ht0\hss}\box0}}
{\setbox0=\hbox{$\scriptstyle\rm O$}\hbox{\hbox
to0pt{\kern0.4\wd0\vrule height0.9\ht0\hss}\box0}}
{\setbox0=\hbox{$\scriptscriptstyle\rm O$}\hbox{\hbox
to0pt{\kern0.4\wd0\vrule height0.9\ht0\hss}\box0}}}

\newcommand{\mQ}{\mathchoice {\setbox0=\hbox{$\displaystyle\rm
Q$}\hbox{\raise 0.15\ht0\hbox to0pt{\kern0.4\wd0\vrule
height0.8\ht0\hss}\box0}}{\setbox0=\hbox{$\textstyle\rm Q$}\hbox{\raise
0.15\ht0\hbox to0pt{\kern0.4\wd0\vrule height0.8\ht0\hss}\box0}}
{\setbox0=\hbox{$\scriptstyle\rm Q$}\hbox{\raise 0.15\ht0\hbox
to0pt{\kern0.4\wd0\vrule height0.7\ht0\hss}\box0}}{\setbox0=
\hbox{$\scriptscriptstyle\rm Q$}\hbox{\raise 0.15\ht0\hbox
to0pt{\kern0.4\wd0\vrule height0.7\ht0\hss}\box0}}}

\newcommand{\mS}{\mathchoice
{\setbox0=\hbox{$\displaystyle     \rm S$}\hbox{\raise0.5\ht0\hbox
to0pt{\kern0.35\wd0\vrule height0.45\ht0\hss}\hbox
to0pt{\kern0.55\wd0\vrule height0.5\ht0\hss}\box0}}
{\setbox0=\hbox{$\textstyle        \rm S$}\hbox{\raise0.5\ht0\hbox
to0pt{\kern0.35\wd0\vrule height0.45\ht0\hss}\hbox
to0pt{\kern0.55\wd0\vrule height0.5\ht0\hss}\box0}}
{\setbox0=\hbox{$\scriptstyle      \rm S$}\hbox{\raise0.5\ht0\hbox
to0pt{\kern0.35\wd0\vrule height0.45\ht0\hss}\raise0.05\ht0\hbox
to0pt{\kern0.5\wd0\vrule height0.45\ht0\hss}\box0}}
{\setbox0=\hbox{$\scriptscriptstyle\rm S$}\hbox{\raise0.5\ht0\hbox
to0pt{\kern0.4\wd0\vrule height0.45\ht0\hss}\raise0.05\ht0\hbox
to0pt{\kern0.55\wd0\vrule height0.45\ht0\hss}\box0}}}
\newcommand{\mT}{\mathchoice {\setbox0=\hbox{$\displaystyle\rm
T$}\hbox{\hbox to0pt{\kern0.3\wd0\vrule height0.9\ht0\hss}\box0}}
{\setbox0=\hbox{$\textstyle\rm T$}\hbox{\hbox
to0pt{\kern0.3\wd0\vrule height0.9\ht0\hss}\box0}}
{\setbox0=\hbox{$\scriptstyle\rm T$}\hbox{\hbox
to0pt{\kern0.3\wd0\vrule height0.9\ht0\hss}\box0}}
{\setbox0=\hbox{$\scriptscriptstyle\rm T$}\hbox{\hbox
to0pt{\kern0.3\wd0\vrule height0.9\ht0\hss}\box0}}}
\newcommand{\mU}{\mathchoice {\setbox0=\hbox{$\displaystyle\rm
U$}\hbox{\hbox to0pt{\kern0.4\wd0\vrule height0.9\ht0\hss}\box0}}
{\setbox0=\hbox{$\textstyle\rm U$}\hbox{\hbox
to0pt{\kern0.4\wd0\vrule height0.9\ht0\hss}\box0}}
{\setbox0=\hbox{$\scriptstyle\rm U$}\hbox{\hbox
to0pt{\kern0.4\wd0\vrule height0.9\ht0\hss}\box0}}
{\setbox0=\hbox{$\scriptscriptstyle\rm U$}\hbox{\hbox
to0pt{\kern0.4\wd0\vrule height0.9\ht0\hss}\box0}}}

\journalname{Journal of Field Robotics}
\begin{document}
\sloppy

\title{Semihierarchical Reconstruction and Weak-area Revisiting for Robotic Visual Seafloor Mapping\\
}

\author{$^*$Mengkun She      \and
        Yifan Song		 \and
        David Nakath	 \and
        Kevin K\"oser
}

\institute{
	$^*$Corresponding author\\
	$^*$M. She, D. Nakath, K. K\"oser
	\at
	Marine Data Science\\
	Christian-Albrechts-University of Kiel\\
	Kiel, Germany\\
	\email{\{mshe,dna,kk\}@informatik.uni-kiel.de}           %
	\and
	$^*$M. She, D. Nakath, Y. Song, K.  K\"oser \at
	GEOMAR Helmholtz Centre for Ocean Research Kiel\\
	Kiel, Germany \\
	Tel.: ++49 431 600 2272,  \\
	\email{ysong@geomar.de}
}

\date{Received: date / Accepted: date}

\maketitle

\begin{abstract}	

Despite impressive results achieved by many on-land visual mapping algorithms in the recent decades, transferring these methods from land to the deep sea remains a challenge due to harsh environmental conditions. 
Images captured by autonomous underwater vehicles (AUVs), equipped with high-resolution cameras and artificial illumination systems, often suffer from heterogeneous illumination and quality degradation caused by attenuation and scattering, on top of refraction of light rays. 
These challenges often result in the failure of on-land SLAM approaches when applied underwater or cause SfM approaches to exhibit drifting or omit challenging images.
Consequently, this leads to gaps, jumps, or weakly reconstructed areas.
In this work, we present a navigation-aided hierarchical reconstruction approach to facilitate the automated robotic 3D reconstruction of hectares of seafloor.
Our hierarchical approach combines the advantages of SLAM and global SfM that is much more efficient than incremental SfM, while ensuring the completeness and consistency of the global map.
This is achieved through identifying and revisiting problematic or weakly reconstructed areas, avoiding to omit images and making better use of limited dive time.
The proposed system has been extensively tested and evaluated during several research cruises, demonstrating its robustness and practicality in real-world conditions.

\keywords{Seafloor mapping \and Underwater imaging \and Structure-from-motion \and Robotic mapping}
\end{abstract}

\section{Introduction}
\label{sec:intro}
\begin{figure*}[!ht]
	\centering
	\includegraphics[height=3.7cm]{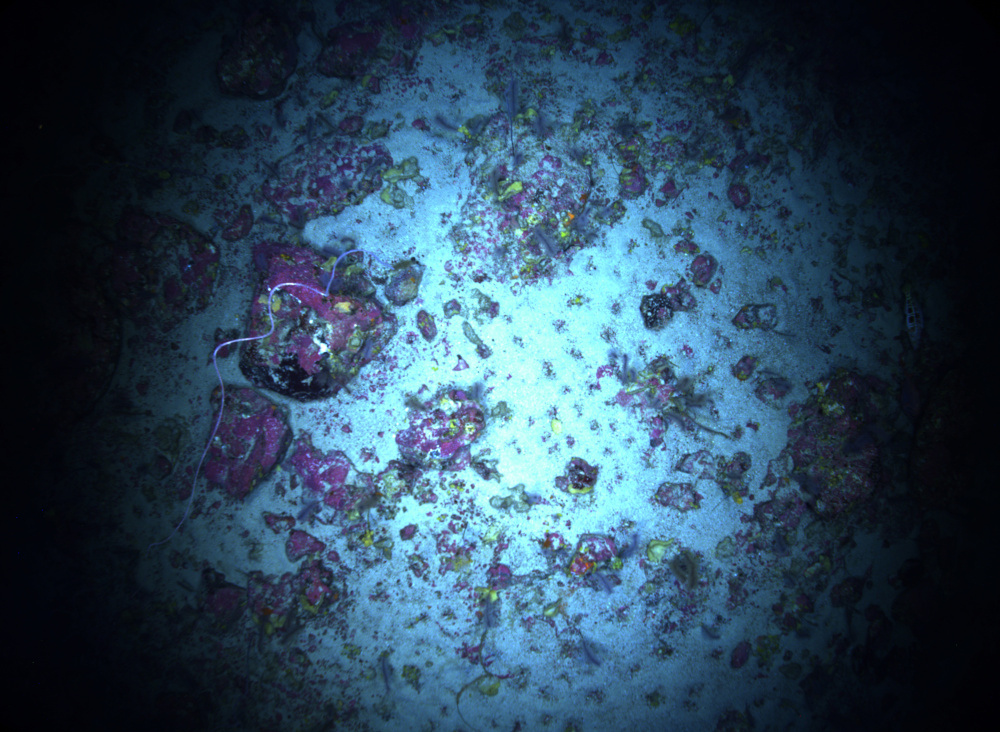}
	\includegraphics[height=3.7cm]{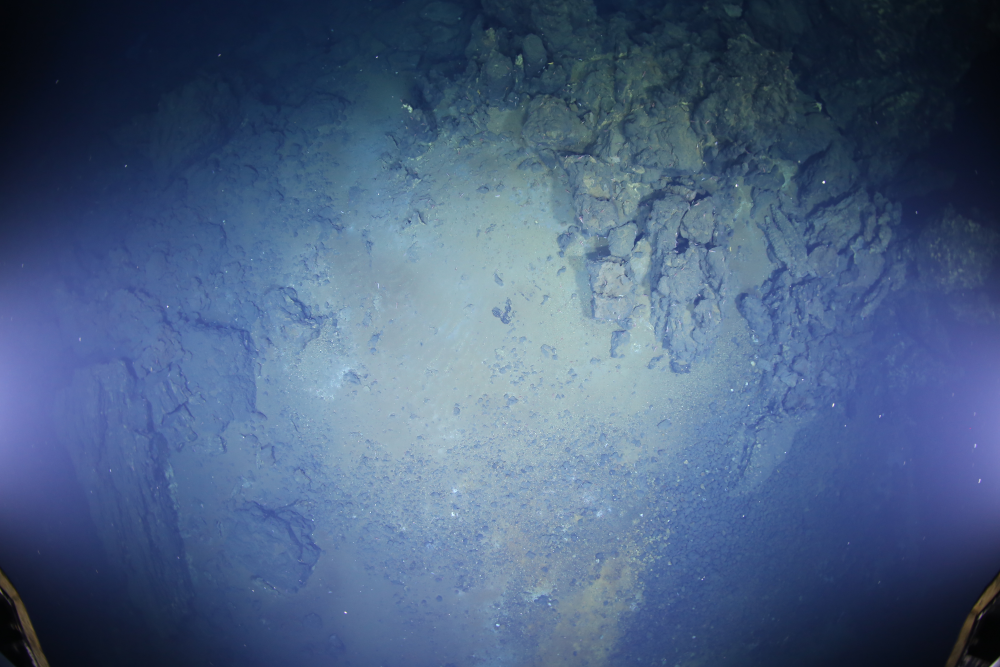}
	\includegraphics[height=3.7cm]{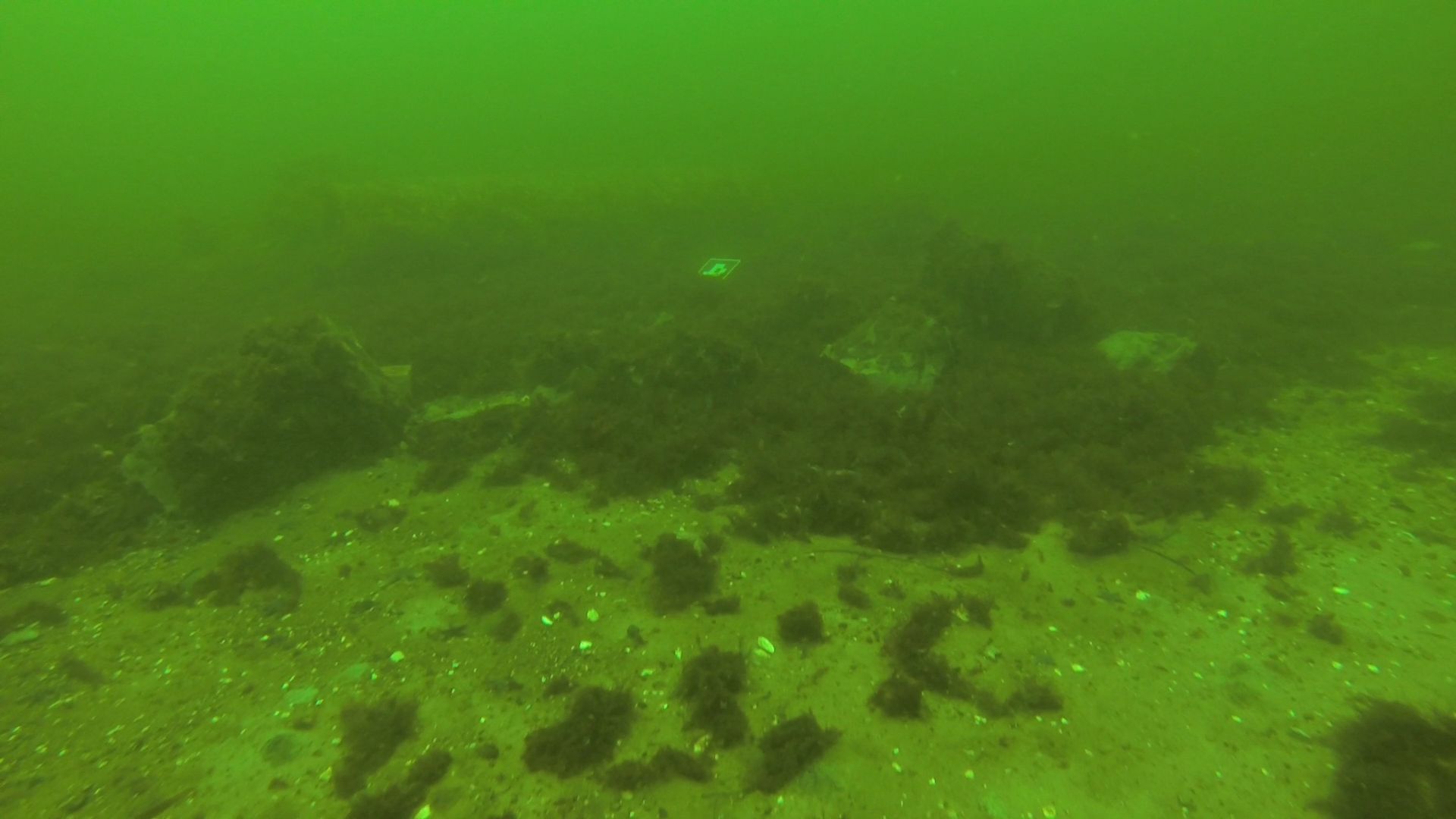}
	\caption{Example underwater images. Due to light attenuation and scattering effects, the images appear to have a strong blue and green hue. \textbf{Left} and \textbf{center} also show the heterogeneous illumination, which is caused by the moving light source mounted on the robot. Image courtesy: GEOMAR and Schmidt Ocean Institution.}
	\label{fig:example_seafloor_imgs}
\end{figure*}

More than half of Earth's surface is covered by the deep ocean with at least one kilometer of water depth and there is a growing interest in exploring this uncharted terrain.
This is typically carried out by robotic platforms with various sensing technologies such as cabled and remotely operated vehicles (ROVs) \cite{Sedlazeck_2009-3DrecUnderwater} or those that navigate autonomously (AUVs) \cite{leonard2016autonomous,Johnson_2010-3DrecAUV}.
In particular, optical images have become increasingly attractive due to their high-resolution and suitability for human interpretation, making them a complementary survey technology to ship-based acoustic methods such as side scan sonar or multi-beam echo sounders \cite{kwasnitschka2016deepsurveycam}.
Over the past decades, numerous visual mapping algorithms on land have achieved remarkable outcomes. 
State-of-the-art Structure-from-Motion (SfM) algorithms, like COLMAP \cite{schonberger2016structure}, demonstrate high robustness in reconstruction of on-land scenarios. 
However, in contrast to reconstructions from highly redundant photo collections  \cite{Snavely_2008bundler,agarwal2010reconstructing,frahm10rome}, robotic photogrammetric surveys of the deep sea have to minimize redundancy for cost reasons and must use the available dive time as efficient as possible. 
Consequently, where photo-collection approaches simply discard difficult images or drop connections for efficiency, deep sea robotic surveys must try to register as many images as possible and at the same time achieve the best reconstruction even under challenging situations.
Unfortunately, in the underwater domain, when applying state-of-the-art Simultaneous Localization and Mapping (SLAM) approaches \cite{mur2017orb,engel2014lsd,leutenegger2015keyframe}, significant positional errors, divergence or tracking loss, or even failure have been observed \cite{joshi2019experimental,koser2020challenges,grimaldi2023investigation}.
As a result, directly transferring these technologies to the deep-sea remains challenging due to the harsh and unique environmental conditions encountered in this setting.

Firstly, artificial illumination systems on deep-diving robots are essential to provide sufficient lighting for image capture in deeper waters. 
However, the presence of co-moving light sources can violate the photo-consistency assumption, particularly when the lighting conditions vary significantly between images \cite{schoentag2022uwcorrespondence}.
This can reduce quantity and robustness of feature correspondences, which are the key to accurately aligning images during reconstruction. 
Additionally, the effective field-of-view of the cameras may be constrained by the illumination cone, potentially resulting in a loss of image content (see Fig. \ref{fig:example_seafloor_imgs}, left and center).
Consequently, this limitation can lead to a reduction in image overlap, especially side overlap, when capturing seafloor images.
Such reduced overlap can result in inconsistent reconstructions due to limited feature matches between adjacent tracks.
Secondly, radiometric degradation caused by light scattering and attenuation can introduce a layer of haze in the image, reducing its contrast, and affect its color.
To facilitate deep sea / artificial light mapping and to create an artefact-free mosaic of the target scene, effective image restoration techniques are required, but are largely missing \cite{song2022optical}, except for simple scenarios (e.g. \cite{koser2021robustly}).
Moreover, from a geometrical perspective, the refraction of incident light rays at the water-pressure-housing interface alters their directions, which complicates the image formation model \cite{JordtSedlazeck_2012RefCalibUnw,she2021refractive}.
Therefore, special calibration methods should be employed to account for refraction effects \cite{she2019adjustment,she2021refractive}. 

The above challenges are specific to the underwater scenario, but there is another crucial aspect that poses a challenge in general visual mapping scenarios: scalability and efficiency.
While many current state-of-the-art SfM \cite{schonberger2016structure} approaches are incremental and have demonstrated their accuracy and robustness, they suffer from inefficiency when dealing with large-scale datasets. 
The computational complexity of incremental SfM increases about cubically with the number of images, making them impractical for large-scale mapping tasks. 
Additionally, they are prone to drift due to error accumulation over time.
To address this, researchers have proposed Global SfM methods to handle large-scale scenes by leveraging the concept of averaging rotational and translational inconsistencies of all images over the entire epipolar graph \cite{wilson2014robust,chatterjee2013efficient,ozyesil2015robust}.
By doing so, they reduce the drift and avoid the need for repeated bundle adjustment when adding images to the reconstruction. 
Furthermore, the divide-and-conquer approach has been employed to address even larger-scale scenes by partitioning the view graph into smaller clusters and performing separate reconstructions in parallel.
Subsequently, these sub-reconstructions are merged to create a global and consistent reconstruction \cite{bhowmick2015divide,zhu2017parallel,chen2020graph}.
However, global SfM approaches come with their own limitations.
One limitation is a lack of effective outlier filtering and an intrinsic degeneracy that relates to the missing scale in visual two-view relations (relative rotation and translation direction): For scenarios, where robots move predominantly forward in a fixed direction like in lawnmower patterns, absolute camera positions become ambiguous at the global SfM level.

Therefore in this work, we focus on the practical challenges and leverage the recent developments in both underwater imaging and visual mapping and present an automated approach that can map large areas of the seafloor (hectares) efficiently while avoiding to discard difficult imagery.%

Our contributions are mainly three-fold:
Firstly, we carefully consider refraction to avoid reconstruction biases \cite{she2019adjustment,she2021refractive}, which has not received much attention in earlier seminal underwater SLAM/SfM works \cite{Johnson_2010-3DrecAUV,bodenmann2017generation,figueira2015accuracy,skinner2017automatic,ridao2010visual,arnaubec2023underwater,joshi2022high}. 
Secondly, we propose a navigation-aided hierarchical reconstruction approach that combines the advantages of SLAM and global SfM to achieve a robust and efficient reconstruction of large underwater scenes.
Specifically, we partition the large scene into smaller clusters and perform local SfM reconstructions concurrently.
Subsequently, a global pose graph optimization is performed to obtain a consistent camera trajectory using the upgraded view graph resulted from the local reconstructions.
Thirdly, to improve the quality of the pose graph optimization, we propose to identify weakly reconstructed areas during the local reconstruction phase and revisit them to reduce inconsistencies in the resulting camera trajectory.
We conduct detailed and extensive evaluations on several real-world datasets with various visual characteristics, demonstrating that the proposed approach achieves much more physically consistent results, i.e., the vehicle exhibits expected motions without gaps, jumps, than vanilla COLMAP on difficult datasets while reducing processing time by on average of $76.7\%$.
Our ablation study further confirms the effectiveness and necessity of the proposed approach.

\section{Related Work}\label{sec:related_work}

Visual mapping is a fundamental process in robotics that involves recovering 3D structure and camera poses of images captured from different viewpoints \cite{schonberger2016structure,jiang2020efficient,joshi2023sm}.
Depending on the purpose of the application, visual mapping can be categorized into two main categories: those that focus on the map quality and those that focus on online operation ability.
The former case mostly refers to approaches that rely on the SfM (Structure-from-Motion) algorithm, while the latter case is mainly referred to as SLAM (Simultaneous Localization and Mapping).
Earlier deep-sea/underwater visual mapping systems mostly focused on 2D image mosaicking \cite{sawhney1998robust,sawhney1999true}.
Ridao et al. \cite{ridao2010visual} improved the photomosaicking results by including a loop-detection module and considering non-consecutive images.
They use this approach for automatically inspecting underwater dams.
Moving beyond 2D information, Johnson‐Roberson et al. \cite{Johnson_2010-3DrecAUV} presented a complete, robust and automated system for large-scale 3D reconstruction and visualization of the seafloor using images fused with vehicle's navigation data.
The underlying SLAM system \cite{mahon2008efficient} used by this work was based on an information filter. 
However, the two systems presented above depend on depth information obtained from a stereo camera system while we work on purely monocular image data.
Although the ability for online operation is a future pursuit, it is currently not practical due to the high cost of losing an expensive deep-sea robot. Furthermore, researchers have identified the challenge of finding good features on visually degraded images as a major reason for the failure of live-SLAM systems \cite{joshi2019experimental,koser2020challenges,grimaldi2023investigation}.
Instead, SfM has greater flexibility in selecting the images and the order in which they are reconstructed.
This also provides the opportunity to choose more robust and high-quality features for the reconstruction process.
Therefore, in this work, our primary focus is on generating high-quality maps in an offline mode.

SfM has shown remarkable performance in various applications, including urban scenes \cite{pollefeys2008detailed} and unordered internet photo collections \cite{Snavely_2008bundler,snavely2006photo,agarwal2010reconstructing,schonberger2016structure}.
And it is commonly used for photogrammetry surveying, not only in on-land environments but also in underwater settings \cite{menna2022combined,Drap12,nocerino2020coral}.
However, it is important to note that these achievements primarily focus on incremental SfM, which is also known to suffer from drift accumulation and inefficiency when dealing with large-scale datasets, as mentioned earlier in Sect. \ref{sec:intro}.
Global SfM approaches on the other hand, consider the entire reconstruction problem at once, typically involving three major steps: 1) view graph construction by feature extraction and matching, 2) estimation of global poses using the view graph and 3) triangulation of scene structure and bundle adjustment.
Since camera poses are solved non-incrementally using relative constraints derived from the view graph, every image is treated equally and all sequential and loop-closure constrains are included in the optimization, helping to mitigate drift.
However, there are two main challenges in global SfM.
First, the epipolar geometries derived from the view graph can be noisy, even with one layer of outlier rejection by RANSAC during geometric verification. 
Second, the epipolar geometry only provides a relative rotation and a translation direction, resulting in a 5-DOFs relative pose estimate, making it difficult to accurately determine absolute camera positions.
To address these issues, many works have focused on robustly averaging rotations from noisy inputs, which has matured in recent years \cite{chatterjee2013efficient,martinec2007robust,sweeney2015optimizing,chen2021hybrid}.
As for the second challenge, 1DSfM \cite{wilson2014robust} addresses this issue by filtering out unreliable measurements through the projection of translation directions to one of the axes, ensuring that the order constraints are maintained.
They also introduce a cost function to minimize translation directions, although they do not address the issue of scale ambiguities.
\"Ozyesil and Singer \cite{ozyesil2015robust} utilized parallel rigidity theory to identify images where the absolute camera positions can be determined uniquely.
However these works mainly focus on the reconstruction of unordered internet photos, whereas in the deep-sea AUV mapping scenario, the AUVs typically move forward and capture images at a constant interval.
The resulting view graph formed by these images typically does not exhibit parallel rigidity, and absolute camera positions cannot be determined uniquely.
Moreover, for completeness and cost reasons, we seek to reconstruct all images if possible even with challenging conditions because each AUV dive is expensive and the target area may only be visited a limited number of times.

To tackle the above challenges, we perform a full incremental SfM on a small local region of images.
This is because small-to-medium-scale incremental SfM has been proven to be accurate and reliable.
By doing so, we obtain an upgraded view graph that contains the full 6-DOFs relative motions, and the reconstructed 3D points in this local region indicate the true inliers.
Using the upgraded view graph, we can then construct a global pose graph optimizer to perform motion averaging and solve for all camera poses.
Although the idea of combining incremental SfM with global SfM has been proposed before by Cui et al. \cite{cui2017hsfm}, we are inherently different since they first perform global rotation averaging to obtain rotations for all cameras, and then incrementally reconstruct all images with fixed rotations.
The most similar work to us is \cite{chen2020graph}, where they propose a novel algorithm for scene partitioning and a framework for performing parallel/distributed incremental SfM on partitioned clusters, followed by a merging step to form a global reconstruction.
However, incremental SfM on each cluster produces an arbitrarily scaled sub-reconstruction, requiring the estimation of similarity transformations based on overlapping images for merging, which makes the partitioning of the scene crucial.
In contrast, our approach addresses a different scenario.
Modern AUVs come with their own navigation systems for self-navigating in the water.
Although we have no strict requirement on the accuracy of the navigation system, we incorporate the navigation data into incremental SfM to produce a fixed-scale reconstruction, thus relaxing the need for scene partitioning.

\section{System Overview and Preparation}
\label{sec:system_overview}

\subsection{The AUVs and the CoraMo Camera System}

\begin{figure*}[!ht]
	\begin{center}
		\includegraphics[height=3.8cm]{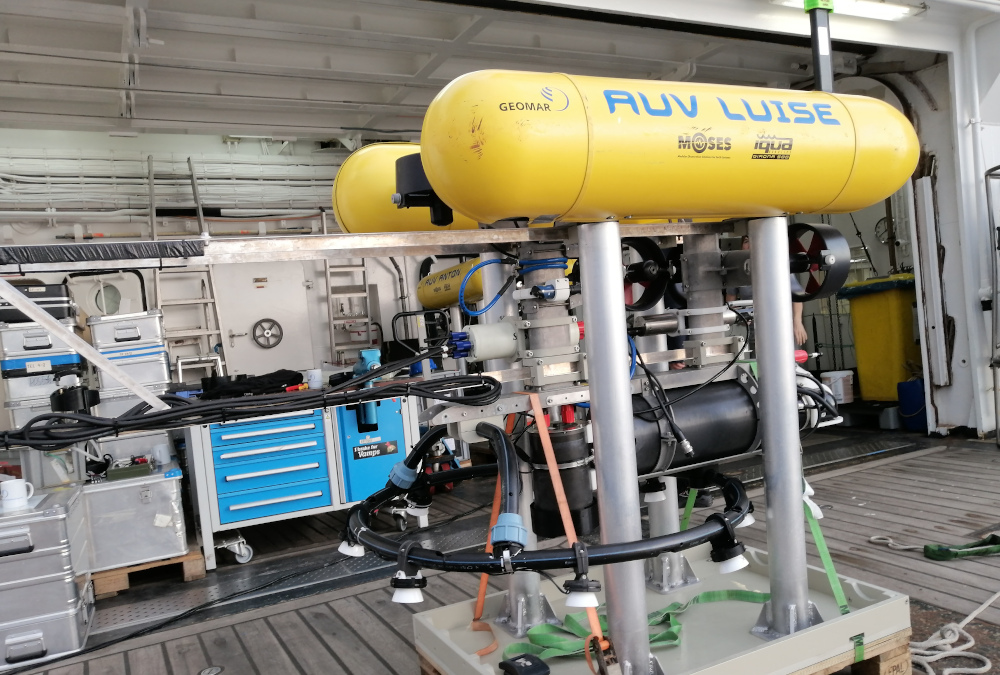}
		\includegraphics[height=3.8cm]{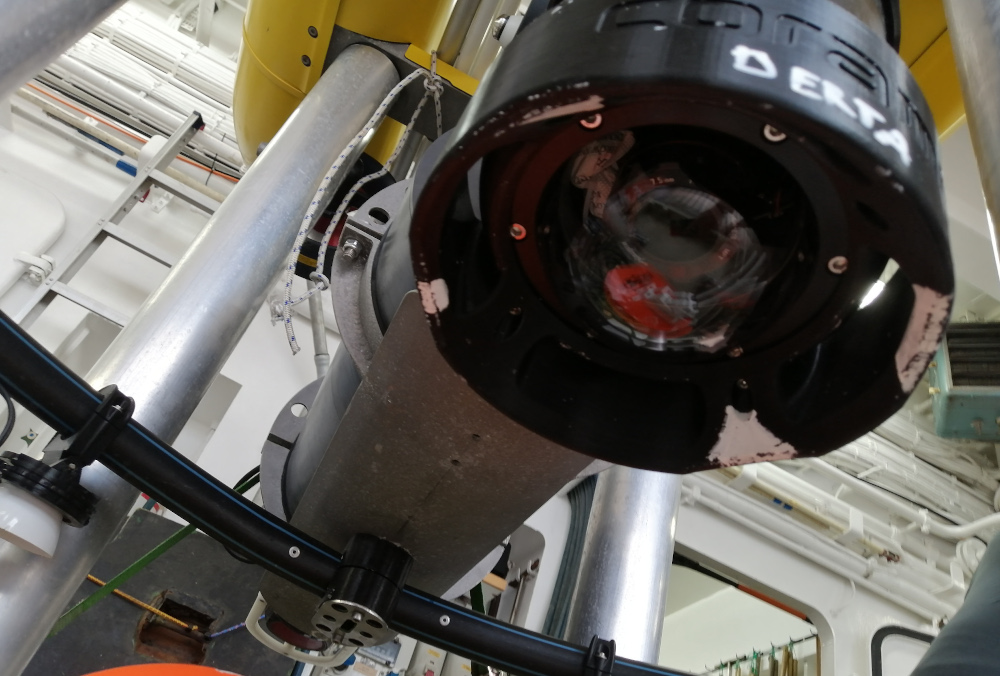}
		\includegraphics[height=3.8cm]{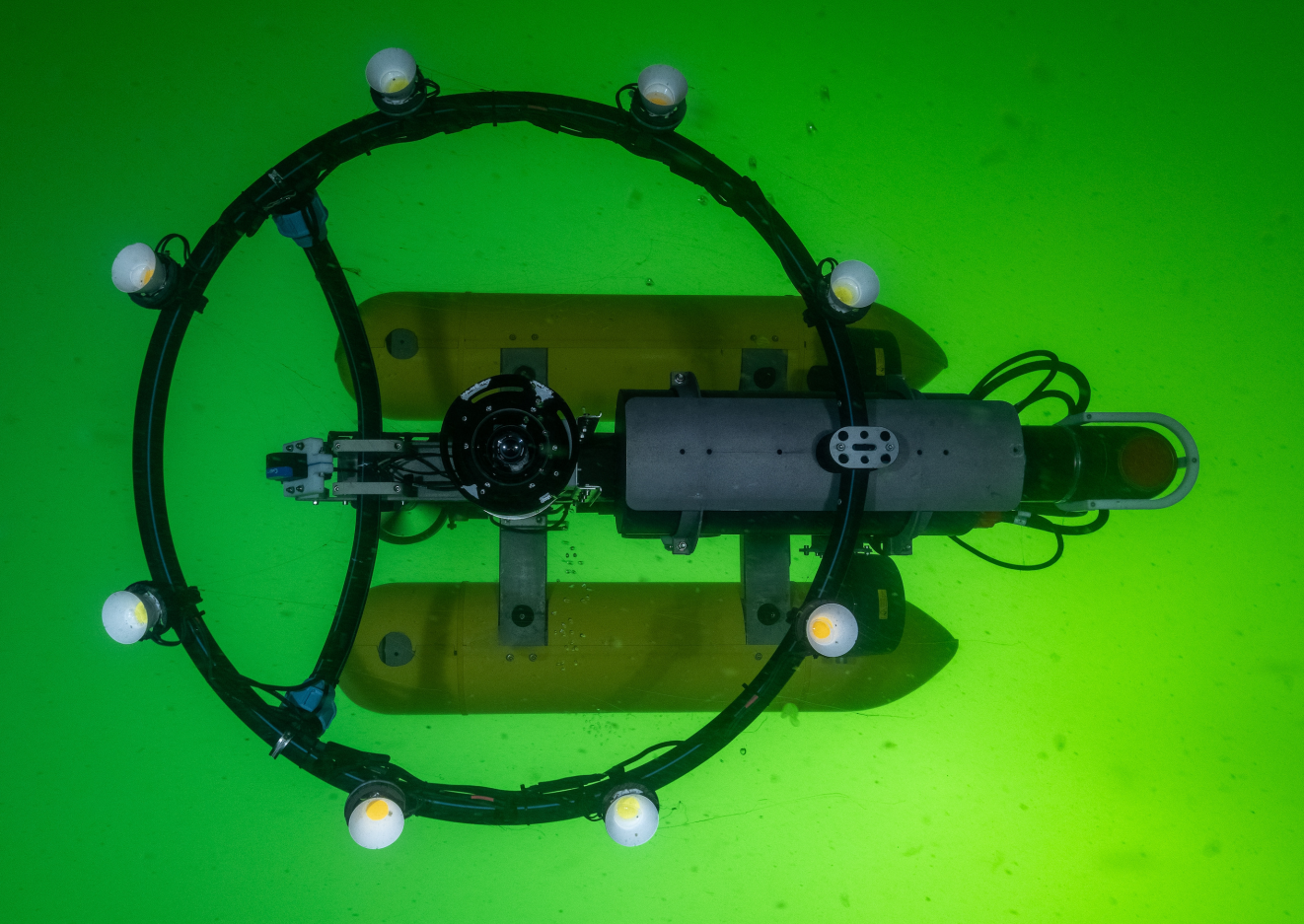}
	\end{center}
	\caption{From left to right: The Girona 500 AUV Anton and Luise; The CoraMo Mk \MakeUppercase{\romannumeral2} camera system; The LED light ring.}
	\label{fig:anton_and_camera}
\end{figure*}

The GEOMAR Helmholtz Centre for Ocean Research Kiel, Germany (GEOMAR) owns two Girona 500 AUVs named "Anton" and "Luise"\footnote{https://www.geomar.de/tlz/auv-autonome-unterwasserfahrzeuge/autonome-unterwasserfahrzeuge} (shown in Fig. \ref{fig:anton_and_camera}, left), developed by the Underwater Robotics Laboratory of the University of Girona, Spain \cite{ribas2011girona}.
The Girona 500 AUVs are specifically designed for slower speeds with hovering capabilities, enabling them to operate in a close distance to the seafloor and to achieving higher resolutions.
The AUV is equipped with several sensors for navigation, including the surface GPS, IMU and DVL for the inertial measurements and USBL for the absolute positioning.
Furthermore, they are designed to be reconfigurable for custom tasks and variety of sensor suites.
More comprehensive informations regarding the design and specifications of the Girona 500 AUVs can be found in \cite{ribas2011girona}.

On top of that basis, the GEOMAR's AUV Team has developed the CoraMo Mk \MakeUppercase{\romannumeral2} camera system \footnote{https://www.geomar.de/tlz/auv-autonome-unterwasserfahrzeuge/programme-und-projekte/coramo} \cite{hissmann2020rv}, as shown in Fig. \ref{fig:anton_and_camera}, right. 
The CoraMo Mk \MakeUppercase{\romannumeral2} camera system incorporates a high-performance machine vision camera (iDS UI-3000SE-C-HQ) featuring a Sony IMX253 1.1 CMOS sensor. 
The camera has a resolution of 4104 x 3006 pixels, and is equipped with a TAMRON fisheye lens. The field of view of the camera is about $88^\circ$.
Recent studies have shown that refraction effects can be effectively minimized in dome-port camera systems when the projection center of the camera is precisely aligned with the center of the dome port housing \cite{she2021refractive}.
Small forward and backward decentering of the lens inside the dome can be largely compensated by radial distortion parameters when perspective underwater calibration is performed.
Therefore, we have upgraded our camera system from a flat port to a $10cm$ diameter dome port, capable of withstanding water depths of up to 6000 meters.
Additionally, a $1m$ diameter LED ring with eight LEDs is mounted around the camera system to provide adequate illumination.

\subsection{Overview of the Visual Seafloor Mapping Pipeline}

\begin{figure}[!h]
	\centering
	\def\svgwidth{0.99\columnwidth}
	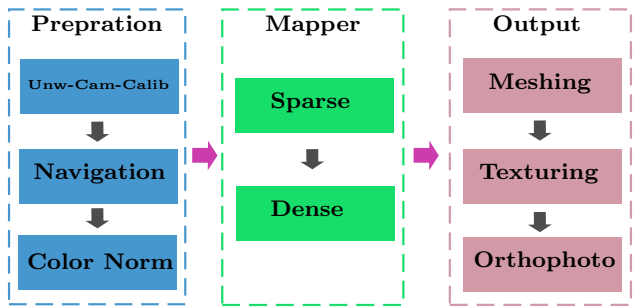
	\caption{Overview of the visual seafloor mapping pipeline.}
	\label{fig:pipeline_overview}
\end{figure}

The visual seafloor mapping pipeline consists of several key steps, as shown in Fig. \ref{fig:pipeline_overview}.
First, an underwater camera calibration is conducted to calibrate the dome-port camera system prior to the mission.
After the mission, the navigation data, obtained from a sensor fusion module, is extracted and used for geo-referencing the captured images.
Next, a color normalization approach, such as the one described in \cite{koser2021robustly}, is applied to the image sequence, removing deep-sea lighting effects.
Afterwards, a navigation-aided hierarchical SfM is performed to create a globally consistent sparse representation of the scene, followed by a chunk-based Multi-View Stereo (MVS) step to produce a densified 3D point cloud.
This process will be elaborated in Sect. \ref{sec:mapper}.
Finally, it is of interest for marine data science application to reconstruct a textured surface mesh and an orthophoto for geographic interpretation.

\subsection{Color Normalization}
\label{sec:color_norm}

Deep-sea AUV mapping tasks are often conducted at depths of hundreds to thousands of meters, artificial light sources are required to provide adequate illumination in the otherwise absolute darkness. However, introducing artificial lighting leads to different light and water effects in images compared to shallow water conditions \cite{song2022optical}, including pronounced light patterns and non-isotropic scattering light cones.
To obtain a 3D model which is free of illumination effects, various methods have been developed to remove artificial light patterns. 
For instance, Pizarro et al. \cite{pizarro2003toward} estimates an illumination image in log space by averaging frames to remove the illumination pattern.
Johnson-Roberson et al. \cite{johnson2017high} employs the gray-world algorithm to estimate gain and offset images based on mean and variance over clustered image sets. 
Additionally, Eustice et al. \cite{eustice2000image} applies contrast limited adaptive histogram equalization to homogenize the illumination in images.
Some physical model-based approaches not only remove light patterns but also restore true colors. 
For example, Bryson et al. \cite{bryson2016true} modifies the Jaffe-McGlamery image formation model and estimates its parameters from image correspondences to simultaneously remove light patterns and correct the colors of underwater images. 
Nakath et al. \cite{nakath2021situ} uses a Monte-Carlo based differentiable ray-tracing approach to estimate the light and water parameters by minimizing differences between simulated and real images, followed by estimating object colors under a similar rendering pipeline using the previously estimated parameters. 
However, these approaches either require prior knowledge of the lighting system or are computationally intensive and suitable only for small-scale applications. 
In this work, we employ a rather more practical approach presented in \cite{koser2021robustly}, which utilizes the assumption that the seafloor has a constant dominant color ($>$50$\%$ of pixels), and perform a statistics-based estimation of additive and multiplicative nuisances that avoid explicit parameters for light, camera, and the water. 
The normalization procedure involves subtracting the previously estimated additive scatter component and dividing by the multiplicative factor image.
The approach is implemented in CUDA and it runs in (near) real-time, therefore, it is more suitable for large-scale AUV mapping tasks.

\subsection{Underwater Camera Calibration}
\label{sec:unw_cam_calib}
\begin{figure}[!h]
	\includegraphics[width=0.32\columnwidth]{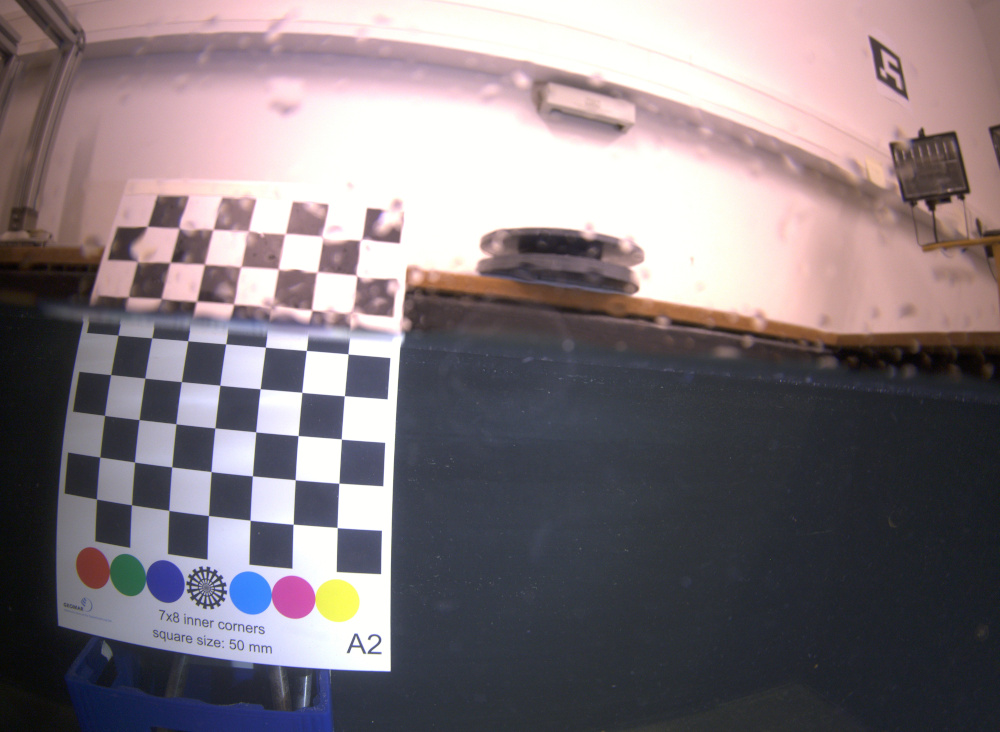}		
	\includegraphics[width=0.32\columnwidth]{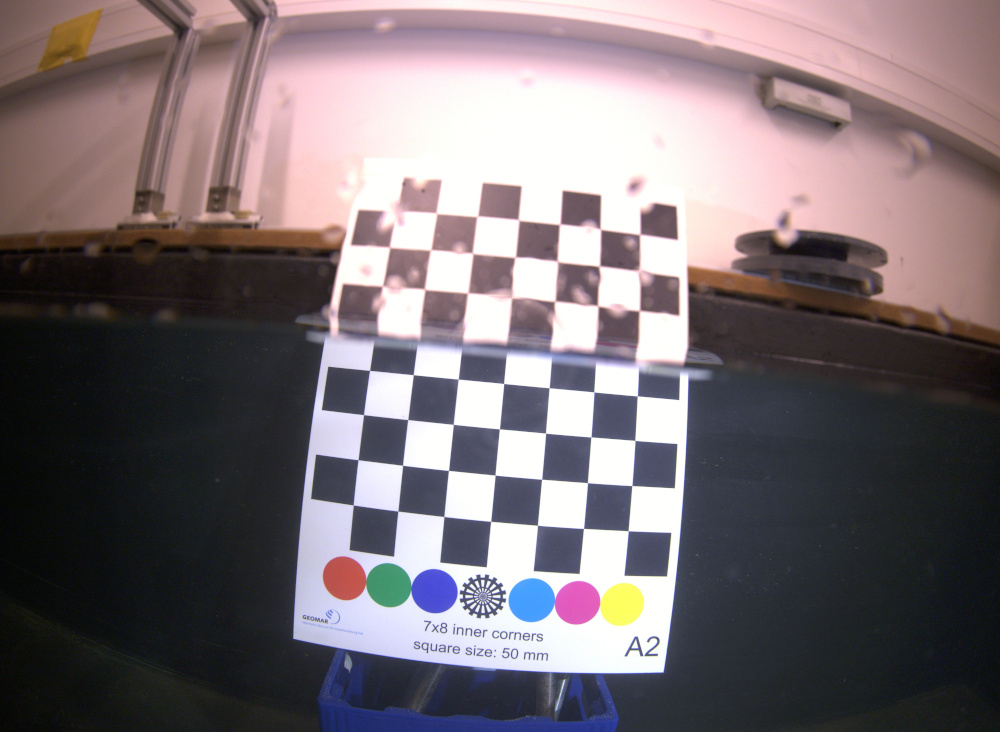}
	\includegraphics[width=0.32\columnwidth]{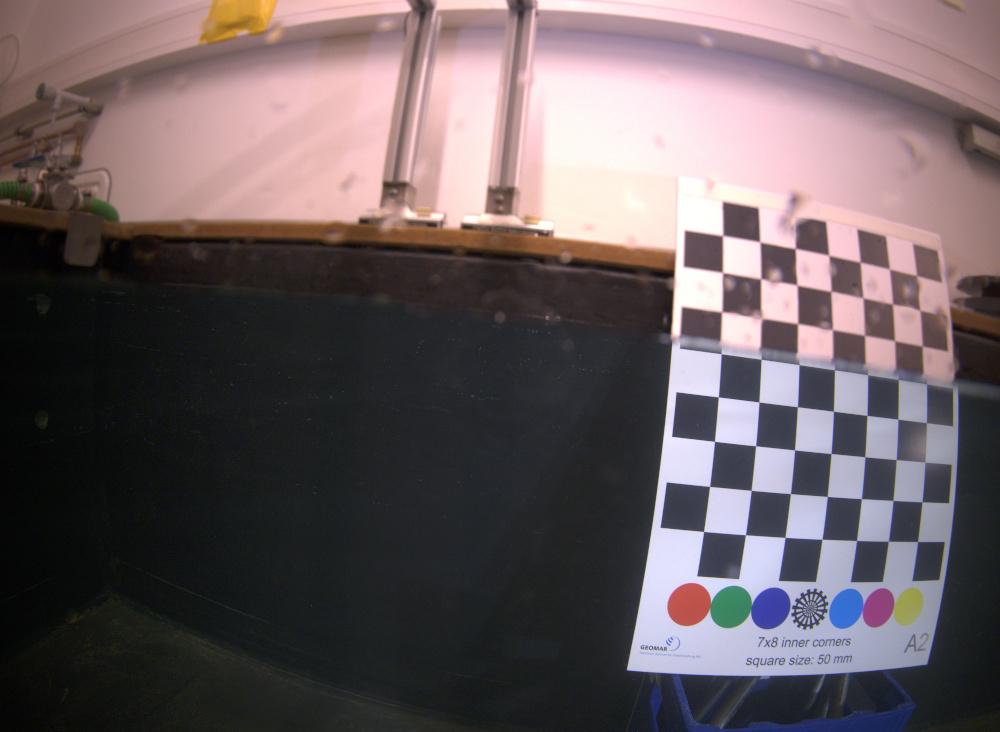}
	\caption{First stage of the underwater camera calibration: centering the camera in the dome port. Sample images have shown that the vertical edges of the calibration target are consistent above and below the water surface, which suggests that the refraction effect has been effectively avoided.}
	\label{fig:calib_air_water}
\end{figure}
We employ a two-staged calibration approach for the underwater camera systems.
In the first stage, we follow the procedure outlined in \cite{she2019adjustment} to align the projection center of the camera with the center of the dome port.
To achieve this, we remove the camera system from the AUV and submerge it halfway underwater in a water tank, ensuring that the camera observes the water surface in a parallel manner.
Subsequently, a calibration target (in our case, a checkerboard) is positioned in front of the camera, also submerged halfway underwater, as depicted in Fig. \ref{fig:calib_air_water}.
By visually examining the continuity of the vertical edges of the checkerboard at the water boundary, we can determine if the camera is properly centered.
The underlying principle is that if the camera is offset by a certain amount, refraction occurs, causing light rays from the submerged portion of the checkerboard to be refracted, while light rays from the above-water portion remain unaffected, resulting in a discontinuity in the vertical edges at the water surface.
Since visual examination is subjective and relies on human observers, multiple photos of the fully submerged checkerboard in water are captured to to estimate the decentering offset parameters.
This estimation can be done using the approach presented in \cite{she2021refractive}.
This process is repeated, and mechanical adjustments are made until the decentering offset parameters fall within the acceptable range of toleration.
Empirically, the values are typically set to be much less than 1$mm$ in the sideward direction and in the range of a millimeter in the forward-backward direction.

In the second stage of the calibration process, the centered dome port camera is treated as a normal perspective camera, and we perform a standard checkerboard-based calibration procedure as described in \cite{zhang2000flexible}, with the calibration target and the camera system submerged underwater.
At this stage, the remaining decentering offset can be compensated by adjusting the camera intrinsics.
Due to the large field of view of the AUV camera, we use the fisheye camera model for the calibration.

\section{Hierarchical Reconstruction with Weak-area Revisit}
\label{sec:mapper}

\begin{figure}[!h]
	\centering
	\def\svgwidth{0.99\columnwidth}
	\begingroup%
  \makeatletter%
  \providecommand\color[2][]{%
    \errmessage{(Inkscape) Color is used for the text in Inkscape, but the package 'color.sty' is not loaded}%
    \renewcommand\color[2][]{}%
  }%
  \providecommand\transparent[1]{%
    \errmessage{(Inkscape) Transparency is used (non-zero) for the text in Inkscape, but the package 'transparent.sty' is not loaded}%
    \renewcommand\transparent[1]{}%
  }%
  \providecommand\rotatebox[2]{#2}%
  \newcommand*\fsize{\dimexpr\f@size pt\relax}%
  \newcommand*\lineheight[1]{\fontsize{\fsize}{#1\fsize}\selectfont}%
  \ifx\svgwidth\undefined%
    \setlength{\unitlength}{239bp}%
    \ifx\svgscale\undefined%
      \relax%
    \else%
      \setlength{\unitlength}{\unitlength * \real{\svgscale}}%
    \fi%
  \else%
    \setlength{\unitlength}{\svgwidth}%
  \fi%
  \global\let\svgwidth\undefined%
  \global\let\svgscale\undefined%
  \makeatother%
  \begin{picture}(1,0.50209205)%
    \lineheight{1}%
    \setlength\tabcolsep{0pt}%
    \put(0,0){\includegraphics[width=\unitlength,page=1]{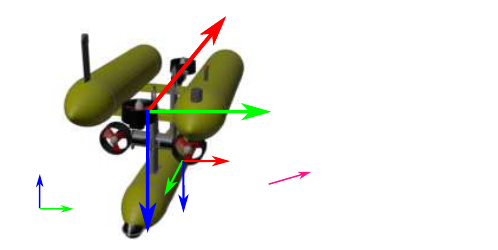}}%
    \put(0.30266706,0.4465584){\makebox(0,0)[lt]{\lineheight{1.25}\smash{\begin{tabular}[t]{l}$^p \mq T_w$\end{tabular}}}}%
    \put(0.03111368,0.02098704){\makebox(0,0)[lt]{\lineheight{1.25}\smash{\begin{tabular}[t]{l}$X$\end{tabular}}}}%
    \put(0.06906234,0.16222341){\makebox(0,0)[lt]{\lineheight{1.25}\smash{\begin{tabular}[t]{l}$Z$\end{tabular}}}}%
    \put(0.12824315,0.05192814){\makebox(0,0)[lt]{\lineheight{1.25}\smash{\begin{tabular}[t]{l}$Y$\\\end{tabular}}}}%
    \put(0.38153877,0.06367837){\makebox(0,0)[lt]{\lineheight{1.25}\smash{\begin{tabular}[t]{l}$^c \mq T_w$\\\end{tabular}}}}%
    \put(0.605512,0.04825632){\makebox(0,0)[lt]{\lineheight{1.25}\smash{\begin{tabular}[t]{l}Local Camera Frame\end{tabular}}}}%
    \put(0,0){\includegraphics[width=\unitlength,page=2]{coordinates.pdf}}%
  \end{picture}%
\endgroup%

	\caption{The notations of different coordinate systems. $^p \mq T_w$ denotes the prior coordinate frame, which represents the vehicle body and $^c \mq T_w$ represents the camera coordinate frame.}
	\label{fig:coord_frames}
\end{figure}

\textbf{Notations.}
We denote the absolute camera poses as $^c\mq T^i_w = \left[ ^c \mq R^i_w \;\mid\; ^c \q t^i_w \right]$, where $^c \mq R^i_w$ and $^c \q t^i_w$ are the rotation and the translation of the $i$-th image. And $^c \mq T^i_w$ transforms a 3D point in the world frame to the local camera frame.
Then, we introduce the prior coordinate frame $p$ to represent the vehicle body of the $i$-th image which is denoted as $^p\mq T^i_w = \left[ ^p \mq R^i_w \;\mid\; ^p \q t^i_w \right]$.
In addition, we introduce $^p \mq T_c$ as the relative transformation from the camera frame to the prior frame.
A simple illustration of the coordinate systems and their relations is shown in Fig. \ref{fig:coord_frames}.

\begin{figure*}[ht!]
	\centering
	\def\svgwidth{0.99\textwidth}
	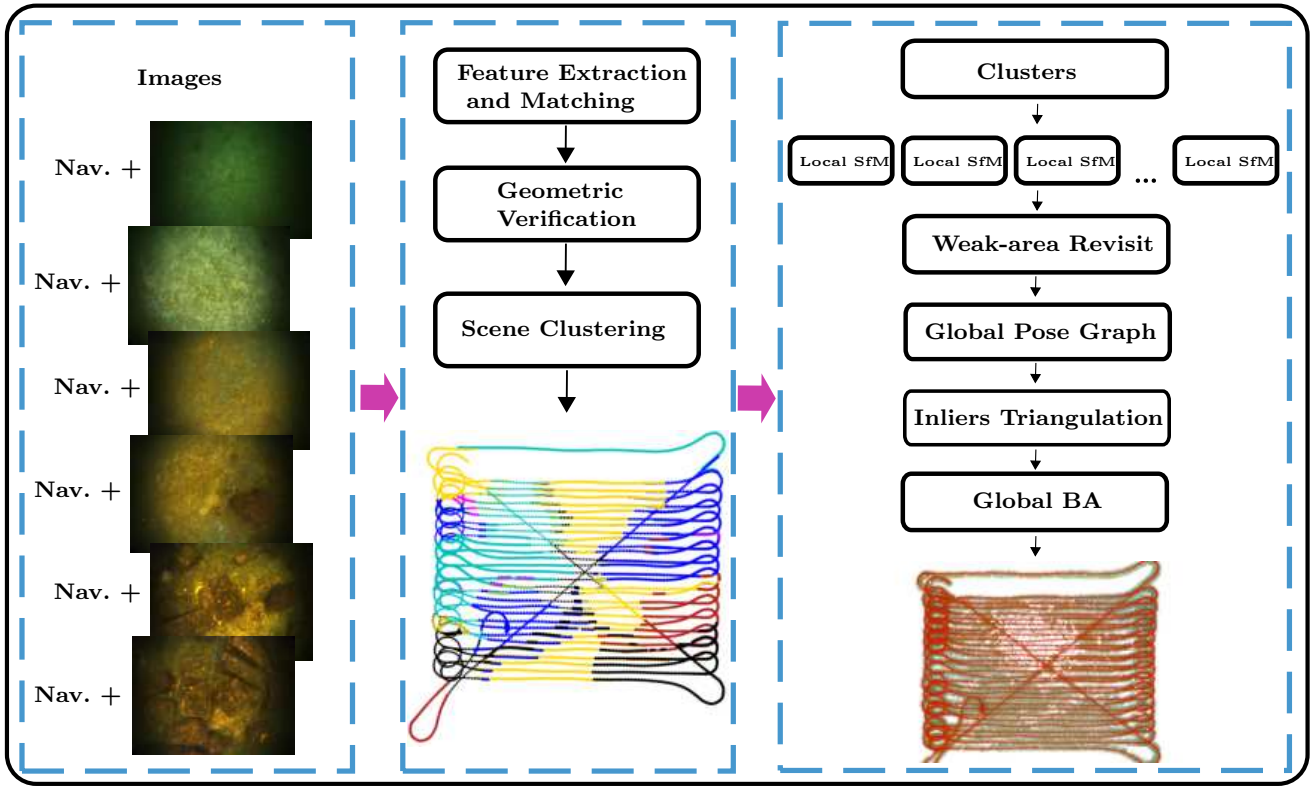
	\caption{The overview of the navigation-aided hierarchical SfM approach.}
	\label{fig:sfm_pipeline}
\end{figure*}

Having obtained the color normalized images, geo-referenced navigation data, and calibrated camera intrinsics, we now introduce our proposed navigation-aided hierarchical SfM approach for achieving efficient and large-scale visual seafloor reconstruction.
The overview of the approach is illustrated in Fig. \ref{fig:sfm_pipeline}.

Similar to previous SfM work from COLMAP \cite{schonberger2016structure}, we first extract SIFT features and their descriptors across the entire image collection. 
Then, we perform spatial matching for each image within localized regions.
This is possible by leveraging the absolute positions of each image from the prior poses.
Afterwards, a geometric verification step is performed to compute the two-view geometries for all possible image pairs and a view graph that encodes the spatial distribution of the features is constructed.
Note that in this process, outliers which do not agree with the estimated two-view geometries are eliminated by RANSAC.
In order to effectively handle large-scale datasets and to mitigate the accumulation drift, we employ a divide-and-conquer strategy similar to \cite{chen2020graph,zhu2017parallel,bhowmick2015divide}. 
Using the view graph computed from the previous step, where images are represented as graph nodes and edges denote the number of inliers between image pairs, we apply a normalized graph-cut algorithm \cite{dhillon2007weighted} to partition the scene into multiple overlapping clusters.

\textbf{Local SfM.} 
Subsequently, we perform (incremental) SfM on each local cluster concurrently. 
Traditional SfM yields reconstructions with arbitrary scales, necessitating the estimation of a similarity transformation between sub-reconstructions that share overlapped images.
However, relying on a few connected images to estimate the similarity transformation is not robust, especially when the view graph is sparse.
This limitation frequently results in inconsistencies and gaps between sub-reconstructions.
To overcome this, we incorporate prior poses into the local SfM process, ensuring that the resulting reconstruction adheres to the same coordinate system and scale \cite{noguchi2022wide,Johnson_2010-3DrecAUV}.
When reconstructing the initial image pair, we begin by estimating the two-view geometry using the inlier matches, and obtain the relative rotation and translation between the image pair, with the translation vector normalized to unit length.
We then set the first camera pose directly as its prior pose.
Next, we scale the translation vector by the length obtained from the prior poses.
Subsequently, we concatenate the relative pose with the first camera pose, yielding the second camera pose.
By doing this, the resulting reconstruction is under the same coordinate system as the prior coordinate system.
It is important to note that, the navigation data is provided in the prior coordinate frame $^p \mq T_w$.
However, to use this information in the camera frame, we must transform it by $^{c^{\prime}} \mq T_w = (^p \mq T_c)^{-1} \cdot \; ^p \mq T_w$.
In our specific setup, the camera is positioned underneath the robot, and is rotated by $90^\circ$ around the $Z$-axis, as depicted in Fig. \ref{fig:anton_and_camera} and Fig. \ref{fig:coord_frames}.
Then, the offset between the camera body and the vehicle body is determined based on measurements taken from the CAD drawing.
We acknowledge that the actual values may deviate from the initial assumptions due to real-world variations.
However, these values are used solely as an initialization and will be optimized during the subsequent bundle adjustment process.

Once the initial reconstruction is established, we proceed with bundle adjustment to refine and optimize the registered images along with the corresponding 3D structures.
In this process, we supervise the optimization to mitigate drift by adding a pose prior term that penalizes the absolute difference between the optimized camera poses and their prior poses.
In essence, the cost function being optimized is as follows:
\begin{multline}
	C = \sum_i\sum_j\Vert \d x^j_i - \pi(\d X_i, \; \mathcal{K}, \; ^c\mq T^j_w)   \Vert^2 + \\ \sum_j \Vert d\left( ^c \mq T^j_w, \;^{c^{\prime}}\mq T^j_w  \right) \Vert^2
	\label{eq:ba_cost}
\end{multline}
$\pi(\cdot)$ denotes the forward projection function that projects the $i$th 3D point $\d X_i$ onto the $j$th image.
$\d x^j_i$ denotes the observation of this 3D point in the $j$th image. 
$\mathcal{K}$ represents the intrinsic parameters of the camera.
We then define the 6-dimensional residual vector $d\left( \mq T_i, \mq T_j \right)$ as the absolute difference between two transformations:
\begin{equation}
	d\left( \mq T_i, \mq T_j \right) = \begin{bmatrix}
	\rho_{r} \cdot 2 \cdot vec(^j\d q_i) \\ \rho_t \cdot ^j\d t_i
	\end{bmatrix}
\end{equation}
where $vec(^j\d q_i)$ returns the real part of the quaternion representation of the rotation from frame $i$ to frame $j$.
$\rho_{r} \in \mathbb{R}^{3\times3}$ and $\rho_{t} \in \mathbb{R}^{3\times3}$ are the weighting factors, which can be obtained by taking the inverse of the covariance matrix associated with the prior poses.
We follow the procedure in \cite{schonberger2016structure} to incrementally reconstruct the remaining images within the cluster.

\textbf{Weak-area Revisit.}
\begin{figure}[!h]
	\centering
	\def\svgwidth{0.9\columnwidth}
	\begingroup%
  \makeatletter%
  \providecommand\color[2][]{%
    \errmessage{(Inkscape) Color is used for the text in Inkscape, but the package 'color.sty' is not loaded}%
    \renewcommand\color[2][]{}%
  }%
  \providecommand\transparent[1]{%
    \errmessage{(Inkscape) Transparency is used (non-zero) for the text in Inkscape, but the package 'transparent.sty' is not loaded}%
    \renewcommand\transparent[1]{}%
  }%
  \providecommand\rotatebox[2]{#2}%
  \newcommand*\fsize{\dimexpr\f@size pt\relax}%
  \newcommand*\lineheight[1]{\fontsize{\fsize}{#1\fsize}\selectfont}%
  \ifx\svgwidth\undefined%
    \setlength{\unitlength}{239bp}%
    \ifx\svgscale\undefined%
      \relax%
    \else%
      \setlength{\unitlength}{\unitlength * \real{\svgscale}}%
    \fi%
  \else%
    \setlength{\unitlength}{\svgwidth}%
  \fi%
  \global\let\svgwidth\undefined%
  \global\let\svgscale\undefined%
  \makeatother%
  \begin{picture}(1,1.48535565)%
    \lineheight{1}%
    \setlength\tabcolsep{0pt}%
    \put(0.02610154,0.77305147){\makebox(0,0)[lt]{\lineheight{1.25}\smash{\begin{tabular}[t]{l}\textbf{Weak-area Detected}\end{tabular}}}}%
    \put(0,0){\includegraphics[width=\unitlength,page=1]{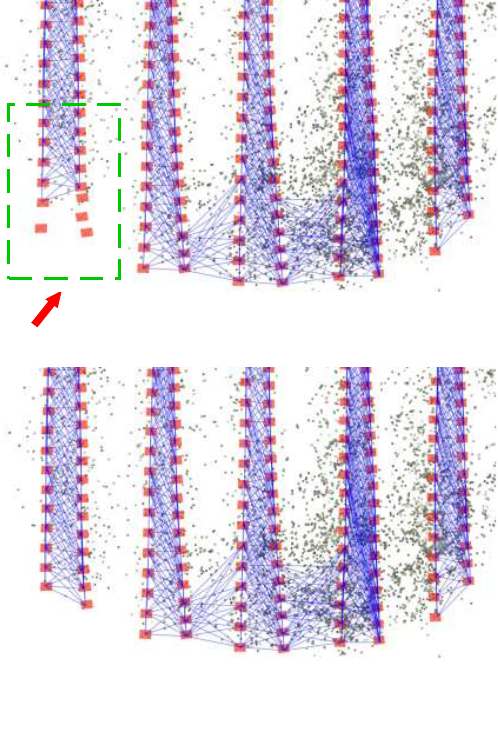}}%
    \put(0.02610155,0.03246697){\makebox(0,0)[lt]{\lineheight{1.25}\smash{\begin{tabular}[t]{l}\textbf{Weak-area Revisited}\end{tabular}}}}%
    \put(0,0){\includegraphics[width=\unitlength,page=2]{wr_revisit.pdf}}%
  \end{picture}%
\endgroup%

	\caption{Weak-area revisit. \textbf{Top}: Weak-area detection; \textbf{Bottom}: Weak-area revisited. The blue lines indicate the connected image pairs in the view graph.}
	\label{fig:wr_revisit}
\end{figure}
In the previous clustered local SfM reconstruction process, there may be images that are not reconstructed within any of the clusters, or image pairs that exhibit a significant number of inlier feature matches but lack shared reconstructed 3D points (see Fig. \ref{fig:wr_revisit}, top).
These situation can arise due to the separation of images into different clusters during the scene clustering process or the inherent difficulty in reconstructing certain images.
Some of the image pairs are even potential loop-closing candidates however not connected and aligned.

To address these issues, previous works have employed a graph-expansion step during the scene clustering phase to increase the number of overlapping images \cite{chen2020graph,zhu2017parallel}. 
However, this technique may not be effective for images that are too challenging to be reconstructed.
In contrast, we propose an alternative approach.
Firstly, we identify the weakly reconstructed areas, which consist of images that remain unreconstructed within any of the clusters and weakly reconstructed image pairs.
Next, we construct new clusters centered around these areas, including the target area and its surrounding images.
Afterwards, we can perform parallel local SfM on each of them.
Since the new clusters are smaller in size, re-performing local SfM on these limited sets of images does not significantly increase the overall computation time.
However, it helps increasing the connectivity between sub-reconstructions and allows for multiple attempts to reconstruct challenging images, which is beneficial for the subsequent global pose graph optimization step.
To identify the weak image pairs, we maintain an upgraded view-graph where we store the total number of common visible 3D points $N_p$ for each image pair.
We then compare this information with the original view-graph obtained from feature matching, where we store the number of feature matches ($N_m$) for each image pair.
Image pairs that satisfy $N_p < 0.2 \cdot N_m$ and $N_m > \mu$ are considered as weak image pairs and are revisited for further reconstruction.
Here, $\mu$ is a hyperparameter that determines whether an image pair contains a sufficient number of feature matches to be considered for reconstruction.
It can be seen in Fig. \ref{fig:wr_revisit}, bottom, that the weak-area is revisited and these challenging images can also be well reconstructed.

\textbf{Global Pose Graph Optimization.}
We align different sub-reconstructions by optimizing a joint global pose graph, where vertices represent camera poses and edges represent relative transformations between two cameras.
The pose graph contains three types of constraints: relative pose constraints, absolute pose constraints, and smooth motion constraints.

To establish the relative pose constraints, we collect all available image pairs within each sub-reconstruction that share common 3D points. 
We compute their relative poses and utilize them as edge measurements within the pose graph. 
This includes both sequential edges and loop-closing edges.
For each vertex in the graph, we impose a weak absolute pose constraint using prior pose measurements. This constraint penalizes vertices if they deviate significantly from the prior poses, helping to maintain consistency with the initial pose estimates.
Furthermore, for completeness reasons, we introduce a third local smooth motion constraint, which ensures that isolated vertices adhere to a plausible and coherent motion pattern within the reconstructed scene.
This enable us to handle vertices that lack any relative constraints, such as images not reconstructed within any clusters, even after revisiting them.

The entire graph of camera poses is then optimized by minimizing the following cost function:
\begin{multline}
	C = \rho_{rel} \sum_{(i, j) \in \mathcal{R}} \Vert d\left( ^j \mq T_i, \;^j \hat{\mq T}_i  \right) \Vert^2 + \\ \rho_{abs} \sum_{j \in \mathcal{V}} \Vert d\left( ^c \mq T^j_w, \;^{c^{\prime}}\mq T^j_w  \right) \Vert^2 +
	\rho_{sm} \sum_{i \in \mathcal{S}} \Vert d\left( ^i \mq T_{i-1}, \;^{i+1} \mq T_i  \right) \Vert^2
\end{multline}
where $\mathcal{R}$ is the set of all relative edges, $\mathcal{V}$ is the set of all vertices and $\mathcal{S}$ is the set of vertices that have no edges connected to them.
Again, the distance function $d(\cdot)$ measures the dissimilarity between the two transformations.
The smooth motion term penalizes the dissimilarity between the relative transformation from vertex $i-1$ to $i$ and the relative transformation from vertex $i$ to $i+1$, effectively encouraging constant velocity.
$\rho_{abs}$, $\rho_{rel}$ and $\rho_{sm}$ are the scalar multipliers to balance the influence of different constraints in the optimization process.

\textbf{Triangulation and Bundle Adjustment.} 
We gather all feature tracks available from each sub-reconstruction and merge them based on their feature matching graph.
The local SfM process incorporates robust outlier rejection mechanisms, such as RANSAC, in several reconstruction steps.
Therefore, we can confidently consider these feature tracks as inliers.
Next, we conduct re-triangulation on the combined feature tracks using the globally optimized camera poses obtained from the previous step to create a comprehensive scene structure.
To further refine the reconstruction, we perform a global bundle adjustment, similar to the approach described in \cite{schonberger2016structure} to minimize the global reprojection error, ensuring that overall reconstruction is of high quality and accuracy.

\textbf{Multi-View Stereo.} 
Once a sparse reconstruction is obtained, the registered images are processed using the open-source package OpenMVS \cite{openmvs2020} to generate a dense representation of the scene through Multi-View Stereo (MVS).
Due to the computational limitations of processing a large number of images, we again employ a divide-and-conquer strategy.
We divide the input reconstruction into several smaller subsets or "chunks" and each chunk is then processed independently using the MVS algorithm.
Note that in this step, we substitute the original images by the color normalized images to compensate for the varying lighting effect in the final output mesh.
\begin{figure*}[!h]
	\centering
	\subfloat[\textit{Easy}]{
		\includegraphics[width=0.2\textwidth]{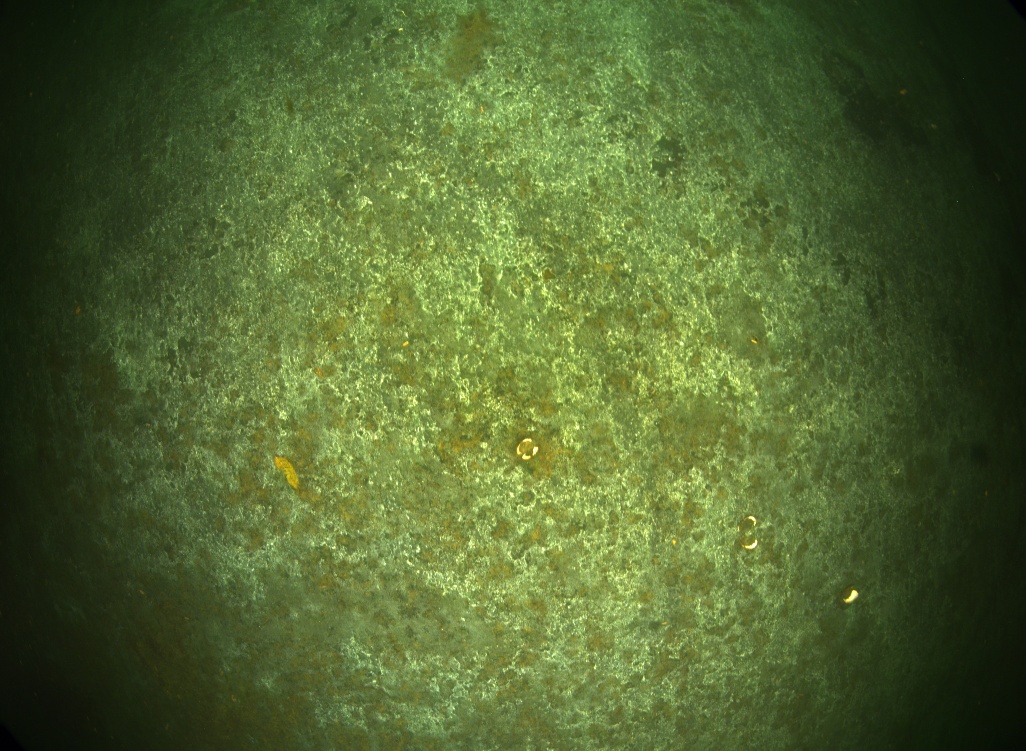}
		\includegraphics[width=0.2\textwidth]{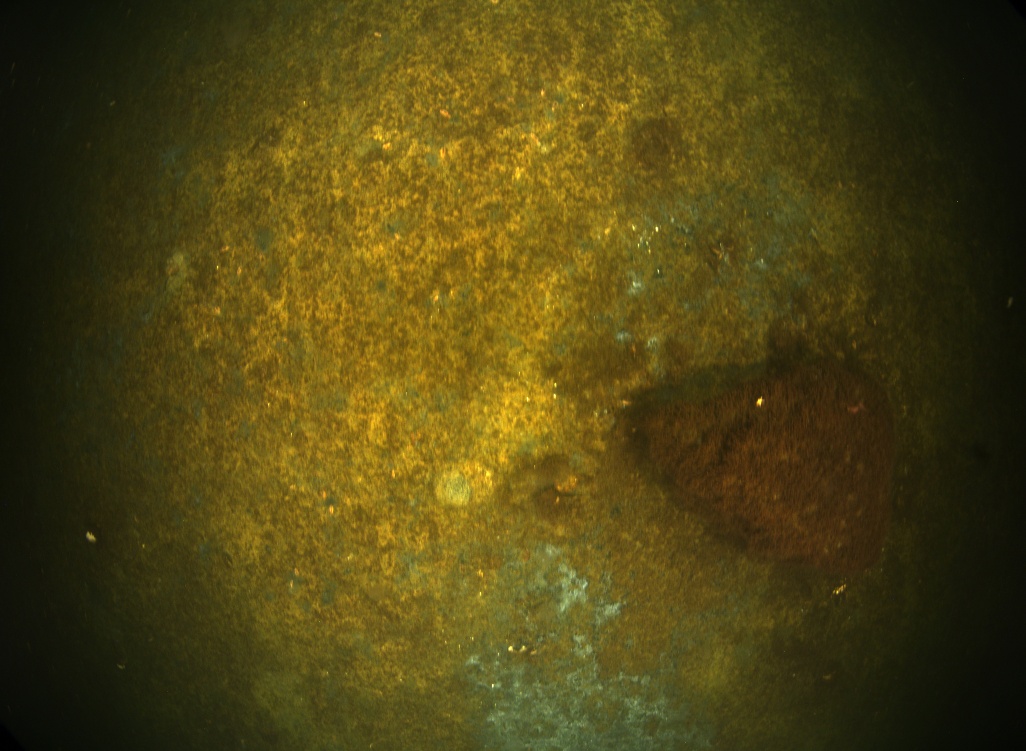}						 \includegraphics[width=0.2\textwidth]{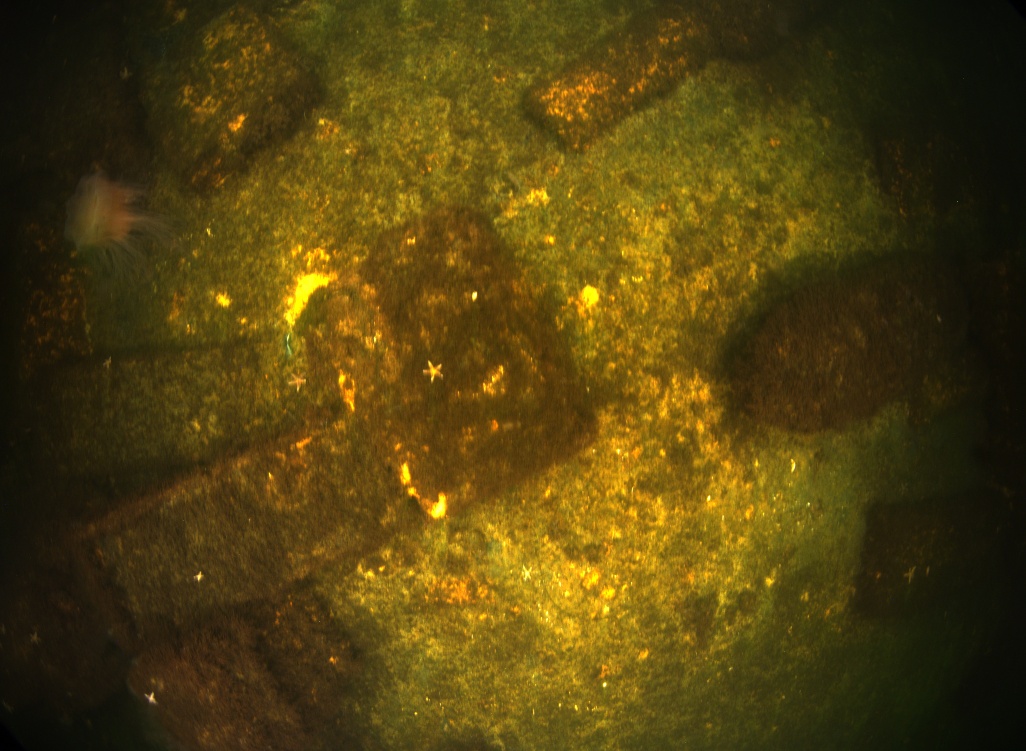}			\includegraphics[width=0.2\textwidth]{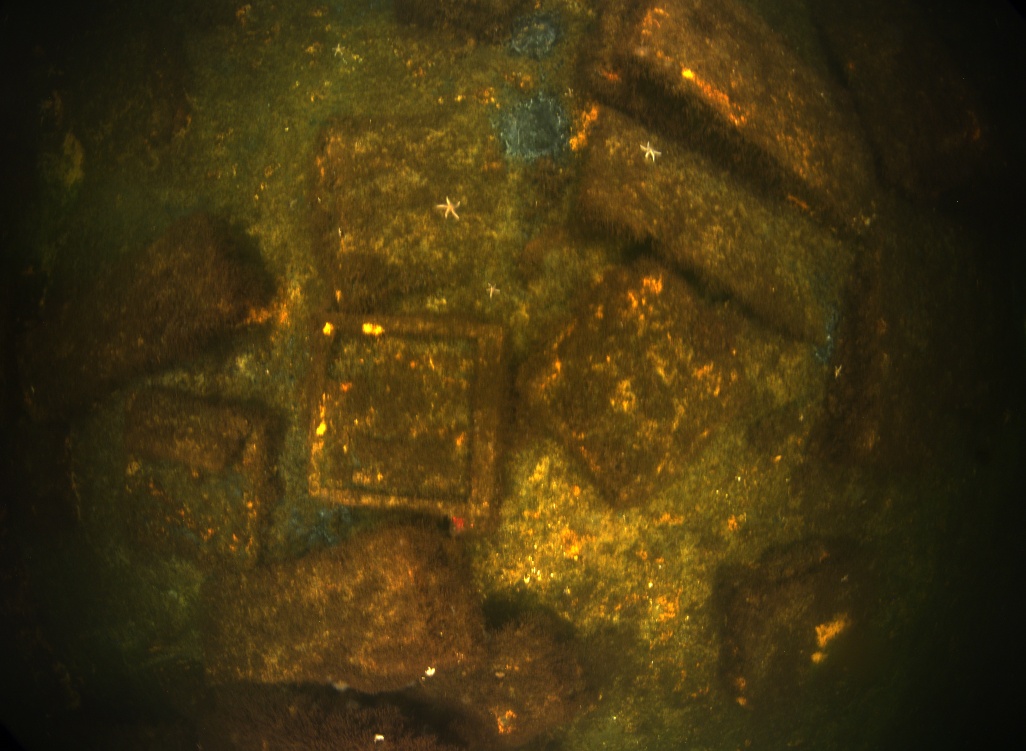}
		\includegraphics[width=0.18\textwidth]{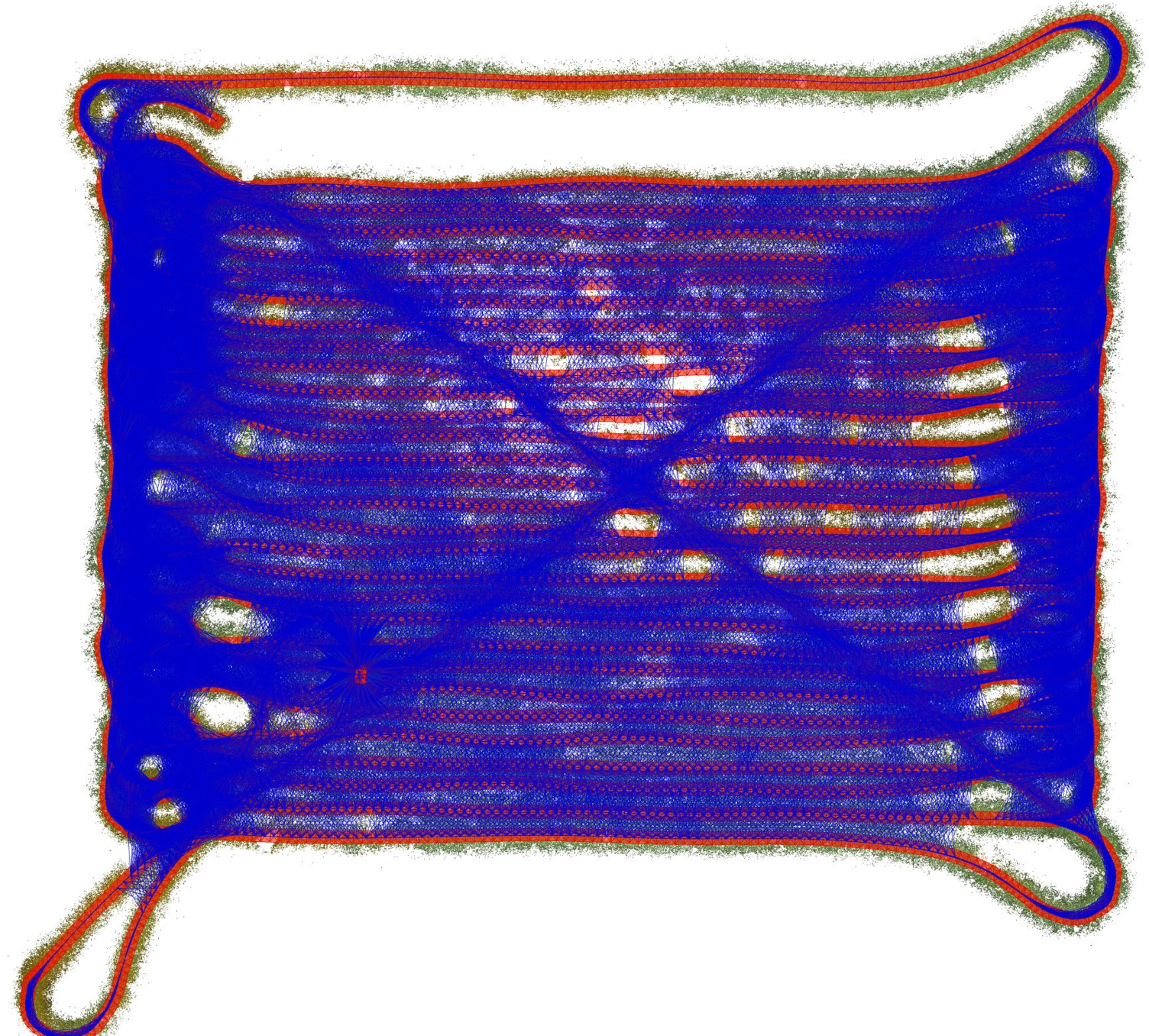}
	}\\
	\subfloat[\textit{Medium1}]{		
		\includegraphics[width=0.2\textwidth]{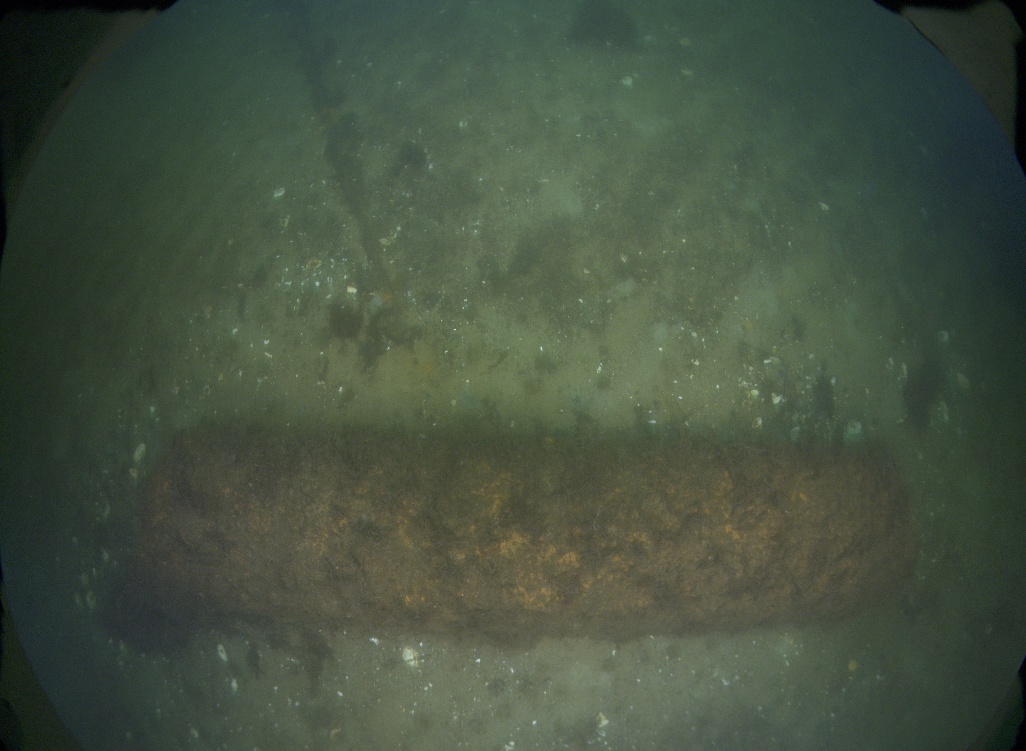}
		\includegraphics[width=0.2\textwidth]{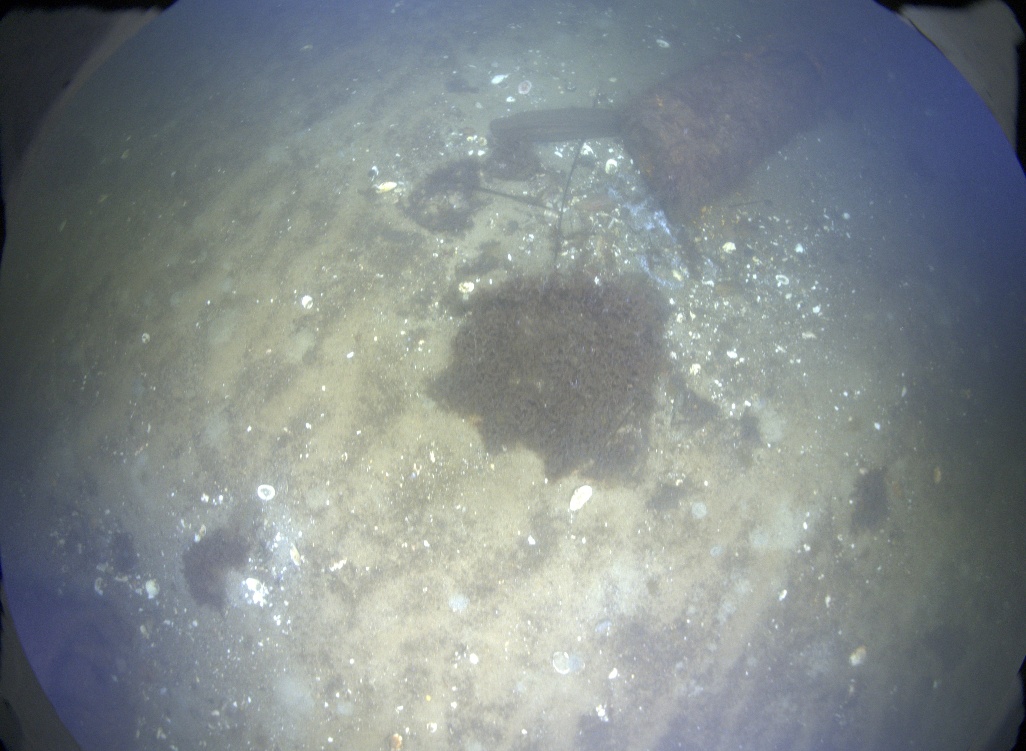}				\includegraphics[width=0.2\textwidth]{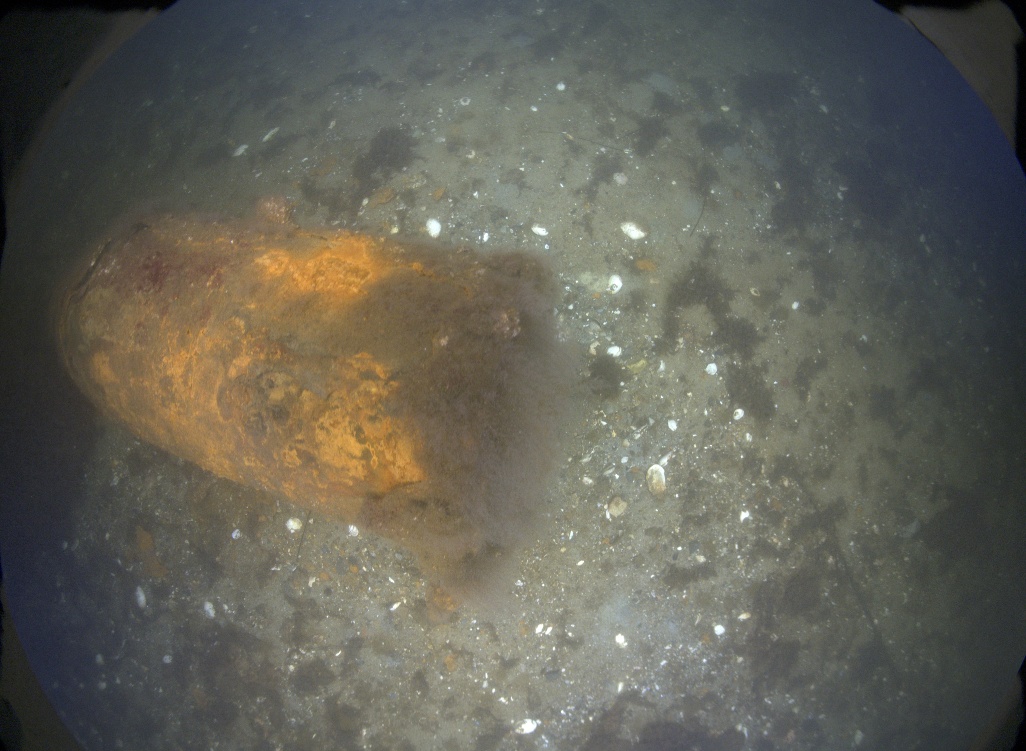}			\includegraphics[width=0.2\textwidth]{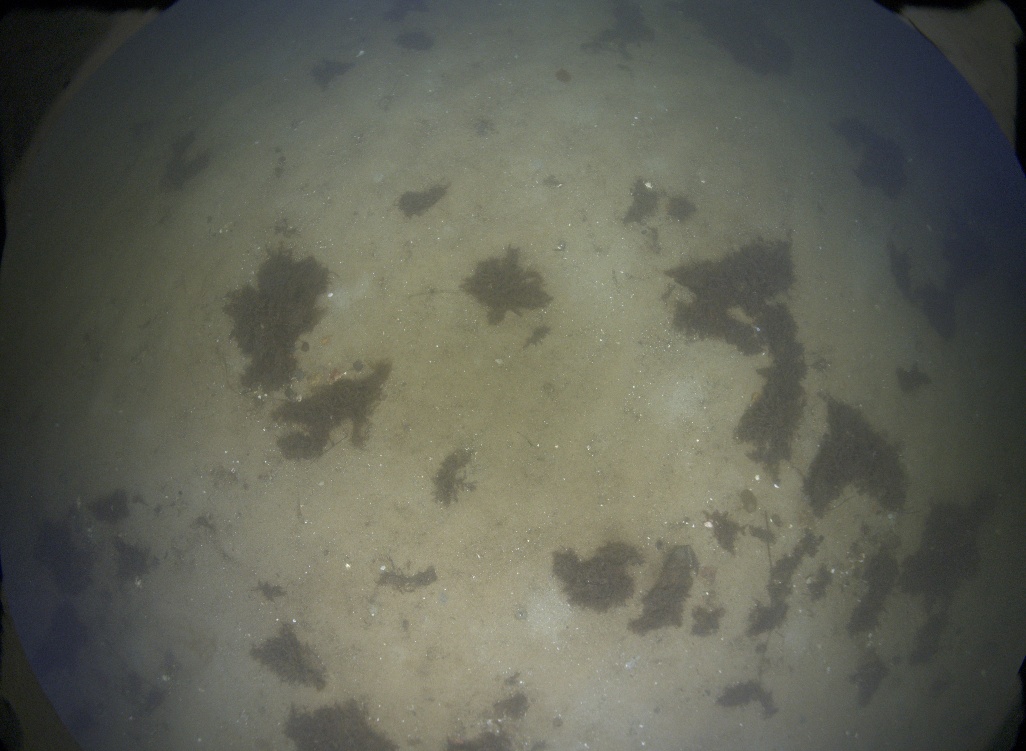}
		\includegraphics[width=0.18\textwidth]{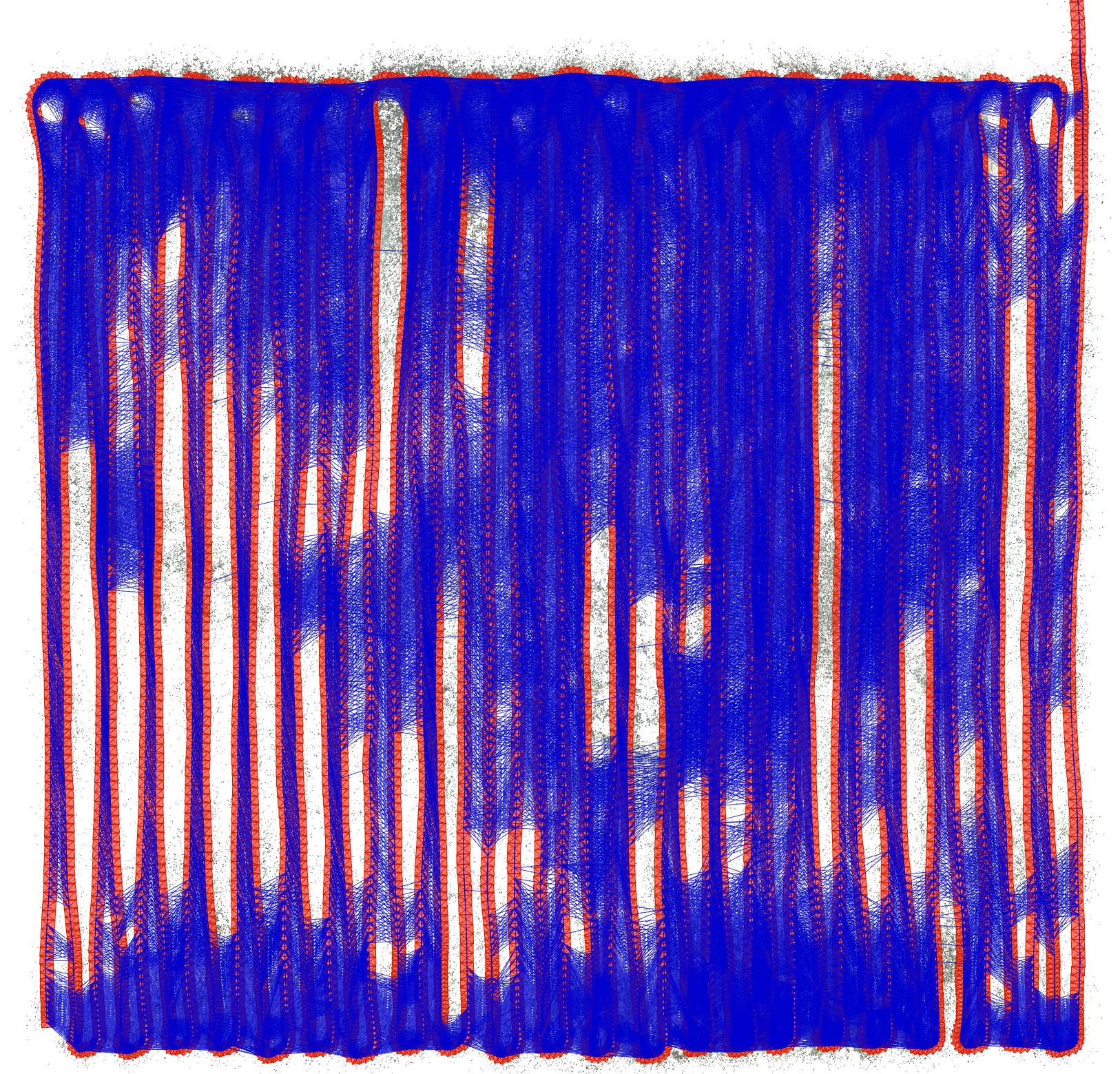}
	}\\
	\subfloat[\textit{Medium2}]{
		\includegraphics[width=0.2\textwidth]{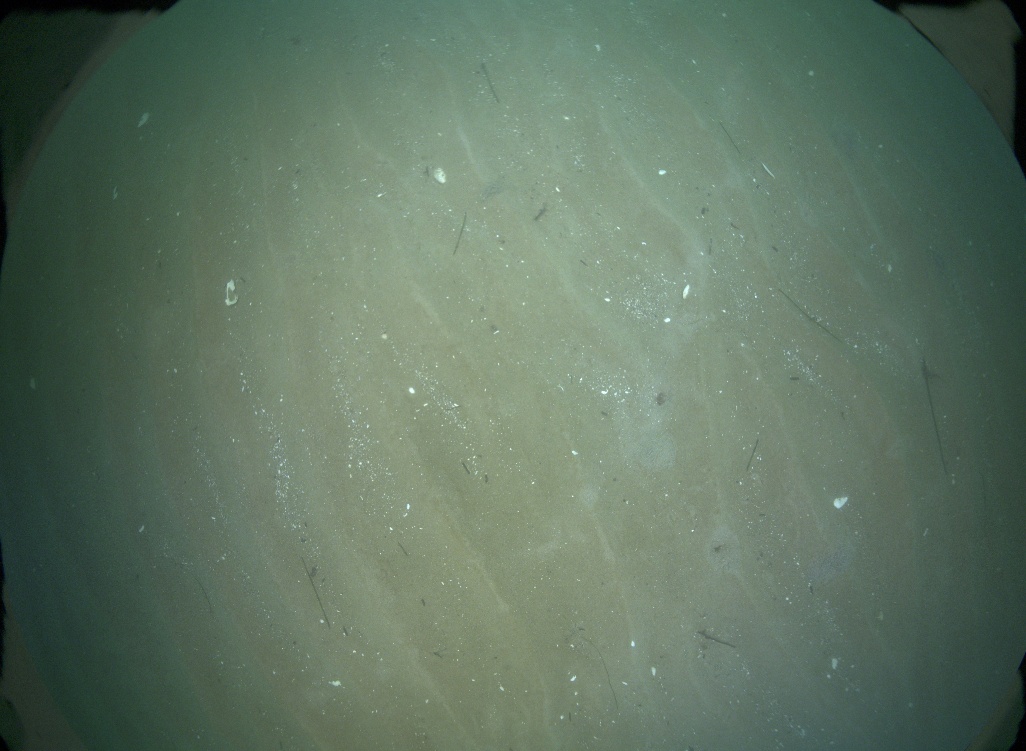}
		\includegraphics[width=0.2\textwidth]{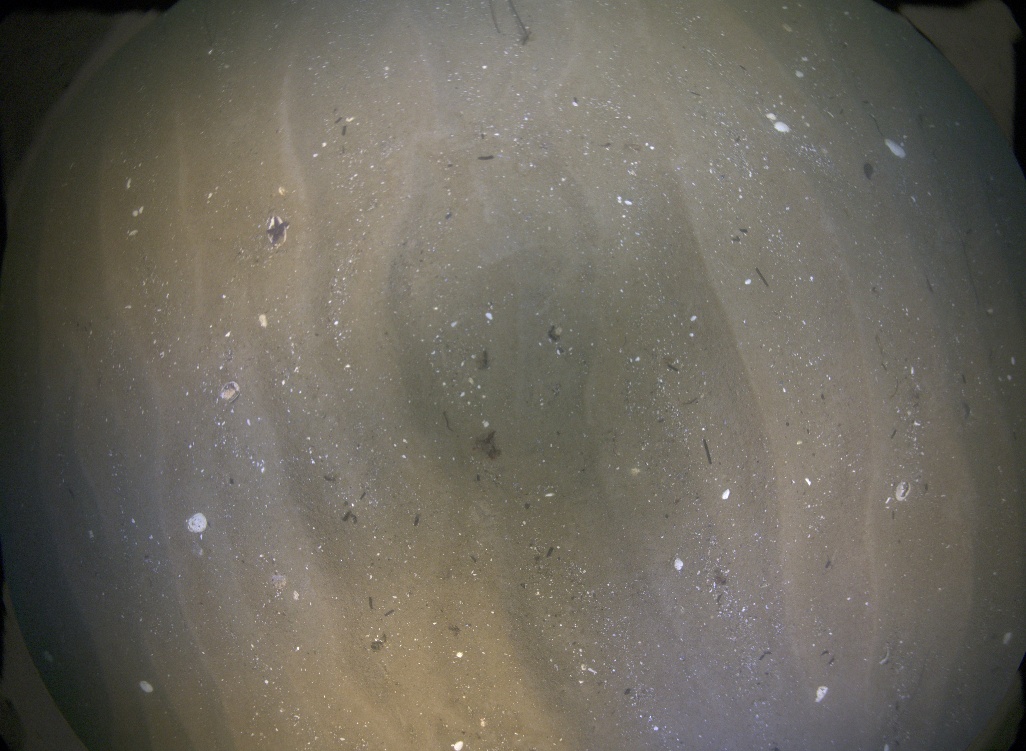}				\includegraphics[width=0.2\textwidth]{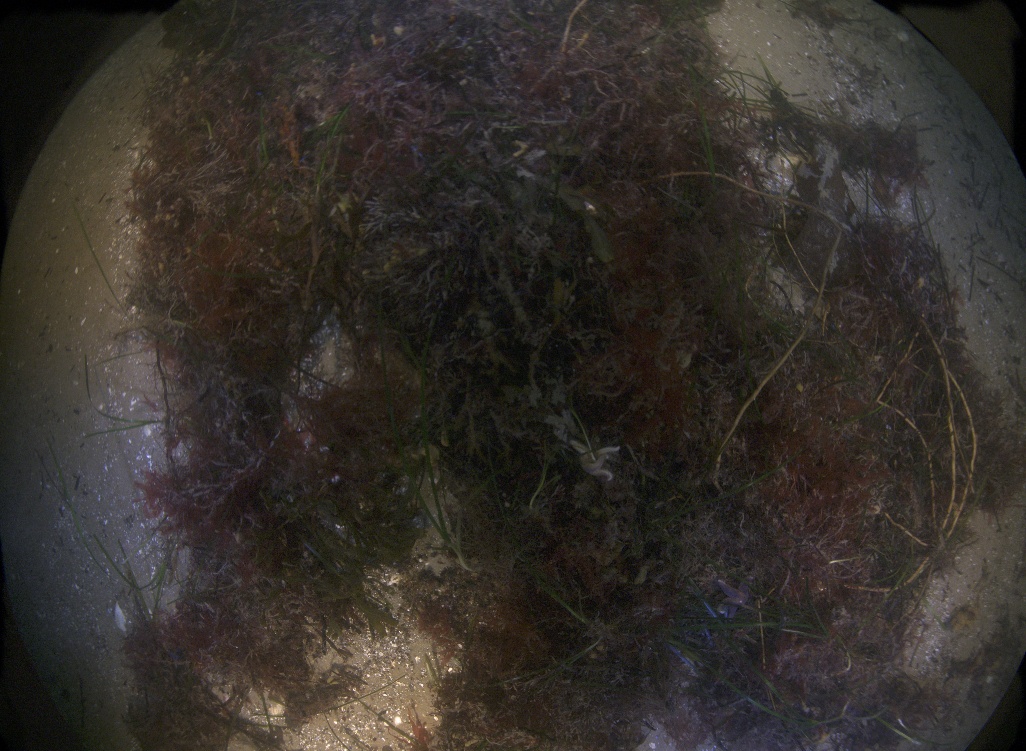}			\includegraphics[width=0.2\textwidth]{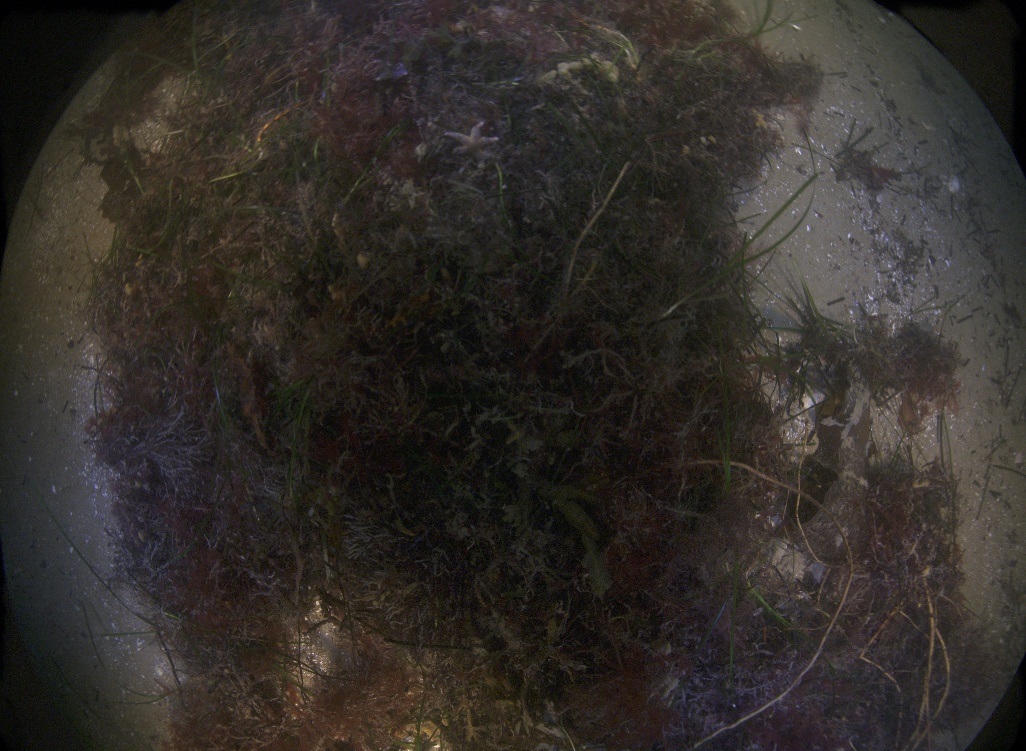}
		\includegraphics[width=0.18\textwidth]{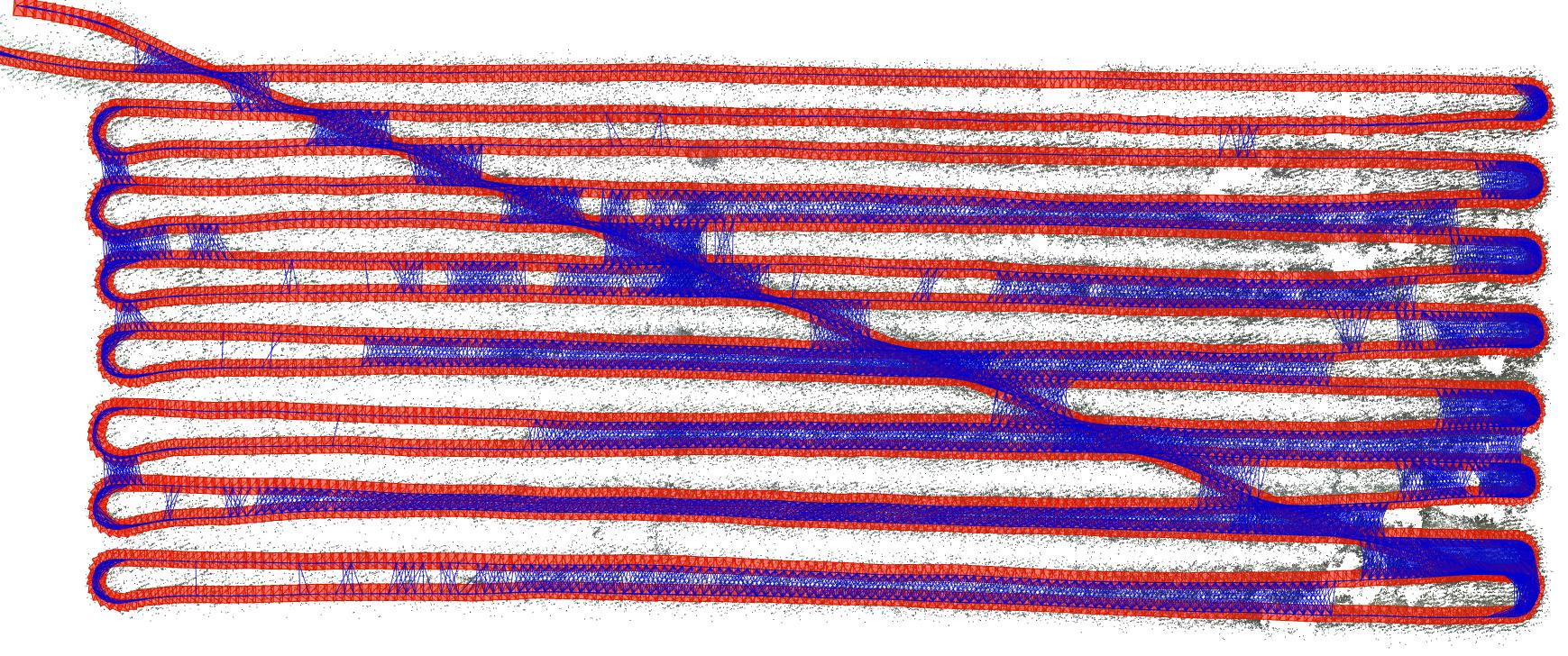}
	}\\
	\subfloat[\textit{Hard1}]{
		\includegraphics[width=0.2\textwidth]{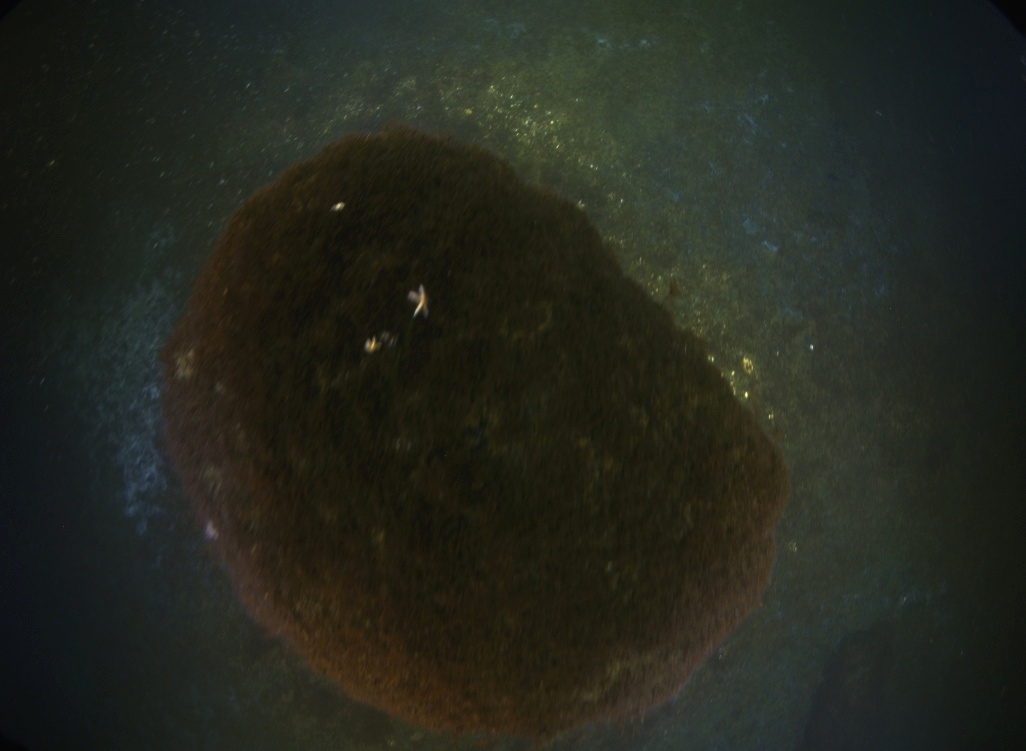}
		\includegraphics[width=0.2\textwidth]{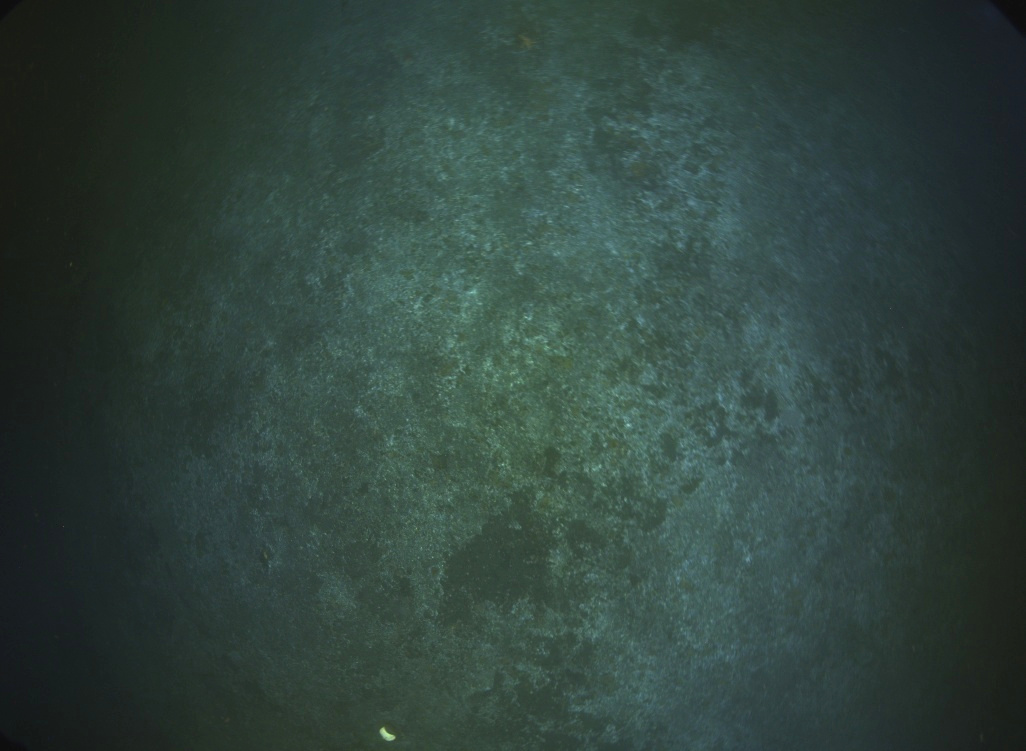}						 \includegraphics[width=0.2\textwidth]{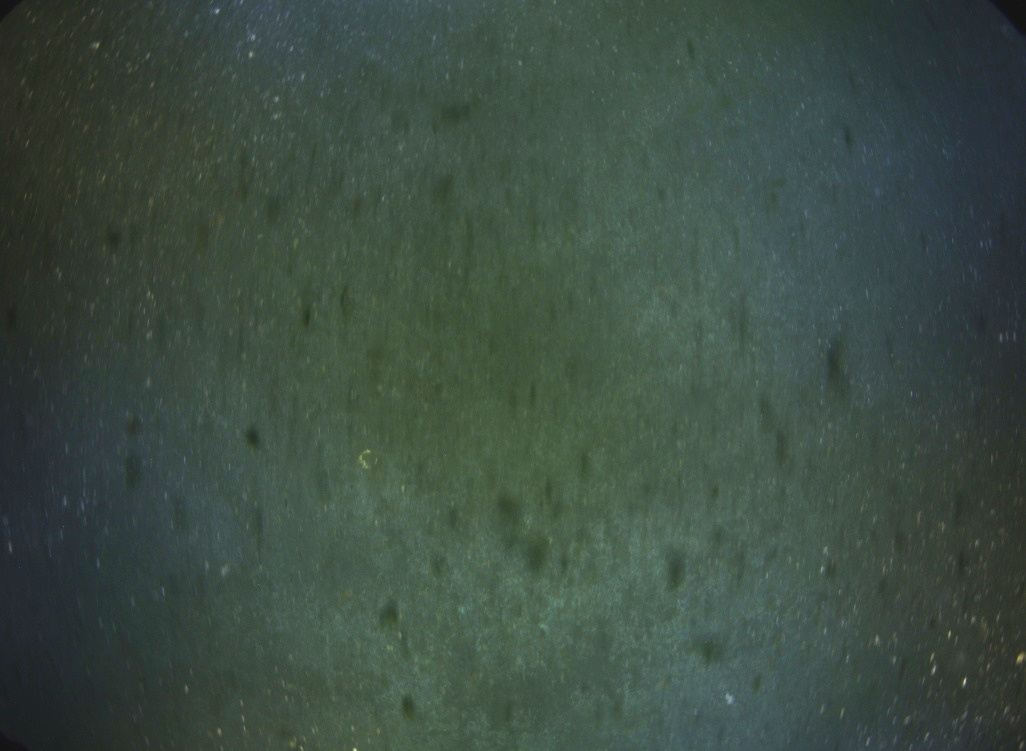}			\includegraphics[width=0.2\textwidth]{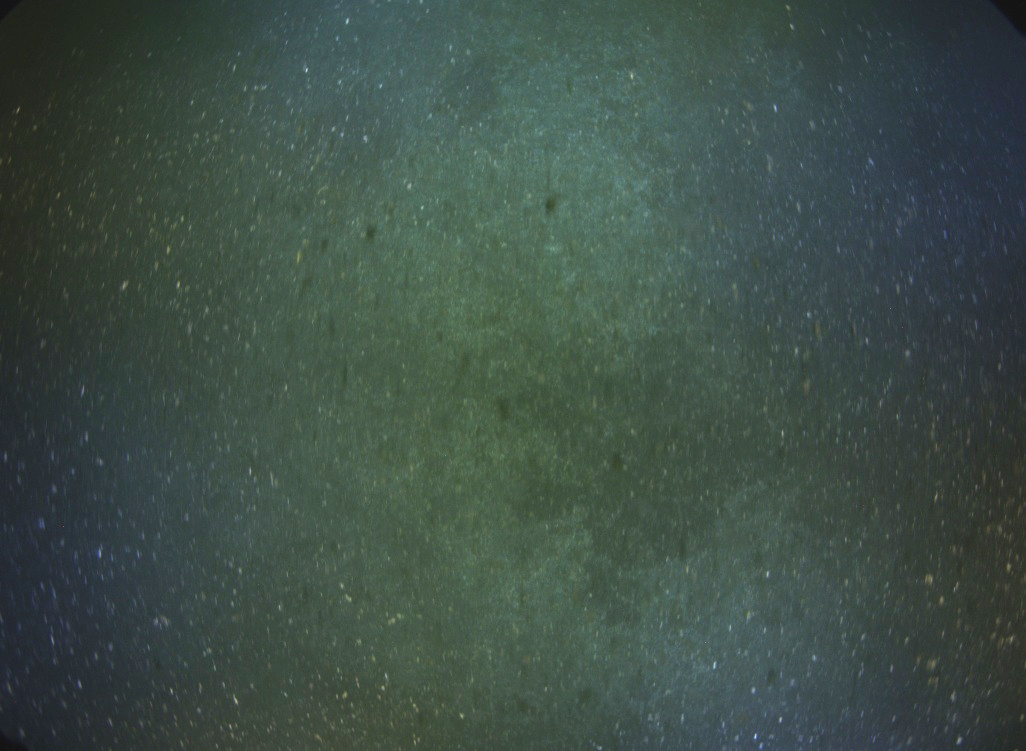}
		\includegraphics[width=0.18\textwidth]{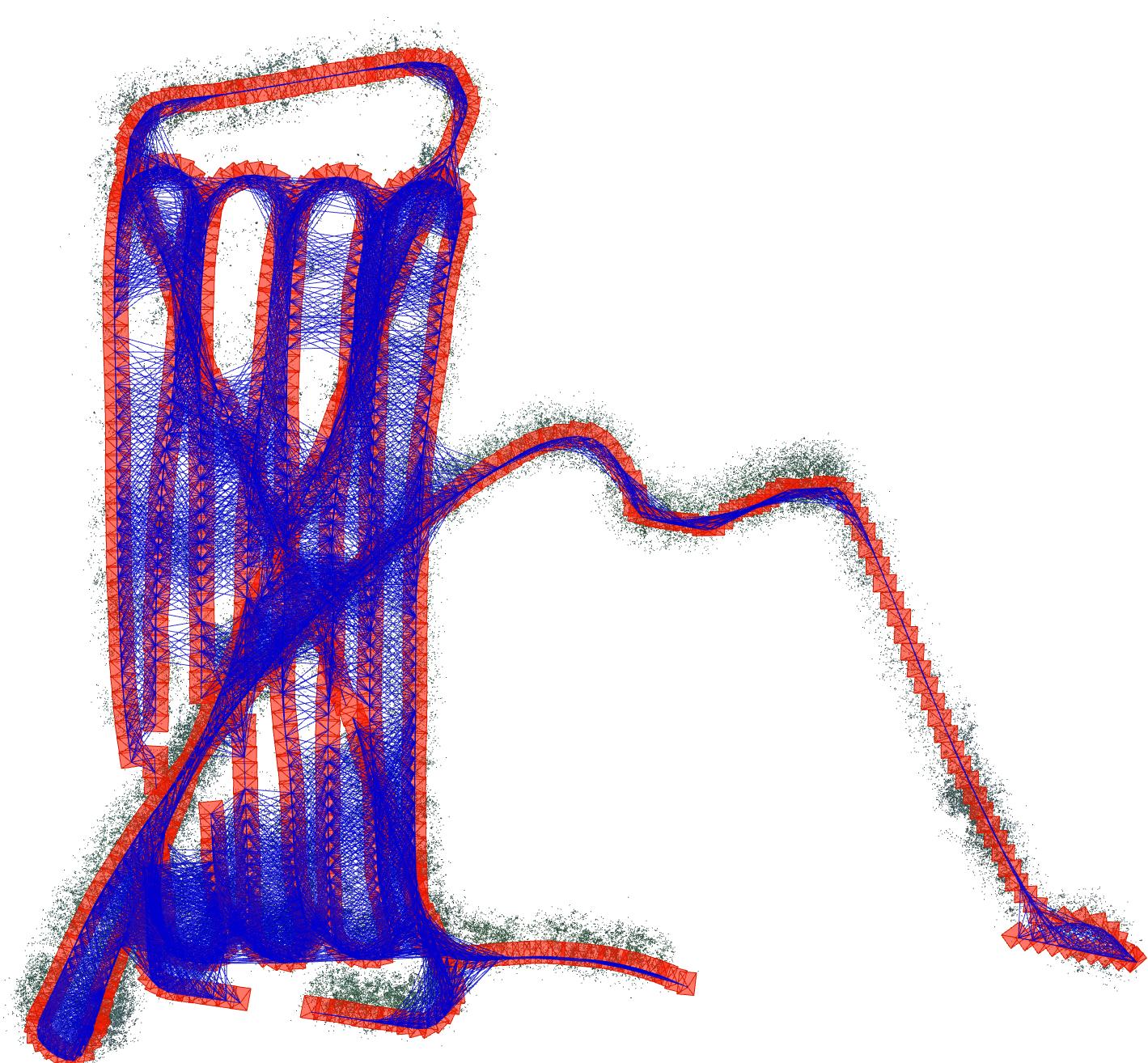}
	}\\
	\subfloat[\textit{Hard2}]{
		\includegraphics[width=0.2\textwidth]{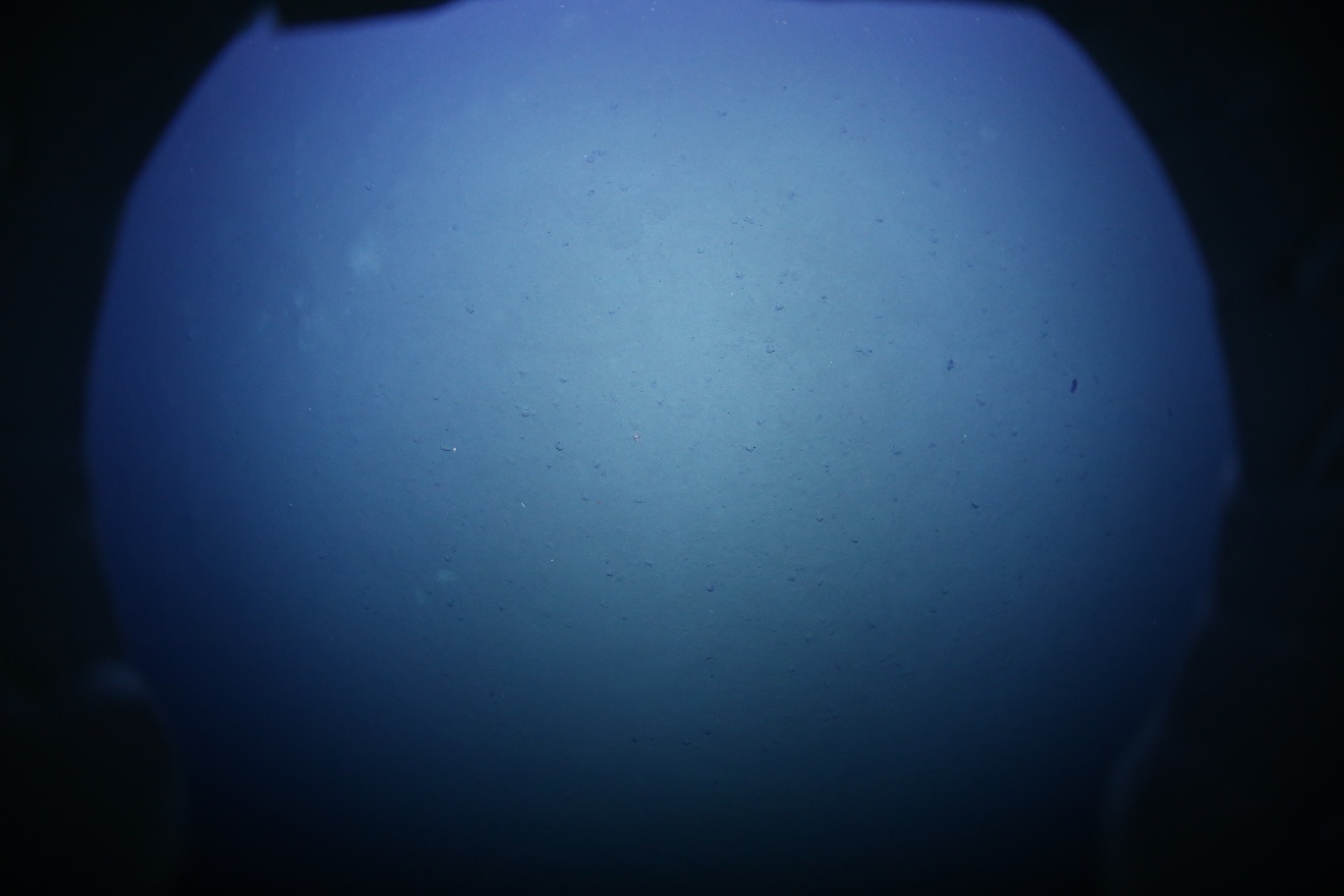}
		\includegraphics[width=0.2\textwidth]{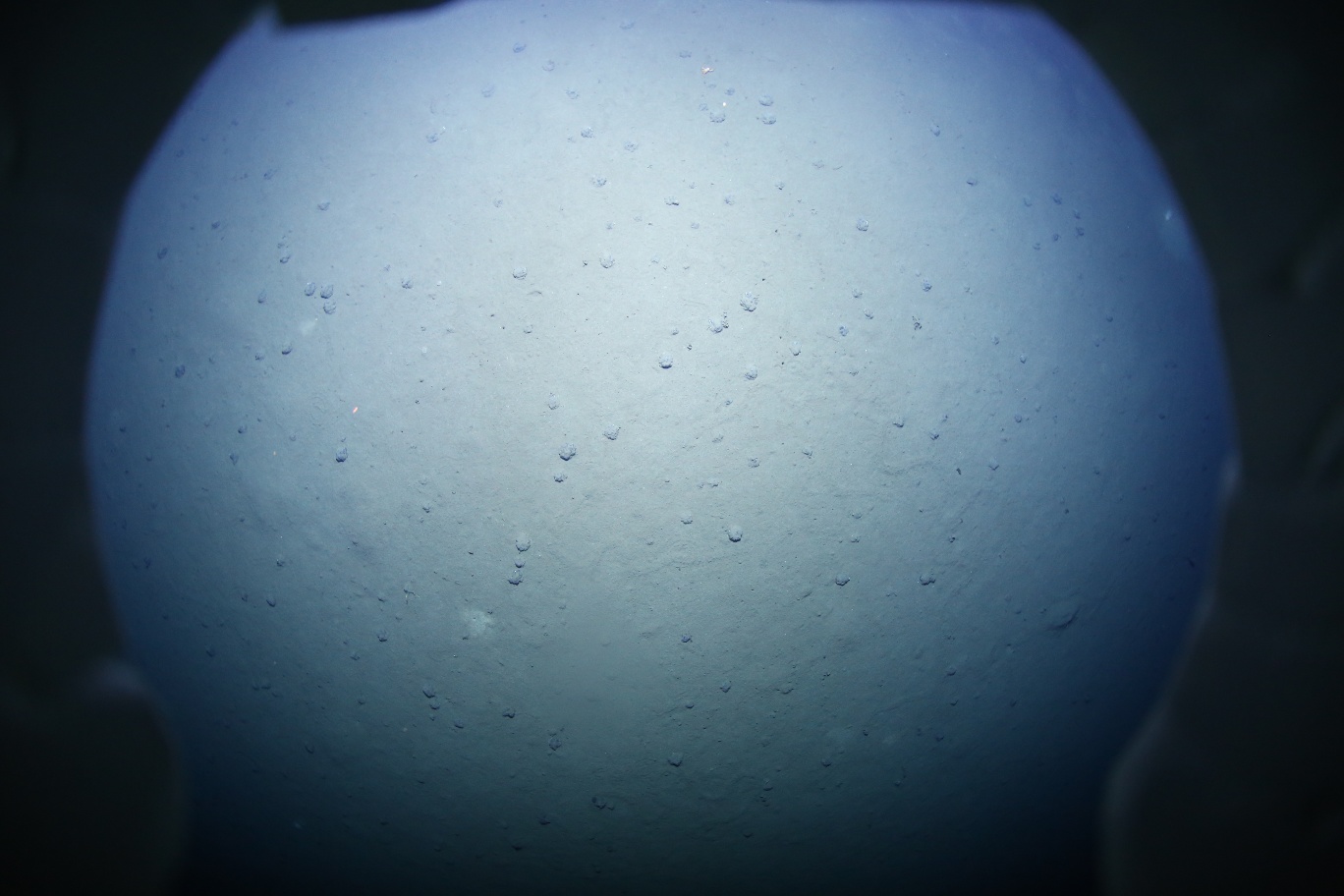}						 \includegraphics[width=0.2\textwidth]{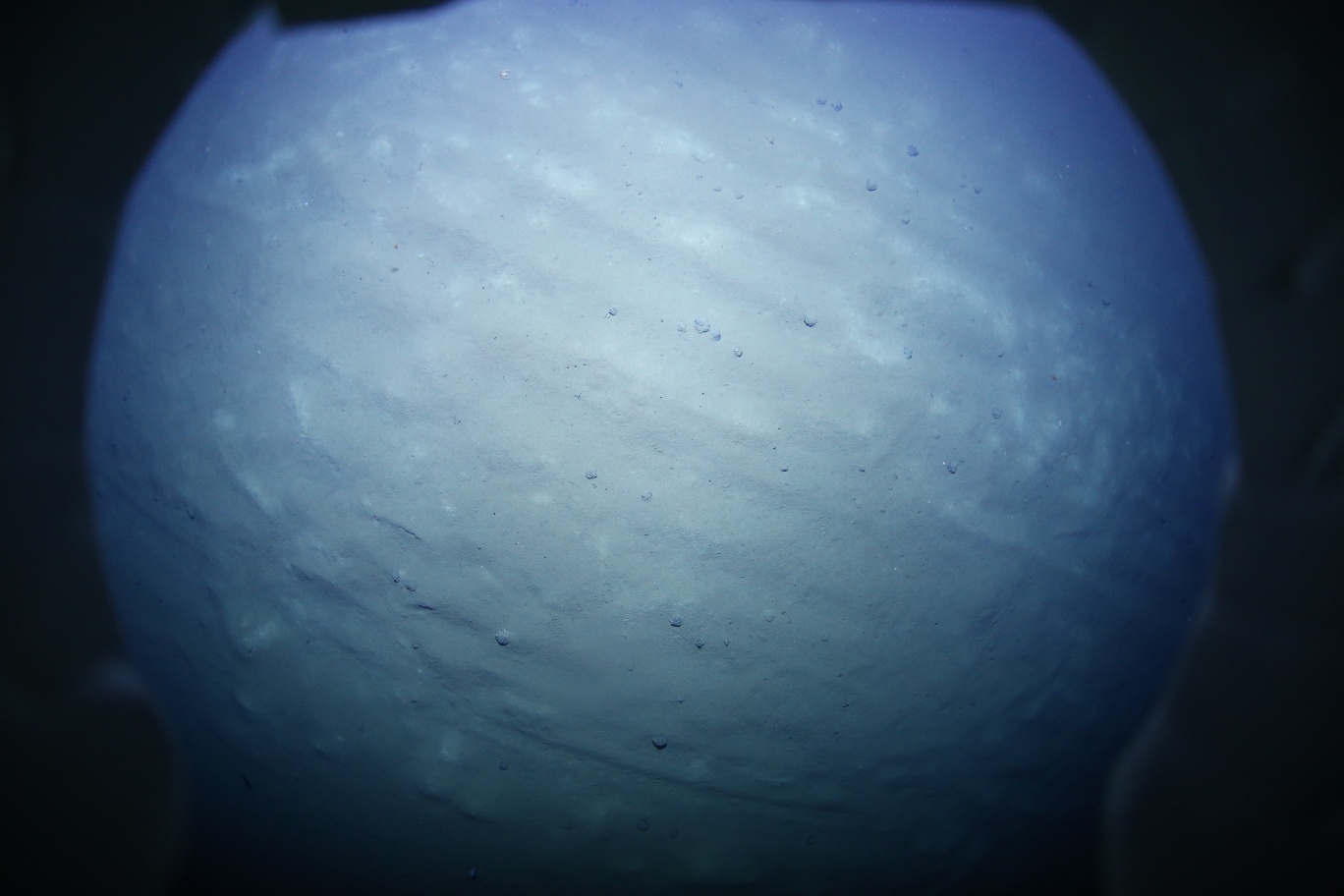}			\includegraphics[width=0.2\textwidth]{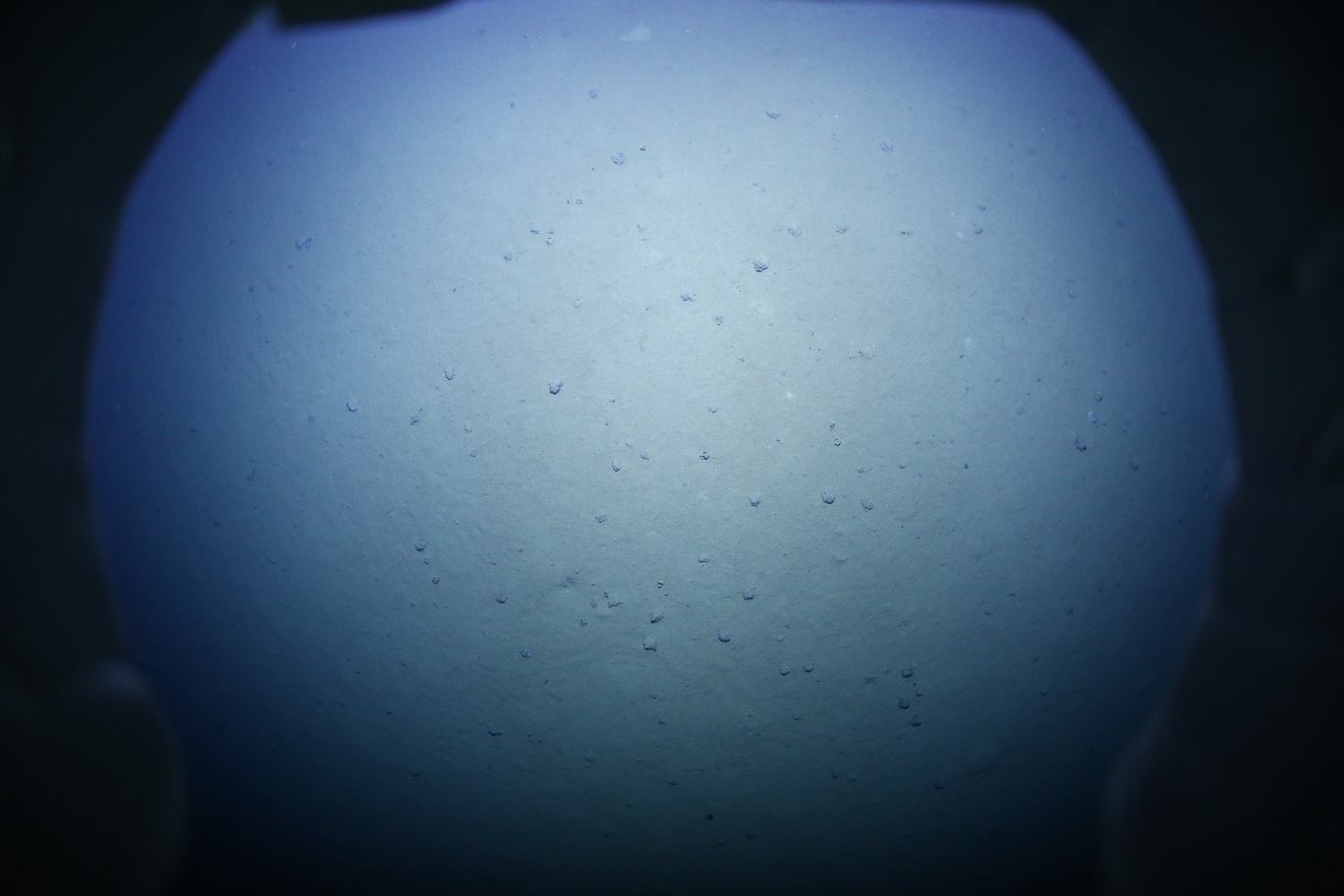}
		\includegraphics[width=0.18\textwidth]{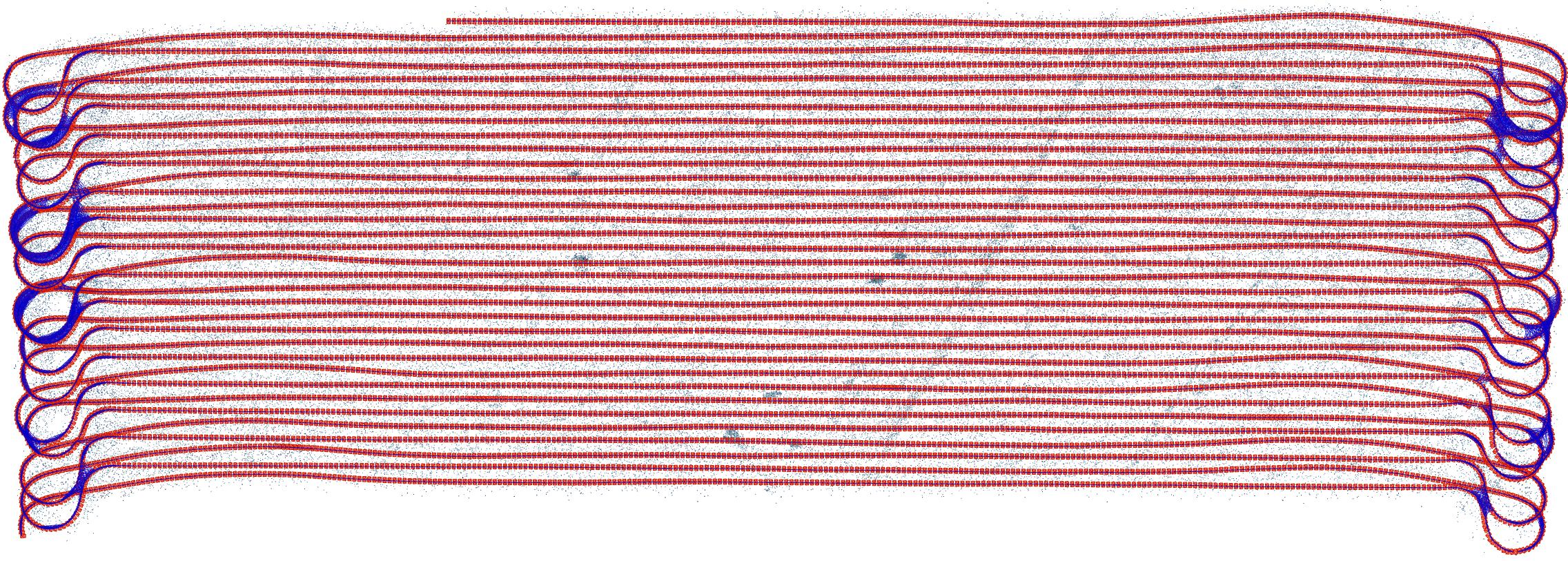}
	}\\
	\caption{An overview of the self-gathered AUV datasets, where each dataset is represented in a row. Four example images from each dataset is displayed to provide a visual representation of the data. Additionally, each rightmost figure shows the overall AUV mission where the camera trajectory is visualized in red, and the blue lines represent the connections between image pairs that share common visible points. The density of the blue lines indicates the strength of the view graph connectivity.}
	\label{fig:datasets_overview}
\end{figure*}

\section{Evaluation}

In this section, we evaluate the performance of our proposed pipeline using various datasets gathered by GEOMAR's AUVs under different sea conditions.
To implement our approach, we utilize the state-of-art SfM software package COLMAP \cite{schonberger2016structure} as the underlying framework.
We first integrate the navigation data into the incremental SfM process of COLMAP, and then we build upon it to implement our hierarchical SfM approach.
For the implementation details, we set the number of times to revisit the weak-area to 2.
For the global pose graph optimization, we set $\rho_{rel}=1.0$, $\rho_{abs} = 0.001$ and $\rho_{sm} = 2.0$.

\textbf{Datasets.} To analyze the performance of our pipeline, we empirically categorize our datasets into different levels of difficulty based on factors such as the imaging conditions, image quality, visibility, water scattering, light variations, and the connectivity of the view graph.
In each dataset, the AUV follows a classical lawn mower pattern, maintaining a stable flying altitude. 
Additionally, the AUV is planned to have sufficient side-track image overlap, and subsequently, a cross-track is planned to facilitate loop-closures.
The navigation data utilized in the dataset contains latitude, longitude and depth for the positional component and yaw, pitch, and roll for the rotational component. 
Figure \ref{fig:datasets_overview} provides an overview of the datasets, showcasing four example images for each dataset. Additionally, the rightmost figure depicts the AUV trajectory during data collection. 
The camera trajectory is visualized in red, and the pink lines represent the connections between image pairs that share common visible points.
The density of the pink lines indicates the strength of the view graph connectivity.

Before evaluating the datasets, we observe distinct characteristics between different datasets.
In the \textit{Easy} dataset (44$m$ $\times$ 35$m$), the input images exhibit clear visibility and are well illuminated, resulting in a dense view graph connection, allowing for strong feature matching and robust reconstruction.
In the \textit{Medium1} (45$m$ $\times$ 42$m$), the water scattering is mild, but the image overlap is less dense compared to the \textit{Easy} dataset in certain areas.
The \textit{Medium2} (51$m$ $\times$ 18$m$) dataset consists of clear images, however, the target scene, which consists of a sea-grass area, lacks distinctive features. 
As a result, feature matches are limited, especially at the image border due to the varying light cone. 
This leads to a weaker side-track connectivity in the view graph.
The \textit{Hard1} (8$m$ $\times$ 21$m$) dataset presents challenges due to strong scattering in the images. 
In certain areas, the water conditions are extremely murky, making it difficult to find any features for reconstruction. 
Consequently, these regions cannot be effectively reconstructed, despite the view graph being sufficiently dense.
The \textit{Hard2} (442$m$ $\times$ 133$m$) dataset represents an exceptionally challenging scenario where the AUV mapped a Maganness nodule field in water depths of more than 4000m to monitor the deep-sea impacts \cite{peukert2018understanding}. 
The target scene lacks distinctive features, and the presence of strong scattering and varying illumination further hinders feature matching. 
Despite a high ratio of side-track image overlap, the side-ward view graph connections remain poor.

\subsection{Results}
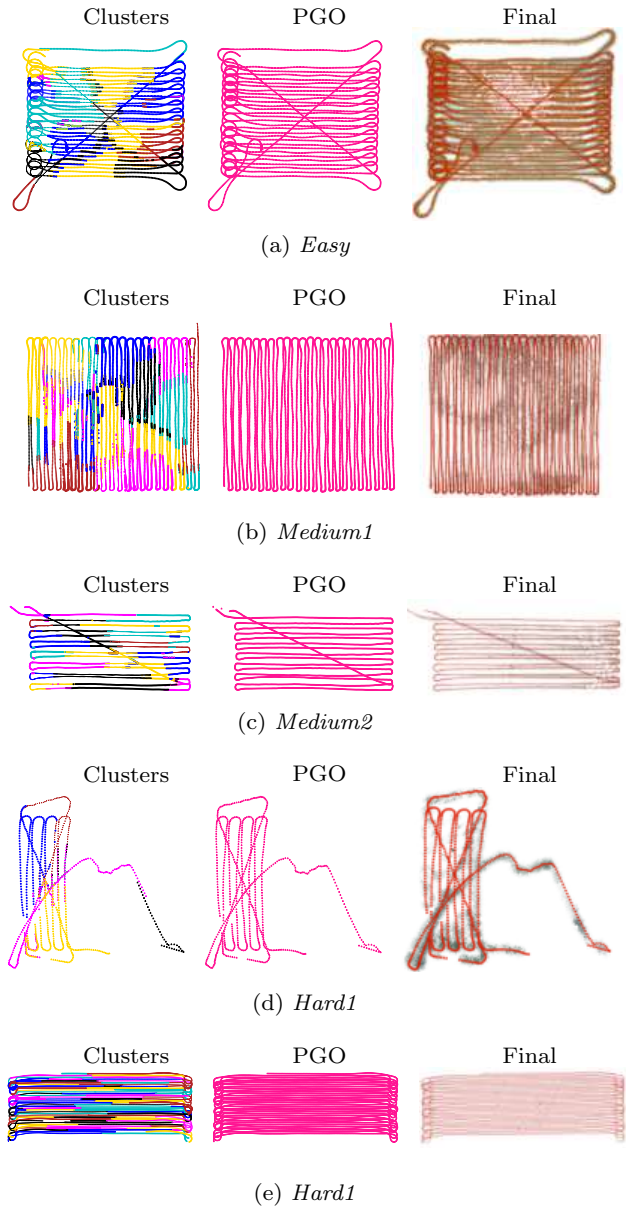
\begin{figure}[!ht]
	\centering
	\subfloat[\textit{Easy}]{
		\def\svgwidth{0.99\columnwidth}
		\begingroup%
  \makeatletter%
  \providecommand\color[2][]{%
    \errmessage{(Inkscape) Color is used for the text in Inkscape, but the package 'color.sty' is not loaded}%
    \renewcommand\color[2][]{}%
  }%
  \providecommand\transparent[1]{%
    \errmessage{(Inkscape) Transparency is used (non-zero) for the text in Inkscape, but the package 'transparent.sty' is not loaded}%
    \renewcommand\transparent[1]{}%
  }%
  \providecommand\rotatebox[2]{#2}%
  \newcommand*\fsize{\dimexpr\f@size pt\relax}%
  \newcommand*\lineheight[1]{\fontsize{\fsize}{#1\fsize}\selectfont}%
  \ifx\svgwidth\undefined%
    \setlength{\unitlength}{239bp}%
    \ifx\svgscale\undefined%
      \relax%
    \else%
      \setlength{\unitlength}{\unitlength * \real{\svgscale}}%
    \fi%
  \else%
    \setlength{\unitlength}{\svgwidth}%
  \fi%
  \global\let\svgwidth\undefined%
  \global\let\svgscale\undefined%
  \makeatother%
  \begin{picture}(1,0.34309623)%
    \lineheight{1}%
    \setlength\tabcolsep{0pt}%
    \put(0.14115057,0.31771319){\color[rgb]{0,0,0}\makebox(0,0)[lt]{\lineheight{1.25}\smash{\begin{tabular}[t]{l}Clusters\end{tabular}}}}%
    \put(0.47236468,0.31808092){\color[rgb]{0,0,0}\makebox(0,0)[lt]{\lineheight{1.25}\smash{\begin{tabular}[t]{l}PGO\end{tabular}}}}%
    \put(0.80445705,0.31771316){\color[rgb]{0,0,0}\makebox(0,0)[lt]{\lineheight{1.25}\smash{\begin{tabular}[t]{l}Final\end{tabular}}}}%
    \put(0,0){\includegraphics[width=\unitlength,page=1]{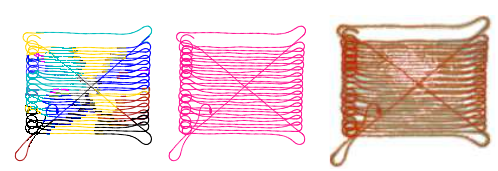}}%
  \end{picture}%
\endgroup%

	}\\
	\subfloat[\textit{Medium1}]{
		\def\svgwidth{0.99\columnwidth}
		\begingroup%
  \makeatletter%
  \providecommand\color[2][]{%
    \errmessage{(Inkscape) Color is used for the text in Inkscape, but the package 'color.sty' is not loaded}%
    \renewcommand\color[2][]{}%
  }%
  \providecommand\transparent[1]{%
    \errmessage{(Inkscape) Transparency is used (non-zero) for the text in Inkscape, but the package 'transparent.sty' is not loaded}%
    \renewcommand\transparent[1]{}%
  }%
  \providecommand\rotatebox[2]{#2}%
  \newcommand*\fsize{\dimexpr\f@size pt\relax}%
  \newcommand*\lineheight[1]{\fontsize{\fsize}{#1\fsize}\selectfont}%
  \ifx\svgwidth\undefined%
    \setlength{\unitlength}{239bp}%
    \ifx\svgscale\undefined%
      \relax%
    \else%
      \setlength{\unitlength}{\unitlength * \real{\svgscale}}%
    \fi%
  \else%
    \setlength{\unitlength}{\svgwidth}%
  \fi%
  \global\let\svgwidth\undefined%
  \global\let\svgscale\undefined%
  \makeatother%
  \begin{picture}(1,0.35146444)%
    \lineheight{1}%
    \setlength\tabcolsep{0pt}%
    \put(0.14115057,0.32608139){\color[rgb]{0,0,0}\makebox(0,0)[lt]{\lineheight{1.25}\smash{\begin{tabular}[t]{l}Clusters\end{tabular}}}}%
    \put(0.47236468,0.32644913){\color[rgb]{0,0,0}\makebox(0,0)[lt]{\lineheight{1.25}\smash{\begin{tabular}[t]{l}PGO\end{tabular}}}}%
    \put(0.80445705,0.32608136){\color[rgb]{0,0,0}\makebox(0,0)[lt]{\lineheight{1.25}\smash{\begin{tabular}[t]{l}Final\end{tabular}}}}%
    \put(0,0){\includegraphics[width=\unitlength,page=1]{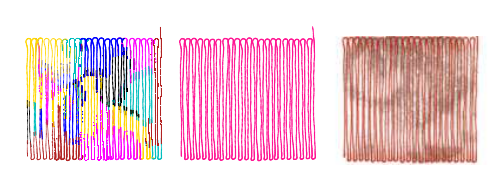}}%
  \end{picture}%
\endgroup%

	}\\
	\subfloat[\textit{Medium2}]{
		\def\svgwidth{0.99\columnwidth}
		\begingroup%
  \makeatletter%
  \providecommand\color[2][]{%
    \errmessage{(Inkscape) Color is used for the text in Inkscape, but the package 'color.sty' is not loaded}%
    \renewcommand\color[2][]{}%
  }%
  \providecommand\transparent[1]{%
    \errmessage{(Inkscape) Transparency is used (non-zero) for the text in Inkscape, but the package 'transparent.sty' is not loaded}%
    \renewcommand\transparent[1]{}%
  }%
  \providecommand\rotatebox[2]{#2}%
  \newcommand*\fsize{\dimexpr\f@size pt\relax}%
  \newcommand*\lineheight[1]{\fontsize{\fsize}{#1\fsize}\selectfont}%
  \ifx\svgwidth\undefined%
    \setlength{\unitlength}{239bp}%
    \ifx\svgscale\undefined%
      \relax%
    \else%
      \setlength{\unitlength}{\unitlength * \real{\svgscale}}%
    \fi%
  \else%
    \setlength{\unitlength}{\svgwidth}%
  \fi%
  \global\let\svgwidth\undefined%
  \global\let\svgscale\undefined%
  \makeatother%
  \begin{picture}(1,0.19665272)%
    \lineheight{1}%
    \setlength\tabcolsep{0pt}%
    \put(0.14115057,0.17126967){\color[rgb]{0,0,0}\makebox(0,0)[lt]{\lineheight{1.25}\smash{\begin{tabular}[t]{l}Clusters\end{tabular}}}}%
    \put(0.47236468,0.17163741){\color[rgb]{0,0,0}\makebox(0,0)[lt]{\lineheight{1.25}\smash{\begin{tabular}[t]{l}PGO\end{tabular}}}}%
    \put(0.80445705,0.17126965){\color[rgb]{0,0,0}\makebox(0,0)[lt]{\lineheight{1.25}\smash{\begin{tabular}[t]{l}Final\end{tabular}}}}%
    \put(0,0){\includegraphics[width=\unitlength,page=1]{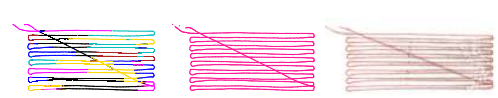}}%
  \end{picture}%
\endgroup%

	}\\
	\subfloat[\textit{Hard1}]{
		\def\svgwidth{0.99\columnwidth}
		\begingroup%
  \makeatletter%
  \providecommand\color[2][]{%
    \errmessage{(Inkscape) Color is used for the text in Inkscape, but the package 'color.sty' is not loaded}%
    \renewcommand\color[2][]{}%
  }%
  \providecommand\transparent[1]{%
    \errmessage{(Inkscape) Transparency is used (non-zero) for the text in Inkscape, but the package 'transparent.sty' is not loaded}%
    \renewcommand\transparent[1]{}%
  }%
  \providecommand\rotatebox[2]{#2}%
  \newcommand*\fsize{\dimexpr\f@size pt\relax}%
  \newcommand*\lineheight[1]{\fontsize{\fsize}{#1\fsize}\selectfont}%
  \ifx\svgwidth\undefined%
    \setlength{\unitlength}{239bp}%
    \ifx\svgscale\undefined%
      \relax%
    \else%
      \setlength{\unitlength}{\unitlength * \real{\svgscale}}%
    \fi%
  \else%
    \setlength{\unitlength}{\svgwidth}%
  \fi%
  \global\let\svgwidth\undefined%
  \global\let\svgscale\undefined%
  \makeatother%
  \begin{picture}(1,0.34309623)%
    \lineheight{1}%
    \setlength\tabcolsep{0pt}%
    \put(0,0){\includegraphics[width=\unitlength,page=1]{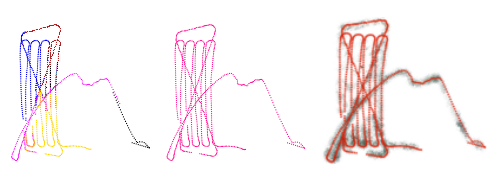}}%
    \put(0.14115057,0.31771319){\color[rgb]{0,0,0}\makebox(0,0)[lt]{\lineheight{1.25}\smash{\begin{tabular}[t]{l}Clusters\end{tabular}}}}%
    \put(0.47236467,0.31808111){\color[rgb]{0,0,0}\makebox(0,0)[lt]{\lineheight{1.25}\smash{\begin{tabular}[t]{l}PGO\end{tabular}}}}%
    \put(0.80445705,0.31771316){\color[rgb]{0,0,0}\makebox(0,0)[lt]{\lineheight{1.25}\smash{\begin{tabular}[t]{l}Final\end{tabular}}}}%
  \end{picture}%
\endgroup%

	}\\
	\subfloat[\textit{Hard1}]{
		\def\svgwidth{0.99\columnwidth}
		\begingroup%
  \makeatletter%
  \providecommand\color[2][]{%
    \errmessage{(Inkscape) Color is used for the text in Inkscape, but the package 'color.sty' is not loaded}%
    \renewcommand\color[2][]{}%
  }%
  \providecommand\transparent[1]{%
    \errmessage{(Inkscape) Transparency is used (non-zero) for the text in Inkscape, but the package 'transparent.sty' is not loaded}%
    \renewcommand\transparent[1]{}%
  }%
  \providecommand\rotatebox[2]{#2}%
  \newcommand*\fsize{\dimexpr\f@size pt\relax}%
  \newcommand*\lineheight[1]{\fontsize{\fsize}{#1\fsize}\selectfont}%
  \ifx\svgwidth\undefined%
    \setlength{\unitlength}{239bp}%
    \ifx\svgscale\undefined%
      \relax%
    \else%
      \setlength{\unitlength}{\unitlength * \real{\svgscale}}%
    \fi%
  \else%
    \setlength{\unitlength}{\svgwidth}%
  \fi%
  \global\let\svgwidth\undefined%
  \global\let\svgscale\undefined%
  \makeatother%
  \begin{picture}(1,0.19665272)%
    \lineheight{1}%
    \setlength\tabcolsep{0pt}%
    \put(0.14115057,0.17126967){\color[rgb]{0,0,0}\makebox(0,0)[lt]{\lineheight{1.25}\smash{\begin{tabular}[t]{l}Clusters\end{tabular}}}}%
    \put(0.47236468,0.17163741){\color[rgb]{0,0,0}\makebox(0,0)[lt]{\lineheight{1.25}\smash{\begin{tabular}[t]{l}PGO\end{tabular}}}}%
    \put(0.80445705,0.17126965){\color[rgb]{0,0,0}\makebox(0,0)[lt]{\lineheight{1.25}\smash{\begin{tabular}[t]{l}Final\end{tabular}}}}%
    \put(0,0){\includegraphics[width=\unitlength,page=1]{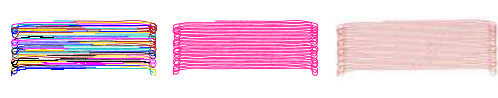}}%
  \end{picture}%
\endgroup%

	}\\
	\caption{Intermediate processing results. From left to right: The resulted camera poses from clustered reconstructions; The pose graph optimization; The final sparse reconstruction.}
	\label{fig:clusters}
\end{figure}

We first present the intermediate processing results of the given datasets in Fig. \ref{fig:clusters} where the left figures depict the clustered reconstruction results and the center figures show the camera poses after the global pose graph optimization step and the rightmost figures show the final sparse reconstructions.

Next, we evaluate the results of our approach and compare it to default COLMAP \cite{schonberger2016structure} as a baseline work.
We refer to our implementation of incremental SfM supervised by the navigation data as INC-NAV, while Ours represents our proposed hierarchical SfM approach.
We therefore refer to the default, non-expert setting of COLMAP as COLMAP-DEF.
However, it is known that COLMAP was originally designed for reconstructing internet photo collections, their default settings might not be suitable for robotic mapping scenarios.
We make certain modifications in the settings of COLMAP to make it better suited for underwater robotic missions, which is referred as COLMAP-PRO.
Specifically, we set the minimum number of inliers for pose estimation to be 6, allowing for the reconstruction of images with a limited number of features, ensuring completeness in the reconstruction.
Moreover, we set the maximum number of trials for re-triangulation to be 5, which allows us to make multiple attempts to re-triangulate image pairs and this helps better to close potential loops.
Since COLMAP produces arbitrarily scaled reconstructions, we employ robust similarity transformation estimation to align the reconstruction with the given navigation trajectory for metric evaluations.

\begin{figure*}[!ht]
	\centering
	\subfloat[\textit{Easy}]{
		\def\svgwidth{0.8\textwidth}
		\begingroup%
  \makeatletter%
  \providecommand\color[2][]{%
    \errmessage{(Inkscape) Color is used for the text in Inkscape, but the package 'color.sty' is not loaded}%
    \renewcommand\color[2][]{}%
  }%
  \providecommand\transparent[1]{%
    \errmessage{(Inkscape) Transparency is used (non-zero) for the text in Inkscape, but the package 'transparent.sty' is not loaded}%
    \renewcommand\transparent[1]{}%
  }%
  \providecommand\rotatebox[2]{#2}%
  \newcommand*\fsize{\dimexpr\f@size pt\relax}%
  \newcommand*\lineheight[1]{\fontsize{\fsize}{#1\fsize}\selectfont}%
  \ifx\svgwidth\undefined%
    \setlength{\unitlength}{495bp}%
    \ifx\svgscale\undefined%
      \relax%
    \else%
      \setlength{\unitlength}{\unitlength * \real{\svgscale}}%
    \fi%
  \else%
    \setlength{\unitlength}{\svgwidth}%
  \fi%
  \global\let\svgwidth\undefined%
  \global\let\svgscale\undefined%
  \makeatother%
  \begin{picture}(1,0.36363636)%
    \lineheight{1}%
    \setlength\tabcolsep{0pt}%
    \put(0.05613118,0.33718962){\makebox(0,0)[lt]{\lineheight{1.25}\smash{\begin{tabular}[t]{l}\textbf{COLMAP-DEF}\end{tabular}}}}%
    \put(0.30809968,0.33718962){\makebox(0,0)[lt]{\lineheight{1.25}\smash{\begin{tabular}[t]{l}\textbf{COLMAP-PRO}\end{tabular}}}}%
    \put(0.57067299,0.33718962){\makebox(0,0)[lt]{\lineheight{1.25}\smash{\begin{tabular}[t]{l}\textbf{INC-NAV}\end{tabular}}}}%
    \put(0.84066949,0.33718962){\makebox(0,0)[lt]{\lineheight{1.25}\smash{\begin{tabular}[t]{l}\textbf{Ours}\end{tabular}}}}%
    \put(0,0){\includegraphics[width=\unitlength,page=1]{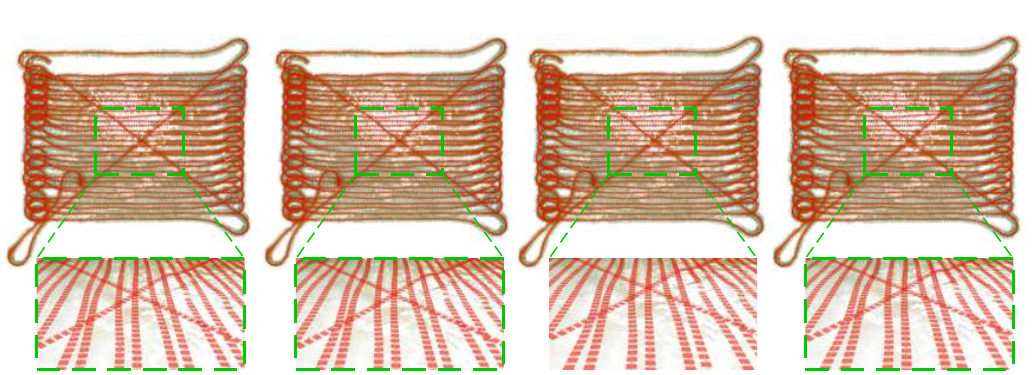}}%
    \put(0.01091573,0.3321876){\makebox(0,0)[lt]{\lineheight{1.25}\smash{\begin{tabular}[t]{l}10m\end{tabular}}}}%
    \put(0,0){\includegraphics[width=\unitlength,page=2]{compare_easy.pdf}}%
  \end{picture}%
\endgroup%

	}\\
	\subfloat[\textit{Medium1}]{
		\def\svgwidth{0.8\textwidth}
		\begingroup%
  \makeatletter%
  \providecommand\color[2][]{%
    \errmessage{(Inkscape) Color is used for the text in Inkscape, but the package 'color.sty' is not loaded}%
    \renewcommand\color[2][]{}%
  }%
  \providecommand\transparent[1]{%
    \errmessage{(Inkscape) Transparency is used (non-zero) for the text in Inkscape, but the package 'transparent.sty' is not loaded}%
    \renewcommand\transparent[1]{}%
  }%
  \providecommand\rotatebox[2]{#2}%
  \newcommand*\fsize{\dimexpr\f@size pt\relax}%
  \newcommand*\lineheight[1]{\fontsize{\fsize}{#1\fsize}\selectfont}%
  \ifx\svgwidth\undefined%
    \setlength{\unitlength}{495bp}%
    \ifx\svgscale\undefined%
      \relax%
    \else%
      \setlength{\unitlength}{\unitlength * \real{\svgscale}}%
    \fi%
  \else%
    \setlength{\unitlength}{\svgwidth}%
  \fi%
  \global\let\svgwidth\undefined%
  \global\let\svgscale\undefined%
  \makeatother%
  \begin{picture}(1,0.36363636)%
    \lineheight{1}%
    \setlength\tabcolsep{0pt}%
    \put(0.05613118,0.33718962){\makebox(0,0)[lt]{\lineheight{1.25}\smash{\begin{tabular}[t]{l}\textbf{COLMAP-DEF}\end{tabular}}}}%
    \put(0.30809968,0.33718962){\makebox(0,0)[lt]{\lineheight{1.25}\smash{\begin{tabular}[t]{l}\textbf{COLMAP-PRO}\end{tabular}}}}%
    \put(0.57067299,0.33718962){\makebox(0,0)[lt]{\lineheight{1.25}\smash{\begin{tabular}[t]{l}\textbf{INC-NAV}\end{tabular}}}}%
    \put(0.84066949,0.33718962){\makebox(0,0)[lt]{\lineheight{1.25}\smash{\begin{tabular}[t]{l}\textbf{Ours}\end{tabular}}}}%
    \put(0,0){\includegraphics[width=\unitlength,page=1]{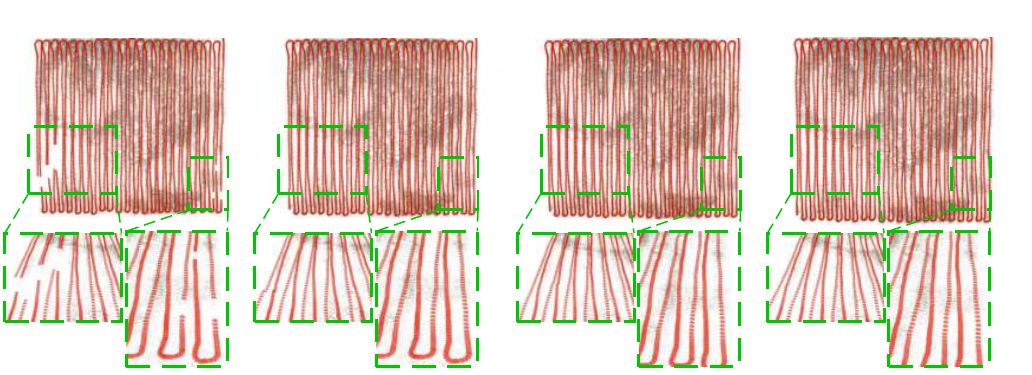}}%
    \put(0.22913775,0.31671292){\makebox(0,0)[lt]{\lineheight{1.25}\smash{\begin{tabular}[t]{l}10m\end{tabular}}}}%
    \put(0,0){\includegraphics[width=\unitlength,page=2]{compare_medium1.pdf}}%
  \end{picture}%
\endgroup%

	}\\
	\subfloat[\textit{Medium2}]{
		\def\svgwidth{0.8\textwidth}
		\begingroup%
  \makeatletter%
  \providecommand\color[2][]{%
    \errmessage{(Inkscape) Color is used for the text in Inkscape, but the package 'color.sty' is not loaded}%
    \renewcommand\color[2][]{}%
  }%
  \providecommand\transparent[1]{%
    \errmessage{(Inkscape) Transparency is used (non-zero) for the text in Inkscape, but the package 'transparent.sty' is not loaded}%
    \renewcommand\transparent[1]{}%
  }%
  \providecommand\rotatebox[2]{#2}%
  \newcommand*\fsize{\dimexpr\f@size pt\relax}%
  \newcommand*\lineheight[1]{\fontsize{\fsize}{#1\fsize}\selectfont}%
  \ifx\svgwidth\undefined%
    \setlength{\unitlength}{495bp}%
    \ifx\svgscale\undefined%
      \relax%
    \else%
      \setlength{\unitlength}{\unitlength * \real{\svgscale}}%
    \fi%
  \else%
    \setlength{\unitlength}{\svgwidth}%
  \fi%
  \global\let\svgwidth\undefined%
  \global\let\svgscale\undefined%
  \makeatother%
  \begin{picture}(1,0.43434343)%
    \lineheight{1}%
    \setlength\tabcolsep{0pt}%
    \put(0.05613118,0.40789669){\makebox(0,0)[lt]{\lineheight{1.25}\smash{\begin{tabular}[t]{l}\textbf{COLMAP-DEF}\end{tabular}}}}%
    \put(0.30809968,0.40789669){\makebox(0,0)[lt]{\lineheight{1.25}\smash{\begin{tabular}[t]{l}\textbf{COLMAP-PRO}\end{tabular}}}}%
    \put(0.57067299,0.40789669){\makebox(0,0)[lt]{\lineheight{1.25}\smash{\begin{tabular}[t]{l}\textbf{INC-NAV}\end{tabular}}}}%
    \put(0.84066949,0.40789669){\makebox(0,0)[lt]{\lineheight{1.25}\smash{\begin{tabular}[t]{l}\textbf{Ours}\end{tabular}}}}%
    \put(0,0){\includegraphics[width=\unitlength,page=1]{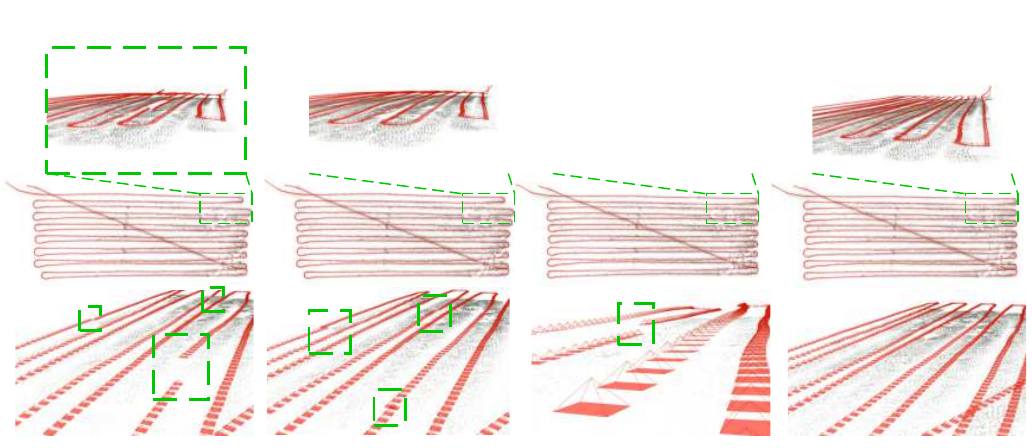}}%
    \put(0.01130866,0.15460717){\makebox(0,0)[lt]{\lineheight{1.25}\smash{\begin{tabular}[t]{l}10m\end{tabular}}}}%
    \put(0,0){\includegraphics[width=\unitlength,page=2]{compare_medium2.pdf}}%
  \end{picture}%
\endgroup%

	}\\
	\caption{Visual comparison results of datasets \textit{Easy}, \textit{Medium1} and \textit{Medium2}.}
	\label{fig:visual_comp_medium}
\end{figure*}

\begin{figure*}[!ht]
	\centering
	\subfloat[\textit{Hard1}]{
		\def\svgwidth{0.8\textwidth}
		\begingroup%
  \makeatletter%
  \providecommand\color[2][]{%
    \errmessage{(Inkscape) Color is used for the text in Inkscape, but the package 'color.sty' is not loaded}%
    \renewcommand\color[2][]{}%
  }%
  \providecommand\transparent[1]{%
    \errmessage{(Inkscape) Transparency is used (non-zero) for the text in Inkscape, but the package 'transparent.sty' is not loaded}%
    \renewcommand\transparent[1]{}%
  }%
  \providecommand\rotatebox[2]{#2}%
  \newcommand*\fsize{\dimexpr\f@size pt\relax}%
  \newcommand*\lineheight[1]{\fontsize{\fsize}{#1\fsize}\selectfont}%
  \ifx\svgwidth\undefined%
    \setlength{\unitlength}{495bp}%
    \ifx\svgscale\undefined%
      \relax%
    \else%
      \setlength{\unitlength}{\unitlength * \real{\svgscale}}%
    \fi%
  \else%
    \setlength{\unitlength}{\svgwidth}%
  \fi%
  \global\let\svgwidth\undefined%
  \global\let\svgscale\undefined%
  \makeatother%
  \begin{picture}(1,0.67676768)%
    \lineheight{1}%
    \setlength\tabcolsep{0pt}%
    \put(0.05613118,0.65032093){\makebox(0,0)[lt]{\lineheight{1.25}\smash{\begin{tabular}[t]{l}\textbf{COLMAP-DEF}\end{tabular}}}}%
    \put(0.30809968,0.65032093){\makebox(0,0)[lt]{\lineheight{1.25}\smash{\begin{tabular}[t]{l}\textbf{COLMAP-PRO}\end{tabular}}}}%
    \put(0.57067299,0.65032093){\makebox(0,0)[lt]{\lineheight{1.25}\smash{\begin{tabular}[t]{l}\textbf{INC-NAV}\end{tabular}}}}%
    \put(0.84066949,0.65032093){\makebox(0,0)[lt]{\lineheight{1.25}\smash{\begin{tabular}[t]{l}\textbf{Ours}\end{tabular}}}}%
    \put(0,0){\includegraphics[width=\unitlength,page=1]{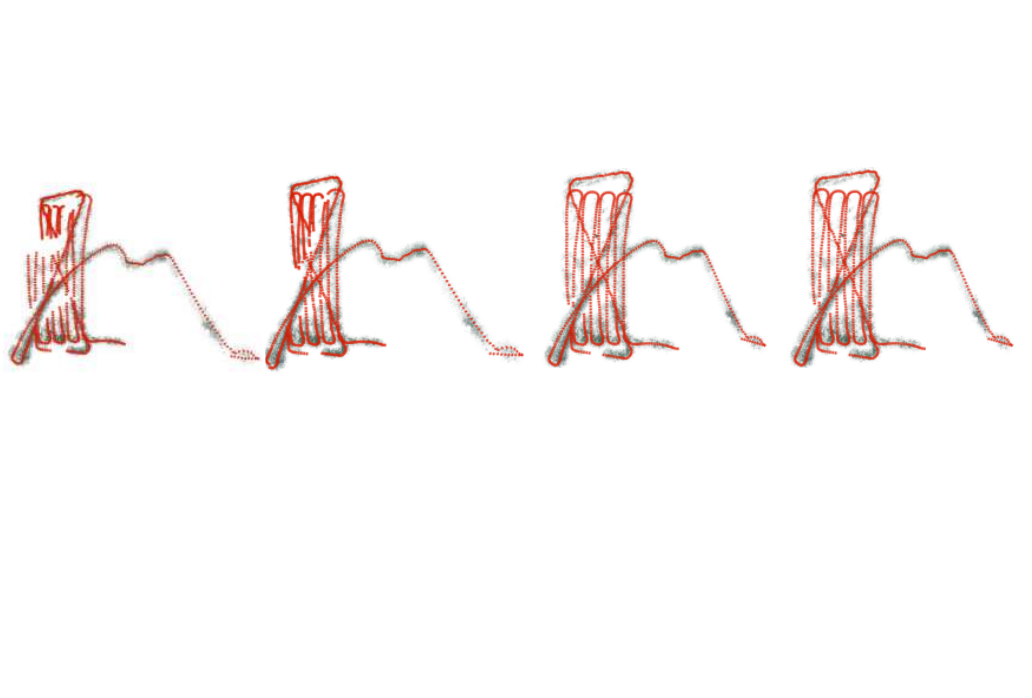}}%
    \put(0.05037924,0.31825561){\makebox(0,0)[lt]{\lineheight{1.25}\smash{\begin{tabular}[t]{l}8m\end{tabular}}}}%
    \put(0,0){\includegraphics[width=\unitlength,page=2]{compare_hard1.pdf}}%
  \end{picture}%
\endgroup%

	}\\
	\subfloat[\textit{Hard2}]{
		\def\svgwidth{0.8\textwidth}
		\begingroup%
  \makeatletter%
  \providecommand\color[2][]{%
    \errmessage{(Inkscape) Color is used for the text in Inkscape, but the package 'color.sty' is not loaded}%
    \renewcommand\color[2][]{}%
  }%
  \providecommand\transparent[1]{%
    \errmessage{(Inkscape) Transparency is used (non-zero) for the text in Inkscape, but the package 'transparent.sty' is not loaded}%
    \renewcommand\transparent[1]{}%
  }%
  \providecommand\rotatebox[2]{#2}%
  \newcommand*\fsize{\dimexpr\f@size pt\relax}%
  \newcommand*\lineheight[1]{\fontsize{\fsize}{#1\fsize}\selectfont}%
  \ifx\svgwidth\undefined%
    \setlength{\unitlength}{495bp}%
    \ifx\svgscale\undefined%
      \relax%
    \else%
      \setlength{\unitlength}{\unitlength * \real{\svgscale}}%
    \fi%
  \else%
    \setlength{\unitlength}{\svgwidth}%
  \fi%
  \global\let\svgwidth\undefined%
  \global\let\svgscale\undefined%
  \makeatother%
  \begin{picture}(1,0.72727273)%
    \lineheight{1}%
    \setlength\tabcolsep{0pt}%
    \put(0.08603547,0.70082599){\makebox(0,0)[lt]{\lineheight{1.25}\smash{\begin{tabular}[t]{l}\textbf{COLMAP-PRO}\end{tabular}}}}%
    \put(0.44935646,0.70082599){\makebox(0,0)[lt]{\lineheight{1.25}\smash{\begin{tabular}[t]{l}\textbf{INC-NAV}\end{tabular}}}}%
    \put(0.79464269,0.70082599){\makebox(0,0)[lt]{\lineheight{1.25}\smash{\begin{tabular}[t]{l}\textbf{Ours}\end{tabular}}}}%
    \put(0,0){\includegraphics[width=\unitlength,page=1]{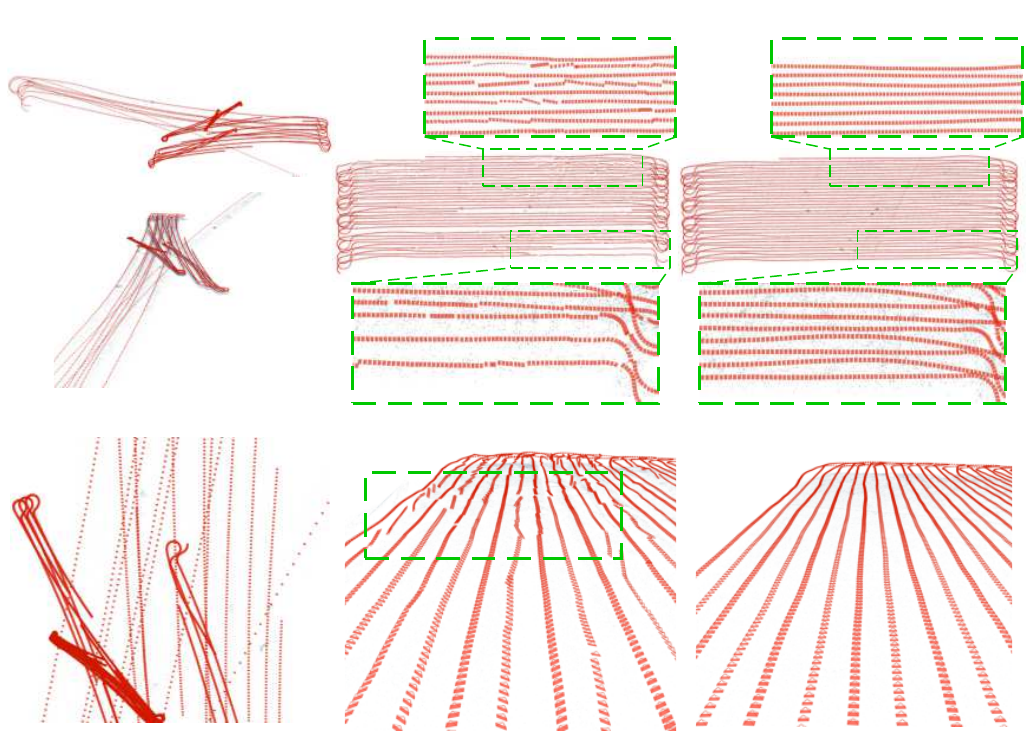}}%
    \put(0.34912614,0.60943709){\makebox(0,0)[lt]{\lineheight{1.25}\smash{\begin{tabular}[t]{l}100m\end{tabular}}}}%
    \put(0,0){\includegraphics[width=\unitlength,page=2]{compare_hard2.pdf}}%
  \end{picture}%
\endgroup%

	}\\
	\caption{Visual comparison results of datasets \textit{Hard1} and \textit{Hard2}.}
	\label{fig:visual_comp_hard}
\end{figure*}

A visual comparison of the evaluation results on the datasets can be seen in Fig. \ref{fig:visual_comp_medium} and Fig. \ref{fig:visual_comp_hard}. 
In addition, quantitative evaluation results are given in Table \ref{tab:eval_quant}.
The table includes the number of reconstructed images $N_c$ and the number of images $N$ available in the dataset.
$L$ represents the average track length of a single 3D point ($\sharp$ of images that observe this 3D point).
To evaluate the accuracy of the reconstruction, we report the Reprojection Error (RE) in pixels and the RMSE of the Absolute Translational Error (ATE) between the reconstructed camera poses and the given navigation data.
Regarding the running time, we record the duration of the SfM process in minutes, excluding the feature extraction and matching steps.
Note however that, neither the Reprojection Error nor the Absolute Translational Error can serve as definitive indicators of the reconstruction quality.
The reprojection error is an internal estimate that the bundle adjustment optimization is based on, and the ATE measures the similarity between the reconstructed camera poses and the given navigation data, which we use to guide our pose estimates during reconstruction.
As a result, it is expected that the reported ATE values in our approach would be relatively low.
Although the navigation data does not represent the ground truth of the vehicle trajectory, our ablation study reveals that it exhibits global reasonability but local inaccuracies.
As a result, although it cannot be used directly for the reconstruction process, it can serve as a reference trajectory.

As can be seen from the figures that our approach consistently produces visually more appealing results across almost all datasets.
In terms of running time, our approach demonstrates faster reconstruction times compared to the other methods shown in the table, except for the \textit{Hard1} dataset.
The reason behind this discrepancy is that the \textit{Hard1} dataset contains a relatively small number of images, around 1000, which limits the advantage of the divide-and-conquer strategy in terms of efficiency.
In terms of statistics, enabling the PRO-setting leads to an increased number of reconstructed images.
Our approach achieves the highest number of reconstructed images with slightly increased reprojection error, which could be attributed to the fact that global SfM approaches are less susceptible to error accumulation as compared to incremental approaches.
Regarding the increased reprojection error, an explanation is that COLMAP solely optimizes for the reprojection error in bundle adjustment, whereas we incorporate additional information by penalizing the disparity between the reconstructed camera poses and their prior poses.

In terms of quality, the original COLMAP performs already very well in the \textit{Easy} dataset even with a dataset size of nearly 5000 images.
In scenarios where the imaging conditions are favorable and the view graph is dense, standard incremental SfM can still produce satisfactory and reliable reconstructions.
Our approach achieves similar reconstruction quality but with a reduced processing time.
However, as the imaging conditions underwater worsen, especially in the deep-sea environments, the level of difficulty increases, which leads to a decrease of the performance of standard incremental SfM, which finally results in inconsistent reconstructions.
For instance, in the \textit{Medium1} and \textit{Medium2} dataset (as depicted in Fig. \ref{fig:visual_comp_medium}), several inconsistencies in the camera trajectory can be observed, indicated by the green boxes.
These inconsistencies can mostly be explained by the accumulation of drift during the incremental reconstruction process.
In the \textit{Medium2} dataset, COLMAP produces a curved reconstruction of the seafloor, while in the INC-NAV case, the reconstruction appears to be flatter due to the supervision by the navigation data.
However, even with navigation data supervision, the accumulated error can still be significant, which in turn may prevent loop-closure.
For the \textit{Hard1} dataset, we specifically employed Domain-size Pooling (DSP-SIFT) \cite{dong2015domain} due to the poor performance of the default SIFT feature.
The images in this dataset suffer from low illumination and strong scattering, which adversely affects feature extraction and matching.
Despite neither approach achieving a complete reconstruction in this challenging dataset, our approach still demonstrates superior reconstruction quality.
In the case of the \textit{Hard2} dataset which contains more than 10000 images and challenging imaging conditions, the default setting of COLMAP fails to generate a valid output, with only 8 images being reconstructed. 
By enabling the PRO-setting, COLMAP is able to reconstruct a larger number of images, achieving a low reprojection error of 0.54 pixels, which is the lowest among all approaches. 
However, despite the low reprojection error, the visual appearance of the reconstruction in Fig. \ref{fig:visual_comp_hard} clearly indicates significant errors, including positional and scale drift.
INC-NAV, on the other hand, is able to address the scale drift issue but produces a poor camera trajectory. 
In contrast, our approach achieves both global consistency and local accuracy in the reconstruction. 
This conclusion can be supported by observing the reconstructed model of the seafloor, which shows the traces left behind by the Manganese nodule mining vehicles during the mining operation (see also the textured mesh in Fig. \ref{fig:meshing_hard2}).

\begin{table*}[!h]
	\centering
	\footnotesize
	\caption{Quantitative evaluation results on the self-gathered AUV datasets. $N_c$ denotes the number of reconstructed images and $N$ denotes the number of images available in the dataset. $L$ represents the average track length of a 3D point ($\sharp$ of images) and $T$ records the running time of the SfM process in minutes. To evaluate the reconstruction accuracy, we report the average reprojection error in pixels and the RMSE of the Absolute Translational Error (ATE) between the reconstructed camera poses and the given navigation.}
	\label{tab:eval_quant}       %
	\scalebox{0.65}{
		\begin{tabular}{ccccccccccccccccccccccccccc}
			\hline\noalign{\smallskip}
			\multirow{2}{*}{Datasets} & \multirow{2}{*}{$N$} & \multicolumn{5}{c}{\textbf{COLMAP-DEF}\cite{schonberger2016structure}} & \multicolumn{5}{c}{\textbf{COLMAP-PRO}} & \multicolumn{5}{c}{\textbf{INC-NAV}} & \multicolumn{5}{c}{\textbf{Ours}}\\
			& & $N_c$ & $L$ & RE & ATE & $T$ & $N_c$ & $L$ & RE & ATE & $T$ & $N_c$ & $L$ & RE & ATE & $T$ & $N_c$ & $L$ & RE & ATE & $T$\\
			\noalign{\smallskip}\hline\noalign{\smallskip}
			\textit{Easy} & 4746 & \textbf{4710} & 5.09 & \textbf{0.68} & 0.407 & 4123.53 & \textbf{4710} & 5.09 & \textbf{0.68} & 0.405 & 4581.42 & \textbf{4710} & 5.09 & \textbf{0.68} & \textbf{0.165} & 5251.75 & \textbf{4710} & \textbf{5.12} & 0.70 & 0.183 & \textbf{1021.692}\\
			\textit{Medium1} & 5752 & 5635 & 4.28 & 0.56 & 0.526 & 1284.55 & \textbf{5740} & 4.25 & 0.56 & 0.515 & 1300.11 & \textbf{5740} & 4.28 & \textbf{0.53} & 0.568 & 1551.92  & \textbf{5740} & \textbf{4.31} & 0.58 & \textbf{0.325} & \textbf{344.25}\\
			\textit{Medium2} & 2977 & 2872 & 3.69 & \textbf{0.51} & 0.513 & 294.16 & \textbf{2881} & 3.68 & 0.52 & 0.531 & 519.61 & \textbf{2881} & 3.68 & 0.55 & \textbf{0.114} & 511.50 & \textbf{2881} & \textbf{3.71} & 0.54 & 0.183 & \textbf{123.20}\\
			\textit{Hard1} & 1065 & 808 & 4.25 & 0.75 & 1.182 & \textbf{26.51} & 847 & 4.32 & 0.75 & 1.348 & 36.68 & 847 & 4.47 & \textbf{0.52} & \textbf{0.095} & 68.11 & \textbf{848} & \textbf{4.51} & 0.53 & \textbf{0.095} & 60.41 \\
			\textit{Hard2} & 10715 & 8 & - & - & - & - & 10479 & \textbf{3.24} & \textbf{0.54} & 101.08 & 206.70 & 10025 & 3.15 & 0.69 & \textbf{0.476} & 458.23 & \textbf{10713} & 3.23 & 0.62 & 0.487 & \textbf{126.41}\\
			\hline\noalign{\smallskip}
		\end{tabular}
	}
\end{table*}

\subsection{Ablation Study}
\begin{table}[!h]
	\centering
	\footnotesize
	\caption{Evaluation results of performing direct triangulation of 3D points using camera poses obtained from the navigation data and the pose graph optimization.}
	\label{tab:eval_ablation}       %
	\scalebox{0.9}{
		\begin{tabular}{ccccccccccccccccccccccccccc}
			\hline\noalign{\smallskip}
			\multirow{2}{*}{Datasets} & \multicolumn{2}{c}{\textbf{DT}} & \multicolumn{2}{c}{\textbf{PGO}} & \multicolumn{2}{c}{\textbf{PGO(inlier)}}\\
			& RE & $L$ & RE & $L$ & RE & $L$ \\
			\noalign{\smallskip}\hline\noalign{\smallskip}
			\textit{Easy} & 27.24 & 3.47 & 2.28 & \textbf{4.72} & \textbf{1.56} & 4.70\\
			\textit{Medium1} & 31.50 & 2.85 & 1.62 & \textbf{4.14} & \textbf{1.20} & 4.08\\
			\textit{Medium2} & 42.32 & 2.60 & 1.18 & \textbf{3.67} & \textbf{0.87} & 3.64\\			
			\textit{Hard1} & 21.00 & 3.76 & 2.66 & \textbf{4.07} & \textbf{1.47} & 4.03\\
			\textit{Hard2} & 23.01 & 2.81 & 1.35 & \textbf{3.23} & \textbf{0.92} & 3.19\\
			\noalign{\smallskip}\hline\noalign{\smallskip}
			AVG. & 29.01 & 3.68 & 1.82 & \textbf{3.97} & \textbf{1.20} & 3.93\\
			\hline\noalign{\smallskip}
		\end{tabular}
	}
\end{table}
\textbf{On the Accuracy of the Navigation and the Pose Graph.} To assess the accuracy of the navigation data obtained from a deep-sea AUV, we conduct an experiment to directly triangulate (DT) 3D points based on the given poses.
The underlying premise is that if the navigation data is sufficiently accurate, it should enable direct 3D reconstruction without using Structure-from-Motion.
Instead of presenting the uncertainty measures of the navigation data, this approach allows us to evaluate the consistency between the navigation data and the actual visual measurements.

In addition to the direct triangulation using the prior poses, we also perform direct triangulation of 3D points using camera poses obtained from the global pose graph optimization (PGO) in our proposed hierarchical SfM approach.
This allows us to evaluate the accuracy of the camera poses after the global motion averaging step.
Since we perform local SfM on clusters, we are able to collect inlier 3D points in this step.
Therefore, we additionally report the results of directly triangulating the true inlier feature matches instead of using all available feature matches, which is referred to as PGO(inlier).
It is important to note that the final global bundle adjustment is not performed in this evaluation.

The evaluation results are shown in Table \ref{tab:eval_ablation}, where RE stands for the Reprojection Error in pixels and $L$ is the average track length.
From the table, it can be observed that directly triangulating 3D points from the navigation data alone was not very successful, resulting in an average reprojection error of 29.01 pixels.
However, when applying the same procedure to camera poses obtained from the global pose graph optimization, an average reprojection error of 1.82 pixels was achieved.
Furthermore, by excluding the outliers, an average reprojection error of 1.20 pixels can be achieved, which is only possible in our proposed approach.
These results indicate a significant improvement in the accuracy of the computed camera poses and serve as a good starting point for the final bundle adjustment.

\begin{figure*}[!ht]
	\centering
	\def\svgwidth{0.99\textwidth}
	\begingroup%
  \makeatletter%
  \providecommand\color[2][]{%
    \errmessage{(Inkscape) Color is used for the text in Inkscape, but the package 'color.sty' is not loaded}%
    \renewcommand\color[2][]{}%
  }%
  \providecommand\transparent[1]{%
    \errmessage{(Inkscape) Transparency is used (non-zero) for the text in Inkscape, but the package 'transparent.sty' is not loaded}%
    \renewcommand\transparent[1]{}%
  }%
  \providecommand\rotatebox[2]{#2}%
  \newcommand*\fsize{\dimexpr\f@size pt\relax}%
  \newcommand*\lineheight[1]{\fontsize{\fsize}{#1\fsize}\selectfont}%
  \ifx\svgwidth\undefined%
    \setlength{\unitlength}{495bp}%
    \ifx\svgscale\undefined%
      \relax%
    \else%
      \setlength{\unitlength}{\unitlength * \real{\svgscale}}%
    \fi%
  \else%
    \setlength{\unitlength}{\svgwidth}%
  \fi%
  \global\let\svgwidth\undefined%
  \global\let\svgscale\undefined%
  \makeatother%
  \begin{picture}(1,0.82828283)%
    \lineheight{1}%
    \setlength\tabcolsep{0pt}%
    \put(0.20986499,0.43908786){\makebox(0,0)[lt]{\lineheight{1.25}\smash{\begin{tabular}[t]{l}\textbf{PGO without Weak-area Revisit}\end{tabular}}}}%
    \put(0.2156599,0.03252224){\makebox(0,0)[lt]{\lineheight{1.25}\smash{\begin{tabular}[t]{l}\textbf{PGO with Weak-area Revisit}\end{tabular}}}}%
    \put(0,0){\includegraphics[width=\unitlength,page=1]{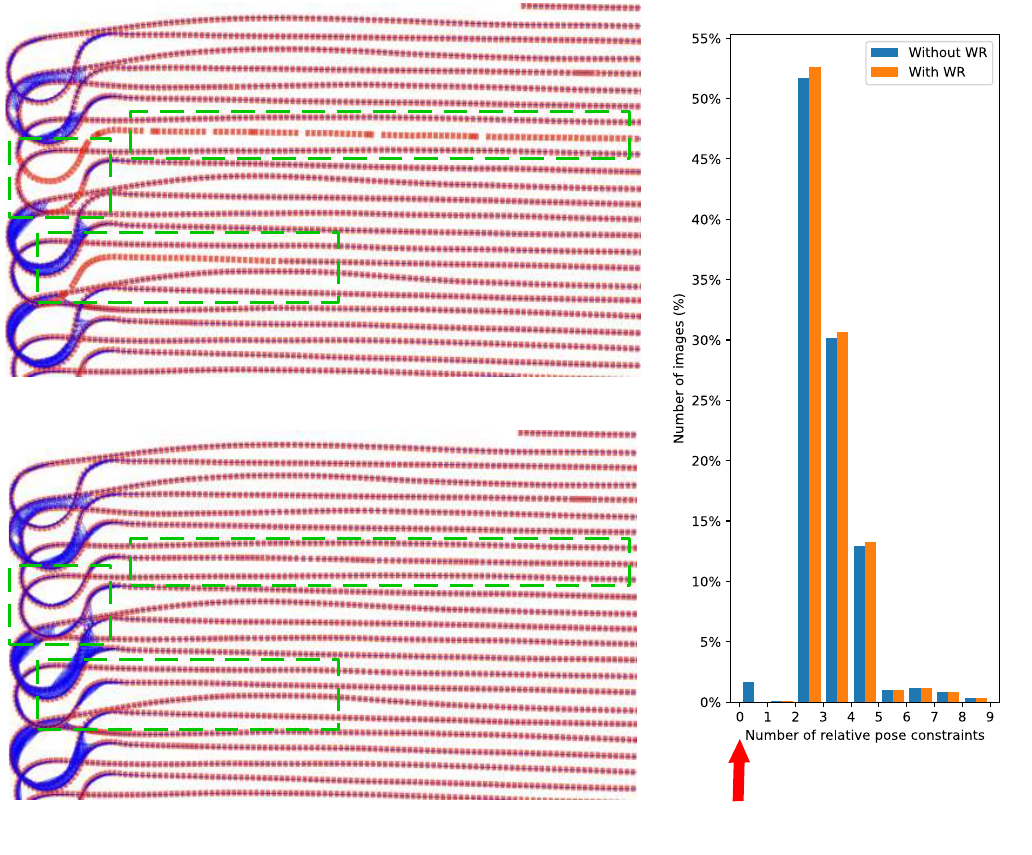}}%
    \put(0.69148737,0.03234312){\rotatebox{0.87892424}{\makebox(0,0)[lt]{\lineheight{1.25}\smash{\begin{tabular}[t]{l}\textbf{Weak-area}\end{tabular}}}}}%
    \put(0,0){\includegraphics[width=\unitlength,page=2]{wr_revisit_sonne.pdf}}%
  \end{picture}%
\endgroup%

	\caption{A comparison of the pose graph optimization results without weak-area revisiting (\textbf{top-left}) and with weak-area revisiting (\textbf{bottom-left}). The green boxes indicate the weak-areas before and after revisiting. The blue lines represent the connected image pairs in the view graph. \textbf{Right} part of the figure shows a histogram of the number of pose graph constraints for each image in the pose graph optimization. A higher number of constraints for an image indicates higher robustness and accuracy.}
	\label{fig:ablation_wr_revisit}
\end{figure*}
\textbf{On Weak-area Revisit.} The experiment conducted to analyze the effectiveness of the proposed weak-area revisiting is presented in Fig. \ref{fig:ablation_wr_revisit}.
The top-left figure shows the reconstruction results of the \textit{Hard2} dataset after the global pose graph optimization step (PGO) without weak-area revisiting, while the bottom-left figure shows the results with weak-area revisiting.
The green boxes in the figures indicate the detected weak-areas where the images receive a limited number of relative pose constraints.
However, after revisiting these weak-areas, the challenging images can also be well-reconstructed, as evident from the improved reconstruction quality in the bottom-left figure.
The right part of the figure presents a histogram of the number of relative pose constraints for each image in the pose graph optimization.
It can be observed that weak-area revisiting effectively eliminates the weak areas by $100\%$, ensuring a more robust and consistent pose graph optimization.

\begin{figure*}[!ht]
	\centering
	\def\svgwidth{0.99\textwidth}
	\begingroup%
  \makeatletter%
  \providecommand\color[2][]{%
    \errmessage{(Inkscape) Color is used for the text in Inkscape, but the package 'color.sty' is not loaded}%
    \renewcommand\color[2][]{}%
  }%
  \providecommand\transparent[1]{%
    \errmessage{(Inkscape) Transparency is used (non-zero) for the text in Inkscape, but the package 'transparent.sty' is not loaded}%
    \renewcommand\transparent[1]{}%
  }%
  \providecommand\rotatebox[2]{#2}%
  \newcommand*\fsize{\dimexpr\f@size pt\relax}%
  \newcommand*\lineheight[1]{\fontsize{\fsize}{#1\fsize}\selectfont}%
  \ifx\svgwidth\undefined%
    \setlength{\unitlength}{495bp}%
    \ifx\svgscale\undefined%
      \relax%
    \else%
      \setlength{\unitlength}{\unitlength * \real{\svgscale}}%
    \fi%
  \else%
    \setlength{\unitlength}{\svgwidth}%
  \fi%
  \global\let\svgwidth\undefined%
  \global\let\svgscale\undefined%
  \makeatother%
  \begin{picture}(1,0.22222222)%
    \lineheight{1}%
    \setlength\tabcolsep{0pt}%
    \put(0.19943097,0.00956504){\makebox(0,0)[lt]{\lineheight{1.25}\smash{\begin{tabular}[t]{l}Original Color\end{tabular}}}}%
    \put(0.68486946,0.00650183){\makebox(0,0)[lt]{\lineheight{1.25}\smash{\begin{tabular}[t]{l}Normalized Color\end{tabular}}}}%
    \put(0,0){\includegraphics[width=\unitlength,page=1]{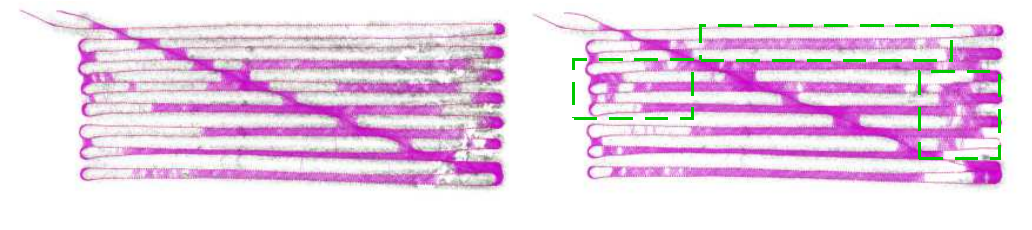}}%
    \put(0.04078487,0.04025947){\makebox(0,0)[lt]{\lineheight{1.25}\smash{\begin{tabular}[t]{l}10m\end{tabular}}}}%
    \put(0,0){\includegraphics[width=\unitlength,page=2]{color_norm_compare_anton131.pdf}}%
  \end{picture}%
\endgroup%

	\def\svgwidth{0.99\textwidth}
	\begingroup%
  \makeatletter%
  \providecommand\color[2][]{%
    \errmessage{(Inkscape) Color is used for the text in Inkscape, but the package 'color.sty' is not loaded}%
    \renewcommand\color[2][]{}%
  }%
  \providecommand\transparent[1]{%
    \errmessage{(Inkscape) Transparency is used (non-zero) for the text in Inkscape, but the package 'transparent.sty' is not loaded}%
    \renewcommand\transparent[1]{}%
  }%
  \providecommand\rotatebox[2]{#2}%
  \newcommand*\fsize{\dimexpr\f@size pt\relax}%
  \newcommand*\lineheight[1]{\fontsize{\fsize}{#1\fsize}\selectfont}%
  \ifx\svgwidth\undefined%
    \setlength{\unitlength}{495bp}%
    \ifx\svgscale\undefined%
      \relax%
    \else%
      \setlength{\unitlength}{\unitlength * \real{\svgscale}}%
    \fi%
  \else%
    \setlength{\unitlength}{\svgwidth}%
  \fi%
  \global\let\svgwidth\undefined%
  \global\let\svgscale\undefined%
  \makeatother%
  \begin{picture}(1,0.23232323)%
    \lineheight{1}%
    \setlength\tabcolsep{0pt}%
    \put(0.17215822,0.02269631){\makebox(0,0)[lt]{\lineheight{1.25}\smash{\begin{tabular}[t]{l}Original Color\end{tabular}}}}%
    \put(0.68183915,0.01963311){\makebox(0,0)[lt]{\lineheight{1.25}\smash{\begin{tabular}[t]{l}Normalized Color\end{tabular}}}}%
    \put(0,0){\includegraphics[width=\unitlength,page=1]{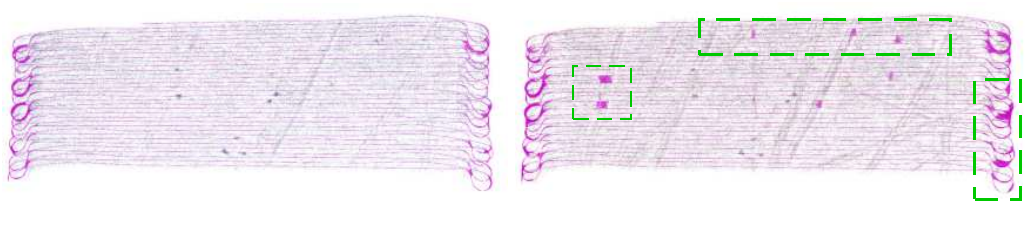}}%
    \put(0.05697112,0.03653391){\makebox(0,0)[lt]{\lineheight{1.25}\smash{\begin{tabular}[t]{l}100m\end{tabular}}}}%
    \put(0,0){\includegraphics[width=\unitlength,page=2]{color_norm_compare_sonne.pdf}}%
  \end{picture}%
\endgroup%

	\caption{A comparison of the view graph connectivity between the reconstruction results obtained using the original images and the color normalized images. The density of the pink lines indicates the strength of the view graph connectivity.}
	\label{fig:ablation_color_norm}
\end{figure*}

\begin{figure*}[!ht]
	\centering
	\def\svgwidth{0.99\textwidth}
	\begingroup%
  \makeatletter%
  \providecommand\color[2][]{%
    \errmessage{(Inkscape) Color is used for the text in Inkscape, but the package 'color.sty' is not loaded}%
    \renewcommand\color[2][]{}%
  }%
  \providecommand\transparent[1]{%
    \errmessage{(Inkscape) Transparency is used (non-zero) for the text in Inkscape, but the package 'transparent.sty' is not loaded}%
    \renewcommand\transparent[1]{}%
  }%
  \providecommand\rotatebox[2]{#2}%
  \newcommand*\fsize{\dimexpr\f@size pt\relax}%
  \newcommand*\lineheight[1]{\fontsize{\fsize}{#1\fsize}\selectfont}%
  \ifx\svgwidth\undefined%
    \setlength{\unitlength}{495bp}%
    \ifx\svgscale\undefined%
      \relax%
    \else%
      \setlength{\unitlength}{\unitlength * \real{\svgscale}}%
    \fi%
  \else%
    \setlength{\unitlength}{\svgwidth}%
  \fi%
  \global\let\svgwidth\undefined%
  \global\let\svgscale\undefined%
  \makeatother%
  \begin{picture}(1,0.21212121)%
    \lineheight{1}%
    \setlength\tabcolsep{0pt}%
    \put(0.1630673,0.01461534){\makebox(0,0)[lt]{\lineheight{1.25}\smash{\begin{tabular}[t]{l}\textbf{Original Color}\end{tabular}}}}%
    \put(0.66062701,0.0145824){\makebox(0,0)[lt]{\lineheight{1.25}\smash{\begin{tabular}[t]{l}\textbf{Normalized Color}\end{tabular}}}}%
    \put(0,0){\includegraphics[width=\unitlength,page=1]{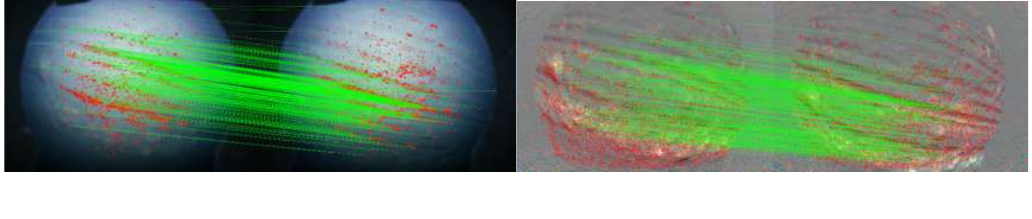}}%
  \end{picture}%
\endgroup%

	\def\svgwidth{0.99\textwidth}
	\begingroup%
  \makeatletter%
  \providecommand\color[2][]{%
    \errmessage{(Inkscape) Color is used for the text in Inkscape, but the package 'color.sty' is not loaded}%
    \renewcommand\color[2][]{}%
  }%
  \providecommand\transparent[1]{%
    \errmessage{(Inkscape) Transparency is used (non-zero) for the text in Inkscape, but the package 'transparent.sty' is not loaded}%
    \renewcommand\transparent[1]{}%
  }%
  \providecommand\rotatebox[2]{#2}%
  \newcommand*\fsize{\dimexpr\f@size pt\relax}%
  \newcommand*\lineheight[1]{\fontsize{\fsize}{#1\fsize}\selectfont}%
  \ifx\svgwidth\undefined%
    \setlength{\unitlength}{495bp}%
    \ifx\svgscale\undefined%
      \relax%
    \else%
      \setlength{\unitlength}{\unitlength * \real{\svgscale}}%
    \fi%
  \else%
    \setlength{\unitlength}{\svgwidth}%
  \fi%
  \global\let\svgwidth\undefined%
  \global\let\svgscale\undefined%
  \makeatother%
  \begin{picture}(1,0.21212121)%
    \lineheight{1}%
    \setlength\tabcolsep{0pt}%
    \put(0.1630673,0.01461534){\makebox(0,0)[lt]{\lineheight{1.25}\smash{\begin{tabular}[t]{l}\textbf{Original Color}\end{tabular}}}}%
    \put(0.66062701,0.0145824){\makebox(0,0)[lt]{\lineheight{1.25}\smash{\begin{tabular}[t]{l}\textbf{Normalized Color}\end{tabular}}}}%
    \put(0,0){\includegraphics[width=\unitlength,page=1]{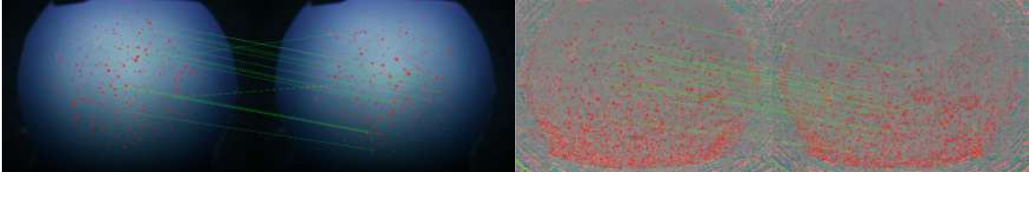}}%
  \end{picture}%
\endgroup%

	\caption{A comparison of the SIFT feature matches using the original images and the color normalized images.}
	\label{fig:ablation_color_norm_features}
\end{figure*}

\textbf{On Color Normalization.} A recent study by Grimaldi et al. \cite{grimaldi2023investigation} demonstrated that pre-processing approaches resulted in improvements in SLAM performance in visually challenging underwater environments.
We additionally conducted an experiment to perform reconstruction using the color normalized images and compare it against reconstruction using the original color images.
As can be seen from Fig. \ref{fig:ablation_color_norm}, the reconstruction using color normalized images exhibits improved connectivity in the resulting view graph, especially between side-tracks.
This can be attributed to the ability of color normalization to effectively reduce non-uniform illumination in the images, thereby enhancing feature matching (see also Fig. \ref{fig:ablation_color_norm_features}).
Moreover, the use of more sophisticated deep-learning-based feature matching methods, specifically designed for underwater conditions, has the potential to further enhance the quality of visual mapping in deep-sea environments. 
However, we leave this for future investigation.

\subsection{Meshing, Texturing and Orthophoto}
\begin{figure*}
	\centering
	\def\s{0.98\textwidth}
	\begingroup%
  \makeatletter%
  \providecommand\color[2][]{%
    \errmessage{(Inkscape) Color is used for the text in Inkscape, but the package 'color.sty' is not loaded}%
    \renewcommand\color[2][]{}%
  }%
  \providecommand\transparent[1]{%
    \errmessage{(Inkscape) Transparency is used (non-zero) for the text in Inkscape, but the package 'transparent.sty' is not loaded}%
    \renewcommand\transparent[1]{}%
  }%
  \providecommand\rotatebox[2]{#2}%
  \newcommand*\fsize{\dimexpr\f@size pt\relax}%
  \newcommand*\lineheight[1]{\fontsize{\fsize}{#1\fsize}\selectfont}%
  \ifx\svgwidth\undefined%
    \setlength{\unitlength}{495bp}%
    \ifx\svgscale\undefined%
      \relax%
    \else%
      \setlength{\unitlength}{\unitlength * \real{\svgscale}}%
    \fi%
  \else%
    \setlength{\unitlength}{\svgwidth}%
  \fi%
  \global\let\svgwidth\undefined%
  \global\let\svgscale\undefined%
  \makeatother%
  \begin{picture}(1,1.11111111)%
    \lineheight{1}%
    \setlength\tabcolsep{0pt}%
    \put(0.42,0.58930252){\makebox(0,0)[lt]{\lineheight{1.25}\smash{\begin{tabular}[t]{l}\textbf{Original Color}\end{tabular}}}}%
    \put(0.42,0.0199332){\makebox(0,0)[lt]{\lineheight{1.25}\smash{\begin{tabular}[t]{l}\textbf{Normalized Color}\end{tabular}}}}%
    \put(0,0){\includegraphics[width=\unitlength,page=1]{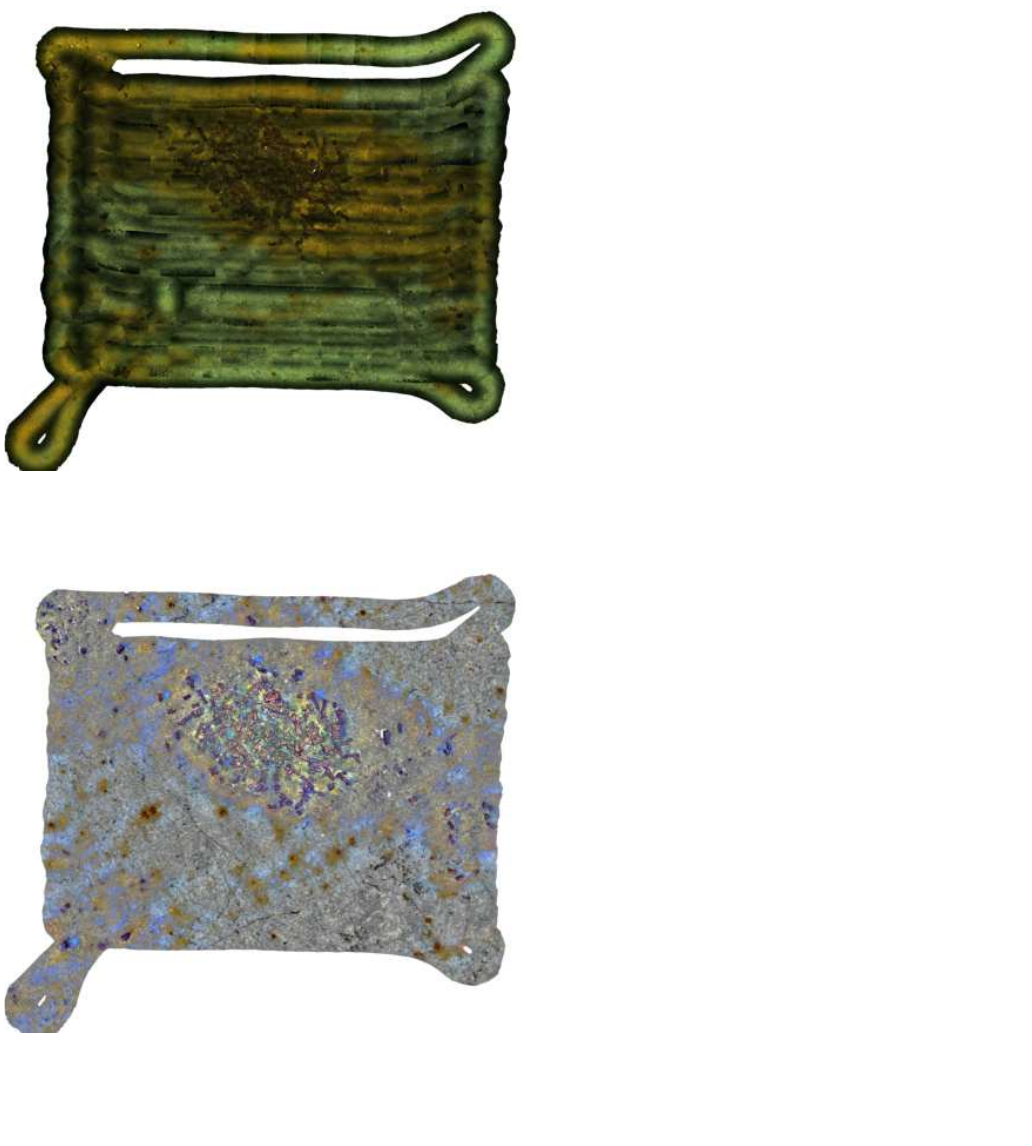}}%
    \put(0.05701359,0.62007035){\makebox(0,0)[lt]{\lineheight{1.25}\smash{\begin{tabular}[t]{l}10$m$\end{tabular}}}}%
    \put(0,0){\includegraphics[width=\unitlength,page=2]{dense_easy.pdf}}%
  \end{picture}%
\endgroup%

	\caption{The final results of meshing, texturing and orthophoto generation on \textit{Easy} dataset.}
	\label{fig:meshing_easy}
\end{figure*}

\begin{figure*}
	\centering
	\def\s{0.98\textwidth}
	\begingroup%
  \makeatletter%
  \providecommand\color[2][]{%
    \errmessage{(Inkscape) Color is used for the text in Inkscape, but the package 'color.sty' is not loaded}%
    \renewcommand\color[2][]{}%
  }%
  \providecommand\transparent[1]{%
    \errmessage{(Inkscape) Transparency is used (non-zero) for the text in Inkscape, but the package 'transparent.sty' is not loaded}%
    \renewcommand\transparent[1]{}%
  }%
  \providecommand\rotatebox[2]{#2}%
  \newcommand*\fsize{\dimexpr\f@size pt\relax}%
  \newcommand*\lineheight[1]{\fontsize{\fsize}{#1\fsize}\selectfont}%
  \ifx\svgwidth\undefined%
    \setlength{\unitlength}{495bp}%
    \ifx\svgscale\undefined%
      \relax%
    \else%
      \setlength{\unitlength}{\unitlength * \real{\svgscale}}%
    \fi%
  \else%
    \setlength{\unitlength}{\svgwidth}%
  \fi%
  \global\let\svgwidth\undefined%
  \global\let\svgscale\undefined%
  \makeatother%
  \begin{picture}(1,1.15151515)%
    \lineheight{1}%
    \setlength\tabcolsep{0pt}%
    \put(0.42,0.01370376){\makebox(0,0)[lt]{\lineheight{1.25}\smash{\begin{tabular}[t]{l}\textbf{Normalized Color}\end{tabular}}}}%
    \put(0.42,0.62282815){\makebox(0,0)[lt]{\lineheight{1.25}\smash{\begin{tabular}[t]{l}\textbf{Original Color}\end{tabular}}}}%
    \put(0,0){\includegraphics[width=\unitlength,page=1]{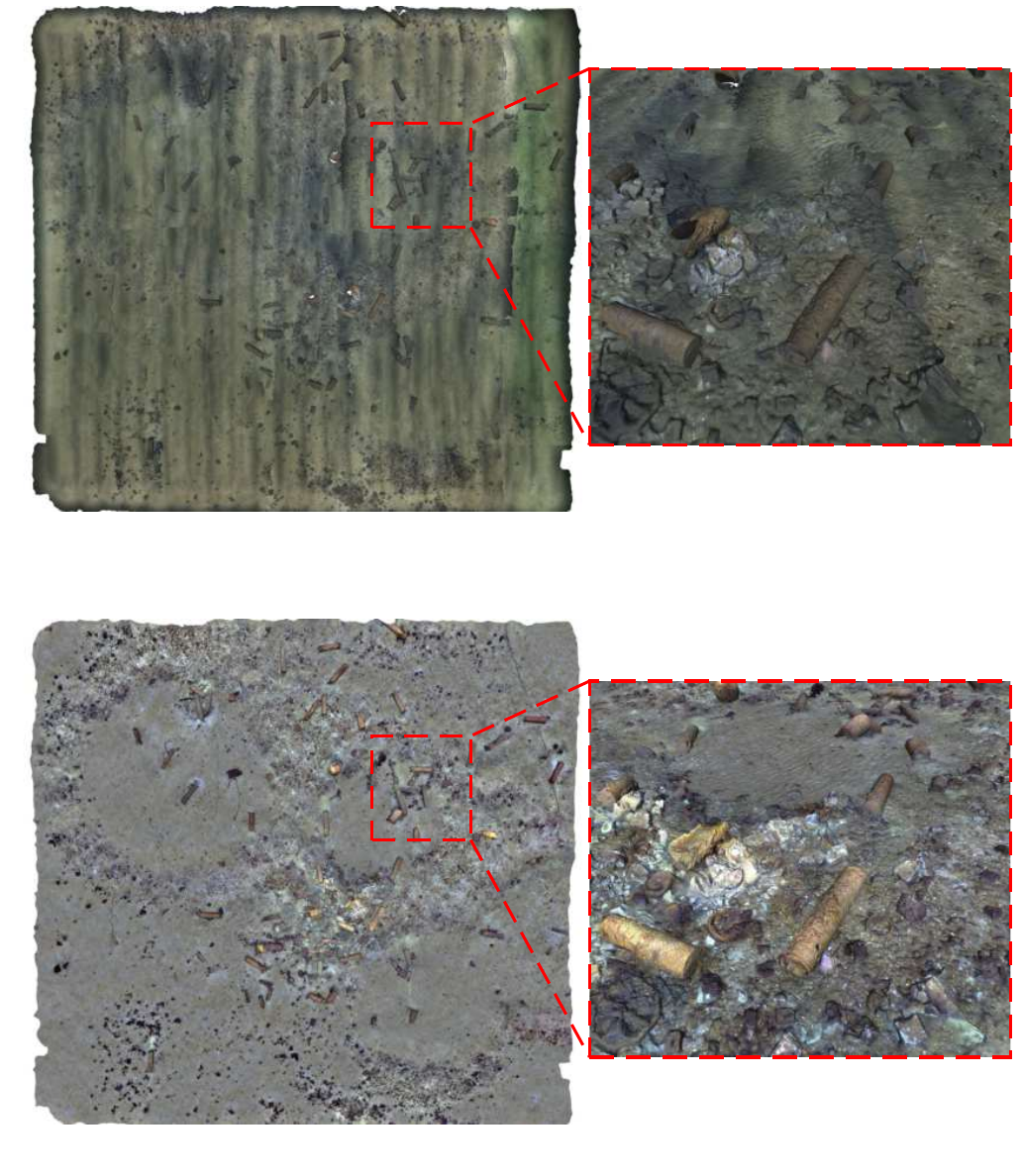}}%
    \put(0.08031121,0.60686062){\makebox(0,0)[lt]{\lineheight{1.25}\smash{\begin{tabular}[t]{l}10$m$\end{tabular}}}}%
    \put(0,0){\includegraphics[width=\unitlength,page=2]{dense_medium1.pdf}}%
  \end{picture}%
\endgroup%

	\caption{The final results of meshing, texturing and orthophoto generation on \textit{Medium1} dataset.}
	\label{fig:meshing_medium1}
\end{figure*}

\begin{figure*}
	\centering
	\def\s{0.98\textwidth}
	\begingroup%
  \makeatletter%
  \providecommand\color[2][]{%
    \errmessage{(Inkscape) Color is used for the text in Inkscape, but the package 'color.sty' is not loaded}%
    \renewcommand\color[2][]{}%
  }%
  \providecommand\transparent[1]{%
    \errmessage{(Inkscape) Transparency is used (non-zero) for the text in Inkscape, but the package 'transparent.sty' is not loaded}%
    \renewcommand\transparent[1]{}%
  }%
  \providecommand\rotatebox[2]{#2}%
  \newcommand*\fsize{\dimexpr\f@size pt\relax}%
  \newcommand*\lineheight[1]{\fontsize{\fsize}{#1\fsize}\selectfont}%
  \ifx\svgwidth\undefined%
    \setlength{\unitlength}{495bp}%
    \ifx\svgscale\undefined%
      \relax%
    \else%
      \setlength{\unitlength}{\unitlength * \real{\svgscale}}%
    \fi%
  \else%
    \setlength{\unitlength}{\svgwidth}%
  \fi%
  \global\let\svgwidth\undefined%
  \global\let\svgscale\undefined%
  \makeatother%
  \begin{picture}(1,1.27272727)%
    \lineheight{1}%
    \setlength\tabcolsep{0pt}%
    \put(0.42,0.01673539){\makebox(0,0)[lt]{\lineheight{1.25}\smash{\begin{tabular}[t]{l}\textbf{Normalized Color}\end{tabular}}}}%
    \put(0.42,0.64010085){\makebox(0,0)[lt]{\lineheight{1.25}\smash{\begin{tabular}[t]{l}\textbf{Original Color}\end{tabular}}}}%
    \put(0.04588725,1.21109964){\makebox(0,0)[lt]{\lineheight{1.25}\smash{\begin{tabular}[t]{l}10$m$\end{tabular}}}}%
    \put(0,0){\includegraphics[width=\unitlength,page=1]{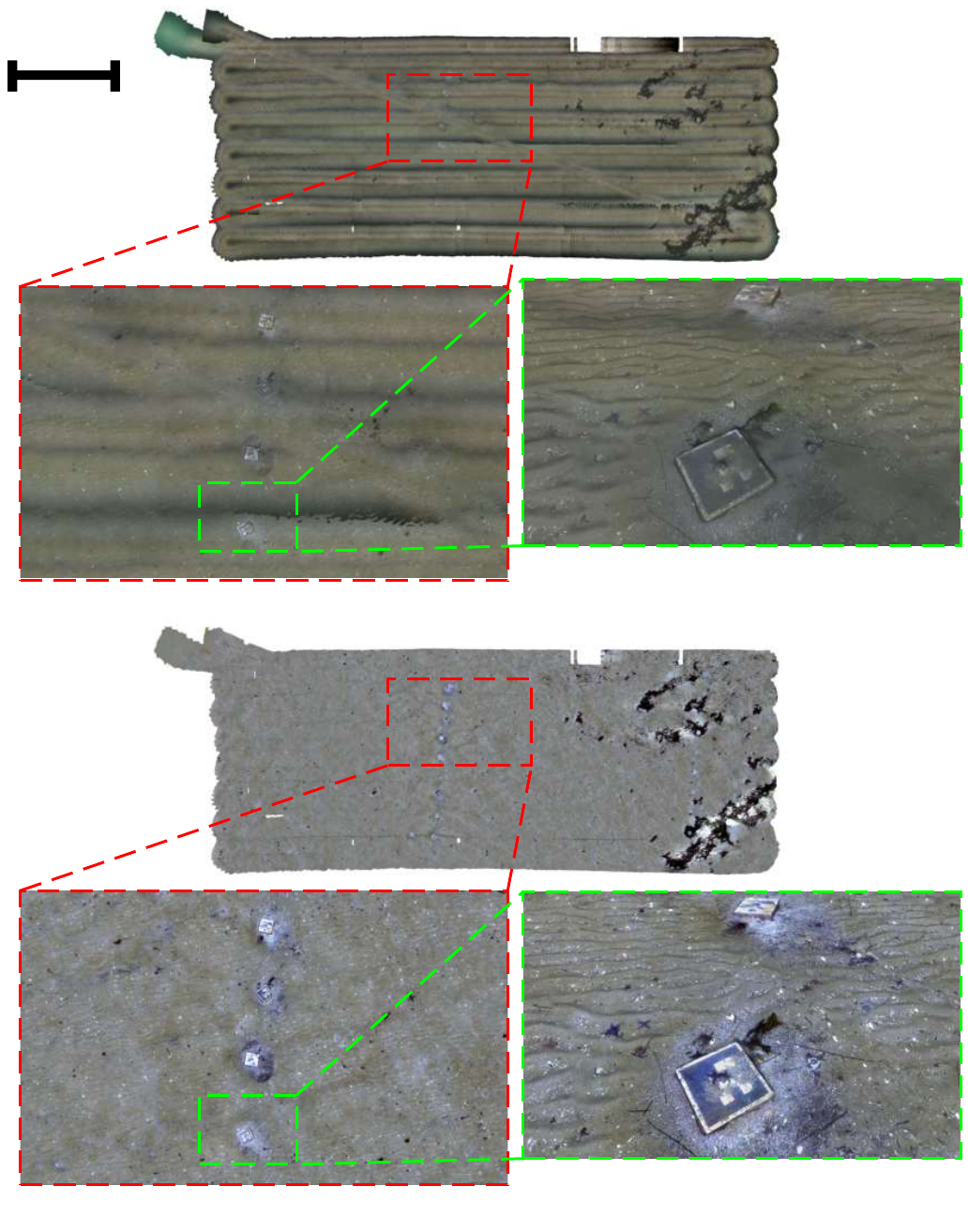}}%
  \end{picture}%
\endgroup%

	\caption{The final results of meshing, texturing and orthophoto generation on \textit{Medium2} dataset.}
	\label{fig:meshing_medium2}
\end{figure*}

\begin{figure*}
	\centering
	\def\s{0.98\textwidth}
	\begingroup%
  \makeatletter%
  \providecommand\color[2][]{%
    \errmessage{(Inkscape) Color is used for the text in Inkscape, but the package 'color.sty' is not loaded}%
    \renewcommand\color[2][]{}%
  }%
  \providecommand\transparent[1]{%
    \errmessage{(Inkscape) Transparency is used (non-zero) for the text in Inkscape, but the package 'transparent.sty' is not loaded}%
    \renewcommand\transparent[1]{}%
  }%
  \providecommand\rotatebox[2]{#2}%
  \newcommand*\fsize{\dimexpr\f@size pt\relax}%
  \newcommand*\lineheight[1]{\fontsize{\fsize}{#1\fsize}\selectfont}%
  \ifx\svgwidth\undefined%
    \setlength{\unitlength}{495bp}%
    \ifx\svgscale\undefined%
      \relax%
    \else%
      \setlength{\unitlength}{\unitlength * \real{\svgscale}}%
    \fi%
  \else%
    \setlength{\unitlength}{\svgwidth}%
  \fi%
  \global\let\svgwidth\undefined%
  \global\let\svgscale\undefined%
  \makeatother%
  \begin{picture}(1,1.11111111)%
    \lineheight{1}%
    \setlength\tabcolsep{0pt}%
    \put(0.42,0.01269374){\makebox(0,0)[lt]{\lineheight{1.25}\smash{\begin{tabular}[t]{l}\textbf{Normalized Color}\end{tabular}}}}%
    \put(0.42,0.60969621){\makebox(0,0)[lt]{\lineheight{1.25}\smash{\begin{tabular}[t]{l}\textbf{Original Color}\end{tabular}}}}%
    \put(0,0){\includegraphics[width=\unitlength,page=1]{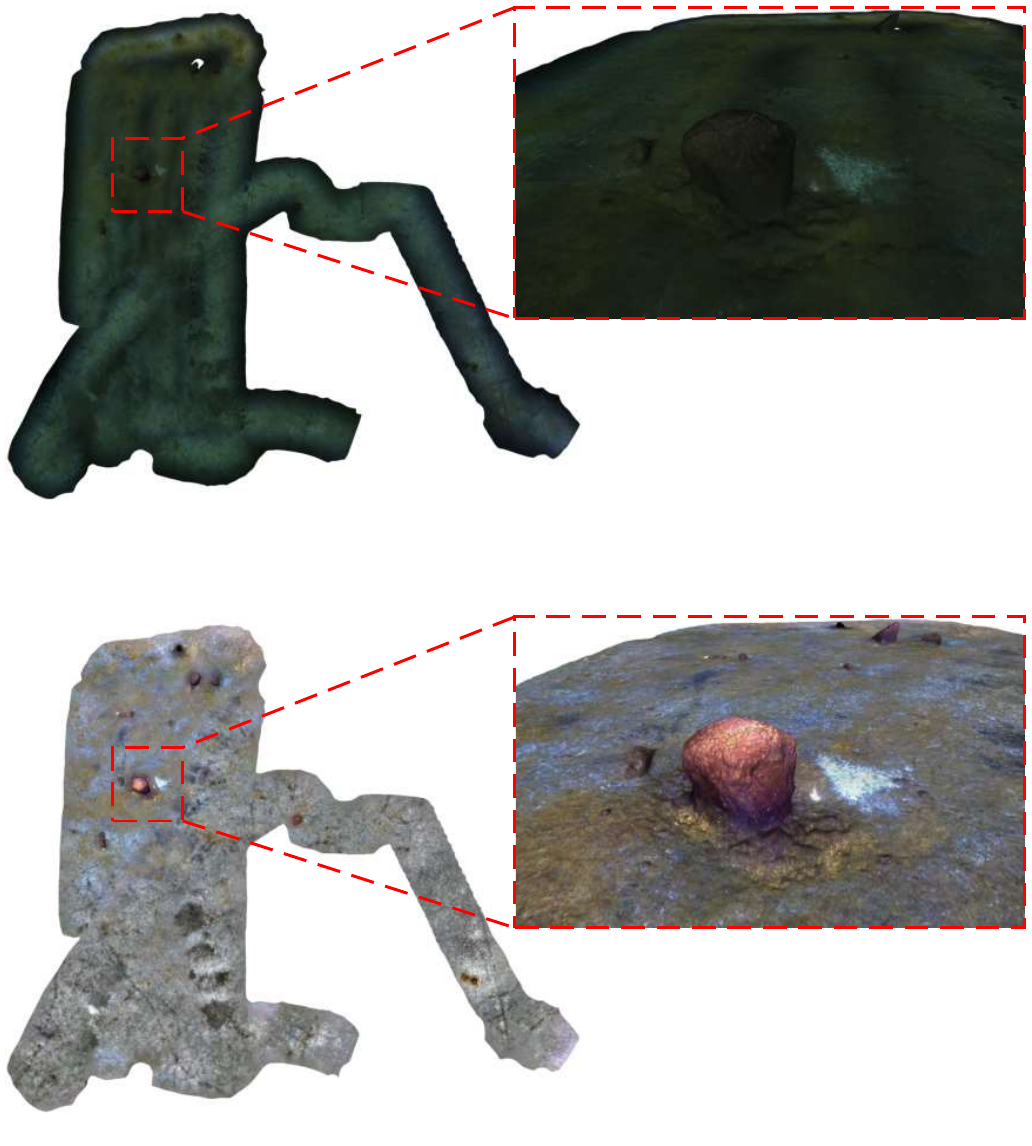}}%
    \put(0.13436062,0.60019063){\makebox(0,0)[lt]{\lineheight{1.25}\smash{\begin{tabular}[t]{l}8$m$\end{tabular}}}}%
    \put(0,0){\includegraphics[width=\unitlength,page=2]{dense_hard1.pdf}}%
  \end{picture}%
\endgroup%

	\caption{The final results of meshing, texturing and orthophoto generation on \textit{Hard1} dataset.}
	\label{fig:meshing_hard1}
\end{figure*}

\begin{figure*}
	\centering
	\def\s{0.98\textwidth}
	\begingroup%
  \makeatletter%
  \providecommand\color[2][]{%
    \errmessage{(Inkscape) Color is used for the text in Inkscape, but the package 'color.sty' is not loaded}%
    \renewcommand\color[2][]{}%
  }%
  \providecommand\transparent[1]{%
    \errmessage{(Inkscape) Transparency is used (non-zero) for the text in Inkscape, but the package 'transparent.sty' is not loaded}%
    \renewcommand\transparent[1]{}%
  }%
  \providecommand\rotatebox[2]{#2}%
  \newcommand*\fsize{\dimexpr\f@size pt\relax}%
  \newcommand*\lineheight[1]{\fontsize{\fsize}{#1\fsize}\selectfont}%
  \ifx\svgwidth\undefined%
    \setlength{\unitlength}{495bp}%
    \ifx\svgscale\undefined%
      \relax%
    \else%
      \setlength{\unitlength}{\unitlength * \real{\svgscale}}%
    \fi%
  \else%
    \setlength{\unitlength}{\svgwidth}%
  \fi%
  \global\let\svgwidth\undefined%
  \global\let\svgscale\undefined%
  \makeatother%
  \begin{picture}(1,1.36363636)%
    \lineheight{1}%
    \setlength\tabcolsep{0pt}%
    \put(0,0){\includegraphics[width=\unitlength,page=1]{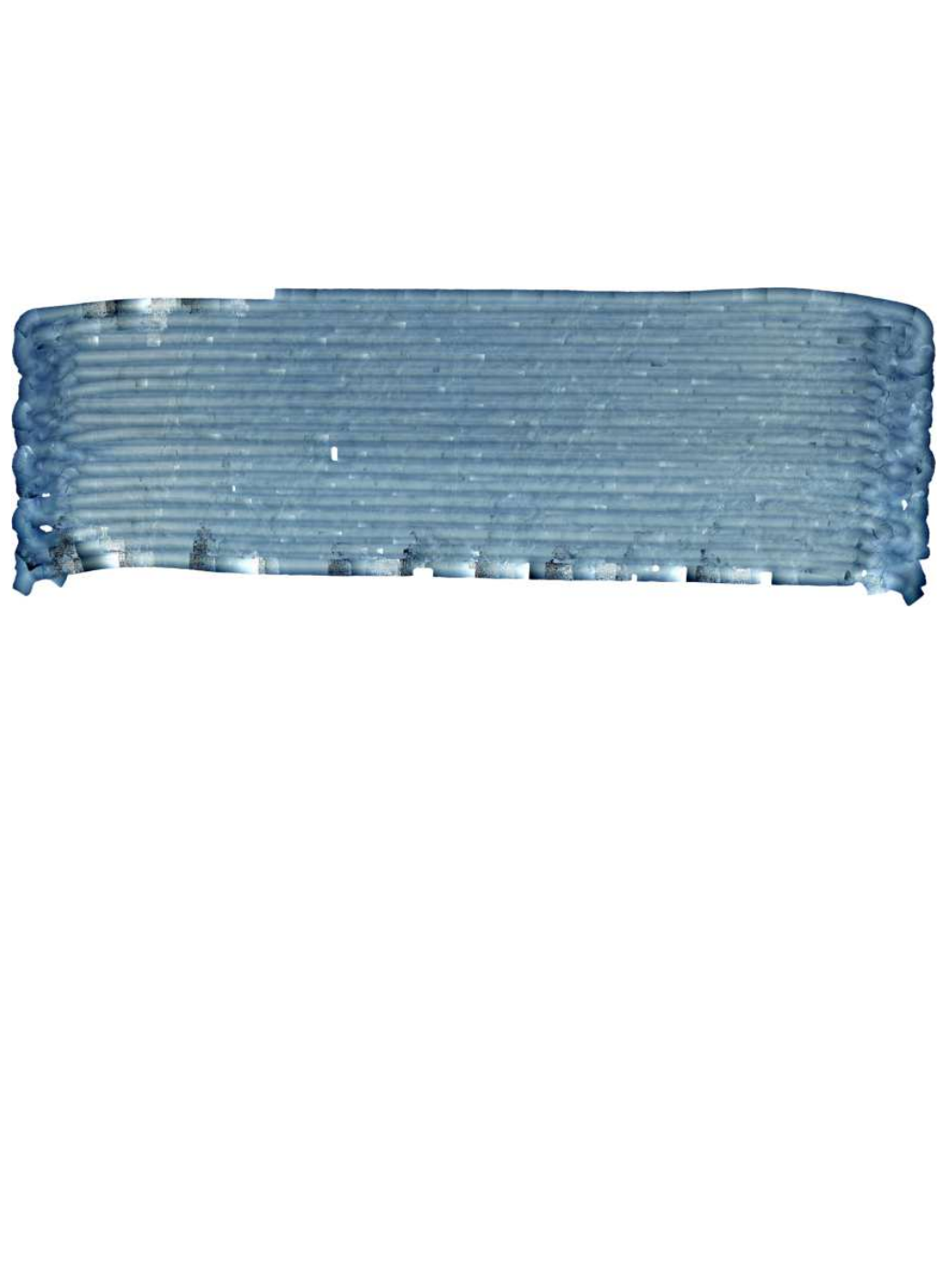}}%
    \put(0.42,0.70093224){\makebox(0,0)[lt]{\lineheight{1.25}\smash{\begin{tabular}[t]{l}\textbf{Original Color}\end{tabular}}}}%
    \put(0.42,0.0167284){\makebox(0,0)[lt]{\lineheight{1.25}\smash{\begin{tabular}[t]{l}\textbf{Normalized Color}\end{tabular}}}}%
    \put(0,0){\includegraphics[width=\unitlength,page=2]{dense_hard2.pdf}}%
    \put(0.12036971,0.72017513){\makebox(0,0)[lt]{\lineheight{1.25}\smash{\begin{tabular}[t]{l}100$m$\end{tabular}}}}%
    \put(0,0){\includegraphics[width=\unitlength,page=3]{dense_hard2.pdf}}%
  \end{picture}%
\endgroup%

	\caption{The final results of meshing, texturing and orthophoto generation on \textit{Hard2} dataset.}
	\label{fig:meshing_hard2}
\end{figure*}

Fig. \ref{fig:meshing_easy} to Fig. \ref{fig:meshing_hard2} display the final results of the last step of the visual mapping pipeline on the given datasets.
We apply chunk-based dense Multi-View Stereo (MVS) estimation and meshing using the OpenMVS \cite{openmvs2020} package on both the original color images and on the normalized color images.
It is evident that the lighting effects are largely compensated in the final output mesh, resulting in a clearer and more visually appealing scene structure.
The zoomed views indicate that we are able to achieve a high-resolution and detailed 3D reconstruction of the seafloor.

\section{Conclusion}
In this work, we presented a fully automated, and comprehensive workflow for mapping large areas of the seafloor, leveraging the recent developments in both underwater imaging and visual mapping techniques.
Our main focus was on achieving a geometrically consistent and accurate reconstruction of the seafloor.
To this end, we carefully consider refraction to avoid reconstruction biases and proposed a navigation-aided hierarchical mapping approach that combines the benefits of SLAM and the global SfM.
Through a thorough evaluation on multiple datasets with varying characteristics, we demonstrated the effectiveness and accuracy of our system.
The proposed work will serve as a baseline for future developments to enable more robust and accurate deep-sea-specific visual mapping.

\begin{acknowledgements}
	This publication has been funded by the German Research Foundation (Deutsche Forschungsgemeinschaft, DFG) Projektnummer 396311425, through the Emmy Noether Programme. We are also grateful for support from the Chinese Scholarship Council (CSC) for M. She (202006050015) and Y. Song (201608080215).
	We would also like to thank CSSF, Schmidt Ocean Institute, GEOMAR AUV and JAGO Team for providing the underwater image materials.
\end{acknowledgements}

\bibliographystyle{spmpsci}      %
\bibliography{reference.bib}   %

\end{document}